\documentstyle[epsf,epsfig,makeidx,amssymb]{style-directory/elsart}

\def\wrt{\mbox{w.r.t.}}
\def\traidelt{{\cal I}_{\dlines}^{23}}
\def\alcd{{\cal ALC}({\cal D})}

\def\rcc8{\mbox{${\cal RCC}$8}}

\def\rectalg{\mbox{${\cal R}$g${\cal A}$}}
\def\prod{\pi}
\def\region{{\mbox{{\em reg}}}}

\def\allenb{<}
\def\allena{>}
\def\allenm{\mbox{{\em m}}}
\def\allenmi{\mbox{{\em mi}}}
\def\alleno{\mbox{{\em o}}}
\def\allenoi{\mbox{{\em oi}}}
\def\allens{\mbox{{\em s}}}
\def\allensi{\mbox{{\em si}}}
\def\allend{\mbox{{\em d}}}
\def\allendi{\mbox{{\em di}}}
\def\allenf{\mbox{{\em f}}}
\def\allenfi{\mbox{{\em fi}}}
\def\alleneq{\mbox{{\em eq}}}
\def\rcc8{\mbox{{\em RCC8}}}

\def\rectalg{\mbox{${\cal R}$g${\cal A}$}}
\def\twoandthreecutone{\mbox{{\em 2and3cut1}}}
\def\pairwisecutting{\mbox{{\em PairwiseCutting}}}
\def\tractablesubset{{\cal S}}
\def\icpa{\mbox{{\em IC-pa}}}
\def\icsa{\mbox{{\em IC-sa}}}
\def\renzbe{\mbox{behind}_=}
\def\renzbd{\mbox{behind}_{\not =}}
\def\renzfe{\mbox{in-front-of}_=}
\def\renzfd{\mbox{in-front-of}_{\not =}}
\def\renzmbe{\mbox{meets-from-behind}_=}
\def\renzmbd{\mbox{meets-from-behind}_{\not =}}
\def\renzmfe{\mbox{meets-in-the-front}_=}
\def\renzmfd{\mbox{meets-in-the-front}_{\not =}}
\def\renzobe{\mbox{overlaps-from-behind}_=}
\def\renzobd{\mbox{overlaps-from-behind}_{\not =}}
\def\renzofe{\mbox{overlaps-in-the-front}_=}
\def\renzofd{\mbox{overlaps-in-the-front}_{\not =}}
\def\renzce{\mbox{contained-in}_=}
\def\renzcd{\mbox{contained-in}_{\not =}}
\def\renzee{\mbox{extends}_=}
\def\renzed{\mbox{extends}_{\not =}}
\def\renzcbe{\mbox{contained-in-the-back-of}_=}
\def\renzcbd{\mbox{contained-in-the-back-of}_{\not =}}
\def\renzefe{\mbox{extends-the-front-of}_=}
\def\renzefd{\mbox{extends-the-front-of}_{\not =}}
\def\renzebe{\mbox{extends-the-back-of}_=}
\def\renzebd{\mbox{extends-the-back-of}_{\not =}}
\def\renzcfe{\mbox{contained-in-the-front-of}_=}
\def\renzcfd{\mbox{contained-in-the-front-of}_{\not =}}
\def\renzeqe{\mbox{equals}_=}
\def\renzeqd{\mbox{equals}_{\not =}}
\def\renzrl{\ell _{Renz}}
\def\dipolesa{{\cal D}_{24}}
\def\dipoleca{{\cal D}_{69}}
\def\dipolese{\mbox{\bf s}}
\def\dipoleee{\mbox{\bf e}}
\def\birel{{\cal I}_U^b}
\def\tirel{{\cal I}_U^t}

\def\algats{{\cal U}\mbox{-at}}

\def\rel-alg{{\cal R}}
\def\fbra{{\cal FB}}
\def\ftra{{\cal FT}}
\def\binrel{\mbox{\em binRel}}
\def\terrel{\mbox{\em terRel}}
\def\btwp{\mbox{\em btw\_p}}
\def\btwdl{\mbox{\em btw\_dl}}
\def\noncoll{\mbox{\em non\_coll}}
\def\angleg{\prec}
\def\angled{\succ}
\def\lhp{\mbox{\em lhp}}
\def\rhp{\mbox{\em rhp}}
\def\dlats{\mbox{$\dlalg${\em -at}}}
\def\cdlats{\mbox{$\cdlalg${\em -at}}}
\def\cdlar{\mbox{$\cdlalg${\em -ar}}}
\def\atraats{\mbox{$\atra${\em -at}}}
\def\dltats{\mbox{$\dltalg${\em -at}}}
\def\gis{\mbox{\em GIS}}

\def\ppartition{\mbox{\em p-partition}}
\def\ppregion{\mbox{\em pp-region}}
\def\lpartition{\mbox{\em line-partition}}
\def\lpregion{\mbox{\em lp-region}}
\def\eee{\mbox{\em eee}}
\def\elll{\mbox{\em ell}}
\def\eoo{\mbox{\em eoo}}
\def\err{\mbox{\em err}}
\def\lel{\mbox{\em lel}}
\def\lll{\mbox{\em lll}}
\def\llo{\mbox{\em llo}}
\def\llr{\mbox{\em llr}}
\def\lor{\mbox{\em lor}}
\def\lre{\mbox{\em lre}}
\def\lrl{\mbox{\em lrl}}
\def\lrr{\mbox{\em lrr}}
\def\oeo{\mbox{\em oeo}}
\def\olr{\mbox{\em olr}}
\def\ooe{\mbox{\em ooe}}
\def\orl{\mbox{\em orl}}
\def\rer{\mbox{\em rer}}
\def\rle{\mbox{\em rle}}
\def\rll{\mbox{\em rll}}
\def\rlr{\mbox{\em rlr}}
\def\rol{\mbox{\em rol}}
\def\rrl{\mbox{\em rrl}}
\def\rro{\mbox{\em rro}}
\def\rrr{\mbox{\em rrr}}
\def\compatible{\mbox{\em comp}}
\def\region{\mbox{\em reg}}

\def\orient{{\cal O}}
\def\BBR{{\rm I\!R}}

\def\cqfd{\vrule height 1.2ex depth 0ex width 1.2ex}
\def\proj{\bigtriangledown}

\def\apra{{\cal CYC}_b}
\def\atra{{\cal CYC}_t}

\def\cyc{\mbox{\em cyc}}
\def\ro1{{\cal R}_{0,1}}
\def\points{\mbox{{\em pts}}}
\def\dlo{\overrightarrow{{\cal L}}_O}
\def\dlines{\overrightarrow{{\cal L}}}
\def\ulines{\overline{{\cal L}}}
\def\dlignes{\mbox{d-lines}}
\def\dligne{\mbox{d-line}}
\def\ulignes{\mbox{u-lines}}
\def\uligne{\mbox{u-line}}
\def\lines{{\cal L}}
\def\dltalg{{\cal TA}_t}
\def\dlalg{{\cal PA}_t}
\def\cdlalg{c{\cal PA}_t}

\def\cuts{\mbox{{\em cuts}}}
\def\coincides{\mbox{{\em coinc-with}}}
\def\parallel{\mbox{{\em s-par-to}}}
\def\incidence{\mbox{{\em inc-with}}}
\def\leftpar{\mbox{{\em l-par-to}}}
\def\rightpar{\mbox{{\em r-par-to}}}
\def\parallelity{\mbox{{\em par-to}}}
\def\cc{\mbox{{\em cc}}}
\def\cp{\mbox{{\em cp}}}
\def\pc{\mbox{{\em pc}}}
\def\pp{\mbox{{\em pp}}}

\def\co1{{\cal C}_{O,1}}
\def\binmat{{\cal B}}
\def\termat{{\cal T}}

\def\ltpa{{\cal PA}}
\def\coone{{\cal C}_{O,1}}

\def\isodeux{{\cal I}_1}
\def\isotrois{{\cal I}_2}
\def\deuxdo{\mbox{2D}{\cal O}}

\begin{document}
\begin{frontmatter}
\title{A ternary Relation Algebra of directed lines}
\thanks{This work was supported partly by the DFG project ``{\em Description
    Logics and Spatial reasoning}" (DLS), under grant {\em NE 279/8-1}, and partly by
    the EU project ``{\em Cognitive Vision systems}" (CogVis), under grant {\em CogVis IST 2000-29375}.}
\author{Amar Isli}
\address{Fachbereich Informatik, Universit\"at Hamburg,\\
         Vogt-K\"olln-Strasse 30, D-22527 Hamburg, Germany\\
         isli@informatik.uni-hamburg.de
        }
\begin{abstract}
We define a ternary Relation Algebra (RA) of relative position relations
on two-dimensional directed lines ($\dlignes$ for short).
A $\dligne$ has two degrees of
freedom (DFs): a rotational DF (RDF), and a translational DF (TDF). The
representation of the RDF of a $\dligne$ will be handled by an RA of 2D
orientations, $\atra$, known in the literature. A second algebra,
$\dltalg$, which will handle the TDF of a $\dligne$, will be defined. The two
algebras, $\dltalg$ and $\atra$, will constitute, respectively, the translational
and the rotational components of the RA, $\dlalg$, of relative position
relations on $\dlignes$: the $\dlalg$ atoms will consist of those pairs
$\langle t,r\rangle$ of a $\dltalg$ atom and a $\atra$ atom that are compatible.
We present in detail the RA $\dlalg$, with its converse
table, its rotation table and its composition tables. We show
that a (polynomial) constraint propagation algorithm, known in the literature, is
complete for a subset of $\dlalg$ relations including almost all of the atomic relations. We will discuss
the application scope of the RA, which includes incidence geometry, GIS
(Geographic Information Systems), shape representation, localisation in
(multi-)robot navigation, and the representation of motion prepositions
in NLP (Natural Language Processing). We then compare the RA to existing
ones, such as an algebra for reasoning about rectangles parallel to the
axes of an (orthogonal) coordinate system, a ``spatial Odyssey'' of Allen's
interval algebra, and an algebra for reasoning about 2D segments.
\begin{keyword}
  Relation algebra,
  Spatial reasoning,
  Qualitative reasoning,
  Geometric reasoning,
  Constraint satisfaction,
  Knowledge representation.
\end{keyword}
\end{abstract}
\newtheorem{corol}{Corollary}
\newtheorem{remk}{Remark}
\newtheorem{thr}{Theorem}
\newtheorem{df}{Definition}
\newtheorem{ex}{Example}
\end{frontmatter}
\maketitle
\section{Introduction}\label{sect1}
Qualitative Spatial Reasoning (QSR), and more generally Qualitative Reasoning
(QR), distinguishes from quantitative reasoning by its particularity of
remaining at a description level as high as possible. In other words, QSR
sticks at the idea of ``making only as many distinctions as necessary''
\cite{Cohn97b,Freksa92b}, idea borrowed to na\"{\i}ve physics
\cite{Hayes85b}. The core motivation behind this is that, whenever the
number of distinctions that need to be made is finite, the reasoning issue
can get rid of the calculations details of quantitative models, and be
transformed into a simple matter of symbols manipulation; in the particular
case of constraint-based spatial and temporal reasoning, this means a finite
relation algebra (finite RA), with tables recording the results of applying
the different operations to the different atoms, and the reasoning issue
reduced to a matter of table look-ups: the best illustration to this is
certainly Allen's \cite{Allen83b} algebra of time intervals.

The main problem in designing a QSR language is certainly to come up with the
right, cognitively adequate, distinctions that need to be made; this problem is
often referred to as the qualitative/quantitative dilemma, or the
finiteness/density dilemma \cite{Habel95a} (how to distinguish
between the infinite number of elements of an ---infinite--- universe using
only a finite number of distinctions?): to say it another way,
because such a language can make only a finite number of distinctions, it
should reflect as good as possible the real world; ideally, such a language
would be such that it distinguishes between two situations if and only if
Humans, or the agents expected to use the language, do distinguish between the
two situations. Qualitative reasoning had to face criticism ---examples include
Forbus, Nielsen and Faltings' \cite{Forbus91a} poverty conjecture, or Habel's
\cite{Habel95a} argument that
such a language, even when built according to cognitive adequacy criteria,
still suffers from not having ``the ability to refine discrete structures if
necessary''. The tendency has since then changed, due
certainly to the success gained by QSR in real applications, such as $\gis$,
robot navigation, or shape description.

QSR has now its place in AI. Its research has focussed for about a decade on
aspects such as topology, orientation and distance. The aspect the most
developed so far is topology, illustrated by the well-known RCC theory
\cite{Randell92a}, from which derives the RCC-8 calculus
\cite{Randell92a,Egenhofer91a}. The RCC theory, on the other hand, stems from
Clarke's ``calculus of individuals'' \cite{Clarke81a}, based on a
binary ``connected with'' relation ---sharing of a point of the arguments.
Clarke's work, in turn, was developed from classical mereology
\cite{Leonard40a,Lesniewski28a} and Whitehead's
``extensionally connected with'' relation \cite{Whitehead29a}. The huge
interest, the last couple of years, in applications such as robot navigation,
illustrated by active and promising RoboCup soccer meetings at the main AI
conferences (IJCAI, AAAI: see, for instance, \cite{robocup99} for
RoboCup'99), had and still have as a consequence
that relative orientation, and, more generally, relative position, considered as
expressing more specific knowledge, are gaining increasing interest from the QSR
community.

The research in QSR has reached a point where the integration of different aspects of knowledge,
such as relative orientation and relative distance, topology
and orientation, or, as in the present work, relative orientation
and relative translation, is more than needed, in order to face the new
demand of real applications. Such an integration of different aspects of knowledge is seen
as position, because a calculus coming from such a combination, if it cannot
represent the position of an object as precisely as do quantitative models, yet
provides a representation more specific than the ones of the combined calculi.
It seems to be the case that all researchers in the area are aware of
the problem \cite{Cohn97b,Escrig97c,Frank92b,Freksa92b}. When looking at what has
really been achieved so far in this direction, apart from the work in
\cite{Clementini97b}, and, more recently, the one in \cite{Gerevini98a}, not much
can be said.

In this work, we consider the geometric element consisting of a
(2-dimensional) directed line ($\dligne$ for short). Such
an element has two degrees of freedom (DFs) \cite{Hartenberg64a,Kramer92a}: a rotational
DF (RDF) and a translational DF (TDF). The RDF, on the one hand, constrains the way
a $\dligne$ can rotate relative to another $\dligne$ (relative orientation); the TDF,
on the other hand, constrains the way a $\dligne$ can translate relative to another,
or other, $\dlignes$, so that once the RDF of a $\dligne$, say $\ell _1$, has been
``absorbed'' (i.e., its orientation has been fixed), we know how to translate
$\ell _1$ (a move parallel to the orientation), so that its desired position gets fixed,
and its TDF absorbed (e.g., translate $\ell _1$ so that its intersecting point with a
second $\dligne$ $\ell _2$ comes before the intersecting point of a third $\dligne$
$\ell _3$ with $\ell _2$, when we walk along $\ell _2$ heading the positive
direction; or so that $\ell _1$ is parallel to both, and does not lie
between, $\ell _2$ and $\ell _3$). We can right now notice a point of high importance
for the TDF of $\dlignes$, which is that their oriented-ness makes them much richer
than undirected lines, or $\ulignes$: contrary to $\ulignes$:
\begin{enumerate}
  \item when walking along a $\dligne$, we know whether we are heading the positive or
    the negative direction; and
  \item when walking perpendicularly to a $\dligne$, we know whether we are heading
    towards the right half-plane or towards the left half-plane bounded by the
    $\dligne$.
\end{enumerate}
We provide a ternary Relation Algebra (RA), $\dlalg$,
of relative position relations on $\dlignes$; the way we
proceed is, somehow, imposed by the two DFs of a $\dligne$:
\begin{enumerate}
  \item a ternary RA of 2D orientations, $\atra$, recently known in the literature
    \cite{Isli98a,Isli00b}, will be the rotational component of $\dlalg$; and
  \item a second ternary algebra, $\dltalg$, which will constitute the translational
    component of $\dlalg$, will be defined.
\end{enumerate}
The $\dlalg$ atoms will consist of those pairs $\langle t,r\rangle$ of a $\dltalg$
atom and a $\atra$ atom that are compatible. The work can be seen as a full
axiomatisation, given as a ternary RA, with its converse table, its rotation table and
its composition tables, of qualitative geometry \cite{Balbiani94a,Bennett00a}, with
parallelity and cutting-ness, and with $\dlignes$ as the primitive entities. It should
be emphasised here that, thanks, again, to the oriented-ness of $\dlignes$:
\begin{enumerate}
  \item  parallelity, on the one hand, splits into six relations,
    ``parallel to, of same orientation as, and lies in the left half-plane bounded by'',
    ``parallel to, of same orientation as, and coincides with'',
    ``parallel to, of same orientation as, and lies in the right half-plane bounded by'',
    ``parallel to, of opposite orientation than, and lies in the left half-plane bounded by'',
    ``parallel to, of opposite orientation than, and coincides with'',
    ``parallel to, of opposite orientation than, and lies in the right half-plane bounded by''.
    This allows distinguishing, on the one hand, between equal orientations and
    opposite orientations, and, on the other hand, between the parallels to a fixed
    $\dligne$ that lie in the left half-plane bounded by, the ones that coincide
    with, and the ones that lie in the right half-plane bounded by, the $\dligne$.
    Had we $\ulignes$ instead of $\dlignes$, parallelity would split into two
    relations, ``coincides with'' and ``strictly parallel to''; and
  \item in a similar way, cutting-ness splits into two relations, ``cuts, and to the left
    of'' and ``cuts, and to the right of''. Cuttingness of $\ulignes$ is atomic.
\end{enumerate}

Using the RA $\dlalg$, we can represent knowledge on relative position of $\dlignes$ as a
CSP (Constraint Satisfaction Problem) \cite{Mackworth77a,Montanari74a}, of which:
\begin{enumerate}
  \item the variables range over the set $\dlines$ of $\dlignes$, and
  \item the constraints consist of $\dlalg$ relations on (triples of) the $\dligne$ variables.
\end{enumerate}
In addition to a full definition of the RA $\dlalg$, with its converse table, its rotation
table and its composition tables, we show the important result that, a (polynomial)
$4$-consistency algorithm known in the literature \cite{Isli98a,Isli00b}, is complete for the
atomic relations of a coarser version, $\cdlalg$, of $\dlalg$: a CSP
such that the constraint on each triple of variables is a $\cdlalg$ atomic relation, can be
checked for consistency using the propagation algorithm in \cite{Isli98a,Isli00b}. Solving a
general $\cdlalg$ CSP, in turn, can be achieved using a solution search algorithm,
such as Isli and Cohn's, also in \cite{Isli98a,Isli00b}. The set of $\cdlalg$ atomic
relations includes almost all $\dlalg$ atomic relations.

The proof of the result that the $4$-consistency algorithm in
\cite{Isli98a,Isli00b} is complete for CSPs expressed in the set, $\tractablesubset$, of
$\cdlalg$ atomic relations, shows
the importance of degrees of freedom \cite{Hartenberg64a,Kramer92a} for this work. If
such an algorithm applied to such a CSP does not derive the empty relation ---in which
case, the result says that the CSP is consistent--- then, in order to find a spatial
scene that is a model of the CSP, we can proceed as follows:
\begin{enumerate}
  \item Start by getting the RDF absorbed for each of the $\dligne$ variables involved
    in the CSP. In other words, start by fixing the orientation for each of the variables.
    This can be done using a result in \cite{Isli98a,Isli00b}, stating that a
    $4$-consistent atomic CSP of 2D orientations is (globally) consistent: the proof
    of this result gives a backtrack-free method for the construction of a solution
    to such a CSP; the solution can be seen as a set of $\dlignes$ all of which are
    incident with a fixed point ---concurrent $\dlignes$.
  \item Once the RDF has been fixed for each of the $\dligne$ variables, the problem
    has been brought down to a 1D problem: a simple translational problem. The TDF,
    for each of the $\dligne$ variables, has to be fixed: how to translate the
    $\dlignes$ relative to one another, so that the TDF of each of the $\dligne$
    variables gets absorbed (i.e., so that all the $\dltalg$ constraints get
    satisfied) ---see the proof of Theorem \ref{sfcthm} for details.
\end{enumerate}

In the light of the preceding lines, we can provide a plausible explanation to the
question of why QSR researchers have not, so far, sufficiently tackled the emerging challenge
of integrating different spatial aspects. Combining, for instance, an algebra of
relative orientation with an algebra of relative distance, may lead to an algebra with
a high number of atoms (a number that can go up to the product of the
numbers of atoms of the combined algebras). A high number of atoms, in turn, means,
among other things, a big composition table (which is
generally hard to build, sometimes even with the help of a computer ---see the
challenge in \cite{Randell92c}!). QSR languages known so far, particularly the
constraint-based ones, could be described as all-aspects-at-once languages, in the
sense that the way the different spatial aspects, such as, for the present work,
relative orientation and relative translation, corresponding to the different DFs
of the objects ---here $\dlignes$--- in consideration, are treated as
undecomposable: as a consequence, the composition table is simply an
$2$-dimensional table, with $d^2$ entries, $d$ being the number of atoms of the
language. The present work is expected to help changing the tendency, since the
composition of the $\dlalg$ relations is brought down to a matter of a cross product
of the composition of the relations of the translational projection, on the one hand,
and the composition of the relations of the rotational projection, on the other
hand. The method could thus be described as a ``divide and conquer'' one:
\begin{enumerate}
  \item project the knowledge onto the different DFs;
  \item process (compose) the knowledge at the level of each of the projections; and
  \item perform the cross product of the different results in order to get the
    composition of the initial knowledge.
\end{enumerate}
As such, the work can be
looked at as answering, at least partly, the challenges in \cite{Randell92c} for the particular
case of QSR: combining different spatial aspects, in the way it is done in the present
work, does not necessarily increase the difficulties related to composition, because
each of the combined aspects corresponds to one of the DFs
\cite{Hartenberg64a,Kramer92a} of the objects in consideration; the different DFs of
an object, in turn, are, in some sense, independent from each other, so that the
composition issue can be tackled using the ``divide and conquer'' method
referred to above.

Another point worth mentioning is that the current work illustrates the importantce of
the work in \cite{Isli98a,Isli00b}, since, as illustrated by the previous lines, the
RA $\atra$ in \cite{Isli98a,Isli00b} is one of the main two components, specifically
the rotational component, of the main RA, $\dlalg$, investigated here.

The application scope of the RA is large: we
discuss the issue for four application domains, Geographical Information Systems
(GIS), shape representation, robot's panorama description, and the representation of motion prepositions. We also consider
incidence geometry and show how to represent with the RA the incidence of a
point with a (directed) line, betweenness of three points, and non-collinearity of
three points.

We then turn to related work met in the literature, consisting mainly of (relation) algebras
for representing and reasoning about polygonally shaped objects: a dipole algebra \cite{Moratz00a},
important for applications such as cognitive robotics \cite{Musto99a} and spatial information systems
\cite{Hoel91a}; an algebra of directed intervals \cite{Renz01a}; and an algebra of rectangles whose
sides are parallel to the axes of an orthogonal coordinate system of the 2-dimensional Euclidean
space \cite{Balbiani98a,Guesgen89a,Mukerjee90a}. We show that in each case the atomic relations
can be expressed in the RA.

Section \ref{csps} provides some background on constraint-based, or relational,
reasoning; an emphasis is given to relation algebras (RAs).
Section \ref{icra} is
devoted to a quick overview of the ternary RA of 2D orientations in
\cite{Isli00b}.
Section \ref{dlalg} presents in detail the ternary RA of relative
position relations on $\dlignes$.
Section
\ref{dlacsps} deals with ternary CSPs of relative position relations on
$\dlignes$, expressed in the new RA; and shows the result that, if such a CSP
is expressed in the set, $\tractablesubset$, of atomic relations of a coarser
version of the new RA, then a known $4$-consistency algorihm
\cite{Isli98a,Isli00b} either detects
its inconsistency, or, if the CSP is consistent, makes it globally consistent.
Section \ref{dlause} describes the use of the new RA in incidence geometry,
and its applications in domains such as GIS, polygonal shape representation,
(self)localisation of a robot, and the representation of motion prepositions
in natural language.
Section \ref{relatedwork} relates the new RA to similar work in the literature:
Scivos and Nebel's work \cite{Scivos01a} on NP-hardness of Freksa's calculus
\cite{Freksa92b,Zimmermann96a};
Moratz et al.'s dipole algebra \cite{Moratz00a};
Renz's spatial Odyssey \cite{Renz01a} of Allen's interval algebra \cite{Allen83b}; and
the rectangle algebra \cite{Balbiani98a,Guesgen89a,Mukerjee90a}.
Section \ref{conclusion} summarises the work.
\subsection*{Degrees of freedom analysis: the metaphor assembly plan}
The work to be presented has been inspired by the approach to solving geometric
constraint systems, known as Degrees of Freedom Analysis, or DFA for short. For
more details on DFA, the reader is referred to \cite{Kramer92a} (see also
\cite{Kramer90a,Kramer92b}). For
the purpose of this work, we just mention quickly how the inspiration came.
``Degrees of freedom analysis employs
the notion of {\em incremental assembly} as a metaphor for solving
geometric constraint systems ...'' (\cite{Kramer92b}, page 34). The term
{\em incremental} refers to the way the method, knowm as the
{\em Metaphor Assembly Plan (MAP)} \cite{Kramer92b}, proceeds, by fixing
step by step the different degrees of freedom of the object variables
involved in the geometric constraint problem, until all of them have
been fixed, or absorbed, at which stage all objects have been assigned
their right positions, and the problem solved. This inspiration led
to the ternary relation algebra of $\dlignes$ to be presented, which
decomposes into two components, a translational component, handling
the translational degrees of freedom of $\dlignes$, and a rotational
component, handling the rotational degrees of freedom of $\dlignes$. The atoms
of the RA consist of pairs of atoms ---an atom of the translational component
and an atom of the rotational component. More importantly, the different
operations on ternary relations ---converse, rotation and composition---
applied to the RA's atoms, reduce to cross products of the operations
applied to the atoms of the translational component, on the one hand, and to the
atoms of the rotational component, on the other hand. The operations, of which
the most important is composition, could thus be parallelised. Furthermore, as
will be seen in the proof of Theorem \ref{sfcthm}, searching for a solution of a
problem expressed in the RA can be done in a way which has some similar side
with the MAP method used in DFA: we first fix the rotational degrees of freedom
of the $\dligne$ variables, by searching for a solution of the rotational
component; we then fix the translational degrees of freedom by translating,
relative to one another, the $\dlignes$ of the rotational solution.
\section{Constraint satisfaction problems}\label{csps}
The aim of this section is to introduce some background on
constraint-based reasoning.

A constraint satisfaction problem (CSP) of order $n$ consists of:
\begin{enumerate}
  \item a finite set of $n$ variables, $x_1,\ldots ,x_n$;
  \item a set $U$ (called the universe of the problem); and
  \item a set of constraints on values from $U$ which may be assigned to the
    variables.
\end{enumerate}
An $m$-ary constraint is of the form $R(x_{i_1},\cdots ,x_{i_m})$, and asserts
that the \mbox{$m$-tuple} of values assigned to the variables $x_{i_1},\cdots ,x_{i_m}$
must lie in the $m$-ary relation $R$ (an $m$-ary relation over the
universe $U$ is any subset of $U^m$). An $m$-ary CSP is one of which the
constraints are $m$-ary constraints.

Composition and converse for binary relations were introduced by De Morgan
\cite{DeMorgan66a}.
Isli and Cohn \cite{Isli98a,Isli00b} extended the operations
of composition and converse to ternary relations, and
introduced for ternary relations the operation of rotation,
which is not needed for binary relations.
For any two
ternary relations $R$ and $S$, $R\cap S$ is the intersection of $R$ and $S$,
$R\cup S$ is the union of $R$ and $S$, $R\circ S$ is the composition of $R$ and $S$, 
$R^\smile$ is the converse of $R$, and $R^\frown$ is the rotation of $R$:
\begin{center}
\begin{eqnarray}
R\cap S      &=&\{(a,b,c):(a,b,c)\in R\mbox{ and }(a,b,c)\in S\} \label{toone}\\
R\cup S      &=&\{(a,b,c):(a,b,c)\in R\mbox{ or }(a,b,c)\in S\} \label{totwo} \\
R\circ S     &=&\{(a,b,c):\mbox{for some }d,(a,b,d)\in R\mbox{ and }(a,d,c)\in S\} \label{tothree} \\
R^\smile     &=&\{(a,b,c):(a,c,b)\in R\} \label{tofour} \\
R^\frown     &=&\{(a,b,c):(c,a,b)\in R\} \label{tofive}
\end{eqnarray}
\end{center}
Three special ternary relations over a universe $U$ are the empty relation
$\emptyset$ which contains no triples at all, the identity relation
$\tirel =\{(a,a,a):a\in U\}$, and the universal relation
$\top _U^t=U\times U\times U$.
\subsection{Constraint matrices}
A ternary constraint matrix of order $n$ over $U$ is an
$n\times n\times n$-matrix, say $\termat$, of ternary relations over $U$ verifying the
following:
\begin{eqnarray}
(\forall i\leq n)     &(\termat _{iii}\subseteq\tirel )           &\mbox{(the identity property)} \nonumber \\
(\forall i,j,k\leq n) &(\termat _{ijk}=(\termat _{ikj})^\smile )        &\mbox{(the converse property)} \nonumber \\
(\forall i,j,k\leq n) &(\termat _{ijk}=(\termat _{kij})^\frown )        &\mbox{(the rotation property)} \nonumber
\end{eqnarray}
Let $P$ be a ternary CSP of order $n$ over a universe $U$.
Without loss of generality, we can make the assumption that for any
three variables $x_i,x_j,x_k$, there is at most one
constraint involving them.
$P$ can be associated with the following
ternary constraint matrix, denoted $\termat ^P$:
\begin{enumerate}
  \item initialise all entries to the universal relation:
    $(\forall i,j,k\leq n)((\termat ^P)_{ijk}\leftarrow \top _U^t)$;
  \item initialise the diagonal elements to the identity relation:
    $(\forall i\leq n)((\termat ^P)_{iii}\leftarrow\tirel )$; and
  \item for all triples $(x_i,x_j,x_k)$ of variables on which
    a constraint $(x_i,x_j,x_k)\in R$ is specified:\\
    $
    \begin{array}{lll}
    (\termat ^P)_{ijk}\leftarrow (\termat ^P)_{ijk}\cap R,     &(\termat ^P)_{ikj}\leftarrow ((\termat ^P)_{ijk})^\smile ,
                                                               &(\termat ^P)_{jki}\leftarrow ((\termat ^P)_{ijk})^\frown ,\\
    (\termat ^P)_{jik}\leftarrow ((\termat ^P)_{jki})^\smile , &(\termat ^P)_{kij}\leftarrow ((\termat ^P)_{jki})^\frown ,
                                                               &(\termat ^P)_{kji}\leftarrow ((\termat ^P)_{kij})^\smile.
    \end{array}
    $\\
\end{enumerate}
We make the assumption that, unless explicitly specified otherwise, a CSP is
given as a constraint matrix.
\subsection{(Strong) $k$-consistency, refinement}
Let $P$ be a ternary CSP of order $n$, $V$ its set of variables and $U$ its universe.
An instantiation of $P$ is any \mbox{$n$-tuple} $(a_1,a_2,\ldots ,a_n)$ of $U^n$,
representing an assignment of a value to each variable.  A consistent instantiation, or solution,
of $P$ is an instantiation satisfying all the constraints.
$P$ is consistent if it has at least one solution; it is inconsistent otherwise. The
consistency problem of $P$ is the problem of verifying whether $P$ is consistent.

Let $V'=\{x_{i_1},\ldots ,x_{i_j}\}$ be a subset of $V$. The sub-CSP of $P$ generated
by $V'$, denoted $P_{|V'}$, is the CSP with set of variables $V'$ and whose constraint
matrix is obtained by projecting the constraint matrix of $P$ onto $V'$.
$P$ is $k$-consistent \cite{Freuder82a} if for any subset $V'$ of $V$
containing $k-1$ variables, and for any variable $X\in V$, every solution to
$P_{|V'}$ can be extended to a solution to $P_{|V'\cup\{X\}}$. $P$ is strongly
$k$-consistent if it is $j$-consistent, for all $j\leq k$.
$1$-consistency, $2$-consistency and $3$-consistency correspond to node-consistency,
arc-consistency and path-consistency, respectively \cite{Mackworth77a,Montanari74a}.
Strong $n$-consistency of $P$ corresponds to what is called global consistency in
\cite{Dechter92a}. Global consistency facilitates the important task of searching
for a solution, which can be done, when the property is met, without backtracking
\cite{Freuder82a}.

A refinement of $P$ is a CSP $P'$ with the same set of variables, and such that
$(\forall i,j,k)((\termat ^{P'})_{ijk}\subseteq (\termat ^P)_{ijk})$.
\subsection{Relation algebras}\label{ras}
We recall some basic notions on relation algebras (RAs). For more details,
the reader is referred to \cite{Tarski41b,Duentsch99a,Duentsch01a,Ladkin94a}
for binary RAs, as first introduced by Tarski \cite{Tarski41b}, who was mainly
interested in formalising the theory of binary relations; and to
\cite{Isli00b} for ternary RAs, motivated by the authors with the fact that
binary RAs are not sufficient for the representation of spatial knowledge, such
as cyclic ordering of three points of the plane, known to be of primary
importance for applications such as robot localisation (how to represent the
konwledge that,
seen from the robot's position, three landmarks, say $L_1$, $L_2$ and $L_3$, are
met in that order, when we scan, say in the anticlockwise direction, a circle
centred at the robot's position, starting from $L_1$) ---cyclic ordering can
be looked at as the cyclic time counterpart of linear time betweenness.\footnote{The
work in \cite{Isli01c} shows how to solve a CSP of cyclic time intervals
\cite{Balbiani00a,Hornsby99a} using results on cyclic ordering of 2D orientations
\cite{Isli98a,Isli00b}, which emphasises the link between cyclic time and
cyclic ordering of 2D orientations.}

A Boolean algebra (BA) with universe ${\cal A}$ is an algebra of the form
$\langle {\cal A},\oplus ,\odot,^-,\bot ,\top\rangle$ which satisfies the following properties,
for all $R,S,T\in {\cal A}$:
\begin{eqnarray}
R\oplus (S\oplus T) &=&(R\oplus S)\oplus T \\
R\oplus S           &=&S\oplus R \\
R\odot S\oplus R    &=&R \\
R\odot S\oplus T    &=&(R\oplus T)\odot (S\oplus T) \\
R\oplus\overline{R} &=&\top
\end{eqnarray}
$\rel-alg$ is a binary RA with universe ${\cal A}$ \cite{Tarski41b,Ladkin94a} if:
\begin{enumerate}
  \item ${\cal A}$ is a set of binary relations; and
  \item $\rel-alg =\langle {\cal A},\oplus ,\odot,^-,\bot ,\top ,\circ ,^\smile ,{\cal I}\rangle$,
    where $\langle {\cal A},\oplus ,\odot,^-,\bot ,\top\rangle$ is a BA
    (called the Boolean part, or reduct, of $\rel-alg$), $\circ$ is a binary
    operation, $^\smile$ is a unary operation, ${\cal I}\in {\cal A}$, and the
    following identities hold for all $R,S,T\in {\cal A}$:
      \begin{eqnarray}
        (R\circ S)\circ T                      &=&R\circ (S\circ T)  \\
        (R\oplus S)\circ T                     &=&R\circ T\oplus S\circ T  \\
        R\circ {\cal I}                        &=&{\cal I}\circ R=R  \\
        (R^\smile )^\smile                     &=&R  \\
        (R\oplus S)^\smile                     &=&R^\smile\oplus S^\smile  \\
        (R\circ S)^\smile                      &=&S^\smile\circ R^\smile  \\
        R^\smile\circ\overline{R\circ S}\odot S&=&\bot
      \end{eqnarray}
\end{enumerate}

Ternary RAs \cite{Isli00b} need a (unary) operation called {\em rotation}, in
addition to an adaptation to the ternary case of the operations of
{\em composition} and {\em converse}, first introduced by De Morgan for
binary relations \cite{DeMorgan66a}. $\rel-alg$ is a ternary RA with universe
${\cal A}$ \cite{Isli00b} if:
\begin{enumerate}
  \item ${\cal A}$ is a set of ternary relations; and
  \item $\rel-alg =\langle {\cal A},\oplus ,\odot,^-,\bot ,\top ,\circ ,^\smile ,^\frown ,{\cal I}\rangle$
    where $\langle {\cal A},\oplus ,\odot,^-,\bot ,\top\rangle$ is a BA
    (called the Boolean part, or reduct, of $\rel-alg$), $\circ$ is a binary
    operation, $^\smile$ and $^\frown$ are unary operations, ${\cal I}\in {\cal A}$, and the
    following identities hold for all $R,S,T\in {\cal A}$:
      \begin{eqnarray}
        (R\circ S)\circ T                       &=&R\circ (S\circ T)\label{tra-pone} \\
        (R\oplus S)\circ T                      &=&R\circ T\oplus S\circ T\label{tra-ptwo} \\
        R\circ {\cal I}                         &=&{\cal I}\circ R         =R\label{tra-pthree} \\
        (R^\smile )^\smile                      &=&R\label{tra-pfour} \\
        (R\oplus S)^\smile                      &=&R^\smile\oplus S^\smile\label{tra-pfive} \\
        (R\circ S)^\smile                       &=&S^\smile\circ R^\smile\label{tra-psix} \\
        R^\smile\circ\overline{R\circ S}\odot S &=&\bot\label{tra-pseven} \\
        ((R^\frown )^\frown )^\frown            &=&R\label{tra-peight} \\
        (R\oplus S)^\frown                      &=&R^\frown\oplus S^\frown\label{tra-pnine}
      \end{eqnarray}
\end{enumerate}
Let $\rel-alg$ be an RA. The elements of $\rel-alg$ are just the relations in its universe.
An atom of $\rel-alg$ is a minimal nonzero element, i.e., $R$ is an atom
if $R\not =\bot$ and for every $S\in {\cal A}$, either
$R\odot S=\bot$ or $R\odot\overline{S}=\bot$. $\rel-alg$ is atomic
if every nonzero element has an atom below it; i.e., if for all nonzero
elements $R$, there exists an atom $A$ such that $A\odot R=A$.
$\rel-alg$ is finite if its universe has finitely many elements. A finite RA is
atomic, and its Boolean part is completely determined by
its atoms. Furthermore, in an atomic RA, the result of applying any of the
operations of the RA to any elements can be obtained from the results
of applying the different operations to the atoms. Specifying a finite,
thus atomic, RA reduces thus to specifying the identity element and
the results of applying the different operations to the different atoms.

The full binary (resp. ternary) RA over a set $U$ is the RA
$\fbra _U=\langle\binrel (U),\cup ,\cap ,$
$^-,\emptyset ,\top _U^b,\circ ,^\smile ,\birel\rangle$
(resp. $\ftra _U=\langle\terrel (U),\cup ,\cap ,^-,\emptyset ,\top _U^t,\circ ,^\smile ,^\frown ,\tirel\rangle$),
where:
\begin{enumerate}
  \item the universe $\binrel (U)$ (resp. $\terrel (U)$) is the set of all binary
    (resp. ternary) relations over
$U$;
  \item $\cup$, $\cap$ and $^-$ are, respectively, the usual set-theoretic
    operations of union, intersection and complement;
  \item $\emptyset$ is the empty relation;
  \item $\top _U^b$ (resp. $\top _U^t$) is the universal binary (resp.
    ternary) relation over $U$:
    $\top _U^b=U\times U$ (resp. $\top _U^t=U\times U\times U$);
  \item $\circ$ and $^\smile$ are, respectively, the operations of
    composition and converse of binary (resp. ternary) relations;
  \item $^\frown$ is the operation of rotation of ternary relations; and
  \item $\birel$ (resp. $\tirel$) is the binary (resp.
    ternary) identity relation over $U$:
    $\birel =\{(a,a)|a\in U\}$ (resp. $\tirel =\{(a,a,a)|a\in U\}$).
\end{enumerate}
A binary (resp. ternary) RA over a set $U$ is an RA
$\rel-alg =\langle {\cal A},\cup ,\cap ,^-,\emptyset ,\top _U^b,\circ ,^\smile ,\birel\rangle$
(resp. $\rel-alg =\langle {\cal A},\cup ,\cap ,^-,\emptyset ,\top _U^t,\circ ,^\smile ,^\frown ,\tirel\rangle$),
with universe ${\cal A}\subseteq\binrel (U)$ (resp.
${\cal A}\subseteq\terrel (U)$), such that:
\begin{enumerate}
  \item ${\cal A}$ is closed under the distinguished operations of
    $\binrel (U)$ (resp. $\terrel (U)$), namely, under the operations $\cup$, $\cap$, $^-$,
    $\circ$ and $^\smile$ (resp. the operations $\cup$, $\cap$, $^-$, $\circ$,
    $^\smile$ and $^\frown$); and
  \item ${\cal A}$ contains the distinguished constants, namely, the relations
    $\emptyset$, $\top _U^b$ and $\birel$ (resp. the relations $\emptyset$,
    $\top _U^t$ and $\tirel$).
\end{enumerate}
Such a binary (resp. ternary) RA is a subalgebra of the full RA  $\fbra _U$ (resp. $\ftra _U$).

Let $\{R_i:i\in I\}\subseteq\binrel (U)$ (resp. $\{R_i:i\in I\}\subseteq\terrel (U)$).
The binary (resp. ternary) RA generated by $\{R_i:i\in I\}$, denoted by
$\langle R_i:i\in I\rangle$, is the RA
$\langle {\cal A},\cup ,\cap ,^-,\emptyset ,\top _U^b,\circ ,^\smile ,\birel\rangle$
(resp. $\langle {\cal A},\cup ,\cap ,^-,\emptyset ,\top _U^t,\circ ,^\smile ,^\frown ,\tirel\rangle$),
such that ${\cal A}$ is the smallest subset of $\binrel (U)$ (resp. $\terrel (U)$)
closed under the distinguished operations of $\binrel (U)$ (resp. $\terrel (U)$). We refer to $\{R_i:i\in I\}$
as a base of $\langle R_i:i\in I\rangle$.

Of particular interest to this work are atomic, finite ternary RAs over a set $U$, of
the form
$\langle 2^{A},\cup ,\cap ,^-,\emptyset ,\top _U^t,\circ ,^\smile ,^\frown ,\tirel\rangle$,
where $A$ is a nonempty finite set of atoms that are Jointly Exhaustive and Pairwise Disjoint
(JEPD): for all triples $(x,y,z)\in U^3$, there exists one and only one atom $t$ from $A$ such that
$t(x,y,z)$. Such a set $A$ of atoms correponds to the finite partitioning,
$\displaystyle\bigcup _{t\in A}t$, of the universal ternary relation over $U$, $\top _U^t$. Such an RA is
nothing else than the RA $\langle t: t\in A\rangle$ generated by $A$. The universe $U$ will be, unless otherwise specified, the
set $\dlines$ of 2D directed lines.

Throughout the rest of the paper, given an $n$-ary algebra $\rel-alg$, with atoms
$r_1,\ldots ,r_m$, and universe $U$, we shall use the notation $\algats$ to refer
to the set $\{r_1,\ldots ,r_m\}$ of all atoms; an $\rel-alg$ relation, say $R$,
is any subset of $\algats$, interpreted as follows:
\begin{eqnarray}
(\forall x_1,\ldots ,x_n\in U)(R(x_1,\ldots ,x_n)\Leftrightarrow\displaystyle\bigvee _{r\in R}r(x_1,\ldots ,x_n))\nonumber
\end{eqnarray}
An $\rel-alg$ atomic relation is an $\rel-alg$ relation consisting of one single
atom (singleton set).
\section{Isli and Cohn's ternary RA of 2D orientations}\label{icra}
We use $\BBR ^2$ as a model of the plane, and assume that $\BBR ^2$ is
associated with a Cartesian coordinate system $(x,O,y)$.
We refer to the set of 2D orientations as $\deuxdo$; to the circle centred
at $O$ and of unit radius, as $\co1$; to the set of directed lines of the
plane as $\dlines$; to the set of undirected lines of the plane as
$\ulines$; to the union $\dlines\cup\ulines$ as $\lines$; and to the set,
subset of $\dlines$, of directed lines of the plane containing (incident
with) $O$, as $\dlo$. Throughout the rest of the paper, we use $\dligne$ and $\uligne$
as abbreviations for ``directed line'' and ``undirected line'', respectively. Given
two distinct points $x$ and $y$ of the plane $\BBR ^2$, we denote by
$\overrightarrow{xy}$ the $\dligne$ containing $x$ and $y$ and oriented
from $x$ to $y$;
given a set $A$, $|A|$ denotes the cardinality (i.e., the number of
elements) of $A$;
given $\ell\in\dlines$, ${\cal O}(\ell )$ refers to the orientation of
$\ell$;
given $\ell\in\lines$, $\points (\ell )$ refers to the set of points of
the plane belonging to $\ell$.

It is common in geometry to consider a line as a set of points, so that
one can write, for a line $\ell$, that $\ell =\points (\ell )$; this is
possible as long as we are concerned only with $\ulignes$, i.e.,
with the set $\ulines$; when the space in consideration is $\dlines$, or
its superset $\lines$, this is not possible any longer, for
$\points (\ell )$ does not contain the information of whether $\ell$ is
a $\dligne$ or a $\uligne$.
\begin{df}\label{isos}
The isomorphisms $\isodeux$ and $\isotrois$ are defined as follows:
\begin{enumerate}
  \item $\isodeux :\deuxdo\rightarrow\coone$; $\isodeux (z)$ is the point $P_z\in\coone$ such
    that the orientation of the $\dligne$ $\overrightarrow{OP_z}$ is $z$.
  \item $\isotrois :\deuxdo\rightarrow\dlo$; $\isotrois (z)$ is the line $\ell _{O,z}\in\dlo$
    of orientation $z$.
\end{enumerate}
\end{df}
\begin{df}\label{angledeuxdlignes}
The angle determined by
two $\dlignes$ $D_1$ and $D_2$,
denoted $\angleg D_1,D_2\angled$, is the one corresponding to the move in an anticlockwise
direction from $D_1$ to $D_2$. The angle
$\angleg z_1,z_2\angled$ determined by orientations $z_1$ and $z_2$ is the angle
$\angleg \isotrois (z_1),\isotrois (z_2)\angled$.
\end{df}
The set $\deuxdo$
can thus be viewed as the set of points of $\coone$ (or of any fixed circle), or as
the set of $\dlignes$ containing $O$ (or any fixed point).
Isli and Cohn \cite{Isli98a,Isli00b} have defined
a binary RA of 2D orientations, $\apra$, that contains four atoms: $e$ (equal), $l$
(left), $o$ (opposite) and $r$ (right). For all $x,y\in\deuxdo$:
\begin{eqnarray}
e(y,x) &\Leftrightarrow &\angleg x,y\angled =0 \nonumber \\
l(y,x) &\Leftrightarrow &\angleg x,y\angled\in (0,\pi ) \nonumber \\
o(y,x) &\Leftrightarrow &\angleg x,y\angled =\pi \nonumber \\
r(y,x) &\Leftrightarrow &\angleg x,y\angled\in (\pi ,2\pi ) \nonumber
\end{eqnarray}
Based on $\apra$, Isli and Cohn \cite{Isli98a,Isli00b} have
built a ternary RA, $\atra$, for cyclic ordering of 2D
orientations: $\atra$ has $24$ atoms, thus $2^{24}$
relations.
The atoms of $\atra$ are written as $b_1b_2b_3$, where
$b_1,b_2,b_3$ are atoms of $\apra$, and such an atom
is interpreted as follows:
$(\forall x,y,z\in\deuxdo )(b_1b_2b_3(x,y,z)\Leftrightarrow b_1(y,x)
 \wedge b_2(z,y)\wedge b_3(z,x))$.
\begin{figure}
\begin{center}
$
\begin{array}{|l|l|l|}  \hline
t       &t^\smile       &t^\frown       \\  \hline\hline
eee     &eee            &eee            \\  \hline
ell     &lre            &lre            \\  \hline
eoo     &ooe            &ooe            \\  \hline
err     &rle            &rle            \\  \hline
lel     &lel            &err            \\  \hline
lll     &lrl            &lrr            \\  \hline
llo     &orl            &lor            \\  \hline
llr     &rrl            &llr            \\  \hline
\end{array}
\hskip 0.3cm
\begin{array}{|l|l|l|}  \hline
t       &t^\smile       &t^\frown       \\  \hline\hline
lor     &rol            &olr            \\  \hline
lre     &ell            &rer            \\  \hline
lrl     &lll            &rrr            \\  \hline
lrr     &rll            &rlr            \\  \hline
oeo     &oeo            &eoo            \\  \hline
olr     &rro            &llo            \\  \hline
ooe     &eoo            &oeo            \\  \hline
orl     &llo            &rro            \\  \hline
\end{array}
\hskip 0.3cm
\begin{array}{|l|l|l|}  \hline
t       &t^\smile       &t^\frown       \\  \hline\hline
rer     &rer            &ell            \\  \hline
rle     &err            &lel            \\  \hline
rll     &lrr            &lrl            \\  \hline
rlr     &rrr            &lll            \\  \hline
rol     &lor            &orl            \\  \hline
rrl     &llr            &rrl            \\  \hline
rro     &olr            &rol            \\  \hline
rrr     &rlr            &rll            \\  \hline
\end{array}
$
\end{center}
\caption{The converse $t^\smile$ and the rotation
         $t^\frown$ of a $\atra$ atom $t$.}\label{con-rot-atra}
\end{figure}
\begin{figure}[t]
\centerline{\epsffile{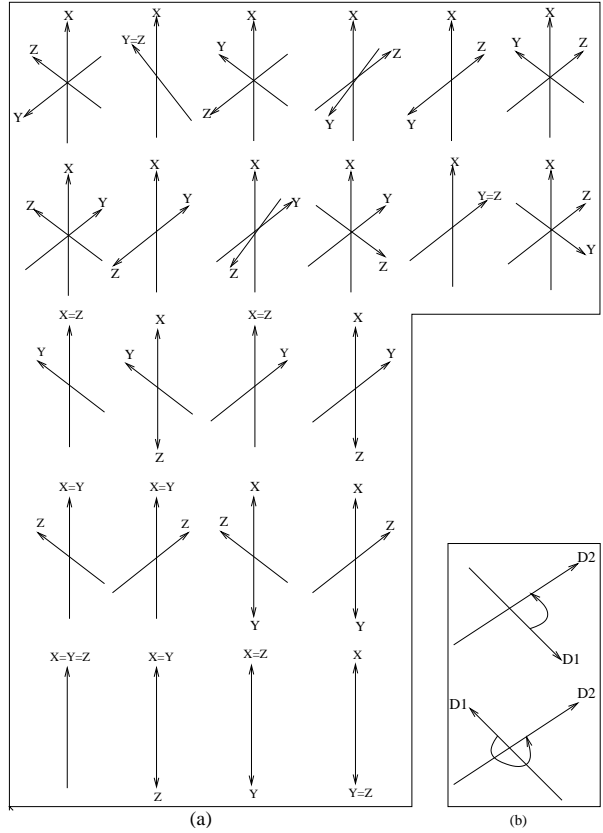}}
\caption{(a) Graphical illustration of the $24$ $\atra$ atoms:
        from top to bottom, left to right, the atoms are
        $\lrl ,\lel ,\lll ,\llr ,\lor ,\lrr ,
         \rll ,\rol ,\rrl ,\rrr ,\rer ,\rlr ,
         \lre ,\llo ,\rle ,\rro ,$
        $\elll ,\err ,\orl ,\olr ,
         \eee ,\eoo ,\ooe ,\oeo$;
        (b) The angle $\langle D_1,D_2\rangle$ determined by two $\dlignes$
        $D_1$ and $D_2$ is the one corresponding to the move in an anticlockwise
        direction from $D_1$ to $D_2$.}\label{CycordsRelation}
\end{figure}
Figure \ref{con-rot-atra} reproduces the $\atra$ converse and rotation
table. Figure \ref{CycordsRelation}
illustrates the $24$ $\atra$ atoms, and the angle determined by two $\dlignes$.
The reader is referred to \cite{Isli98a,Isli00b} for the $\atra$ composition
tables.
\section{The algebra of $\dlignes$}\label{dlalg}
We define in this section our algebra of ternary relations on $\dlignes$. The knowlege the
algebra can express, consists of a combination of translational knowledge and rotational knowledge.
The translational component records as ternary relations knowledge such as, the order in which two
$\dlignes$ cut a third one, or the order in which come three parallel $\dlignes$, when we move
from the left half-plane towards the right half-plane bounded by one of the $\dlignes$.

The rotational component, on the other hand,
records, also as ternary relations, knowledge on the relative angles of the three $\dligne$ arguments;
specifically, on the angles determined by pairs of the three arguments.
\subsection{The translational component}
We start by defining three binary relations, $\cuts$, $\coincides$ (coincides with) and
$\parallel$ (strictly parallel to), over the set $\dlines$ of $\dlignes$, and the
derived relation $\parallelity$ (parallel to) of parallelity. For all
$x,y\in\dlines$:
\begin{eqnarray}
\cuts (x,y)         &\Leftrightarrow     &|\points (x)\cap\points (y)|=1 \\
\coincides (x,y)    &\Leftrightarrow     &\points (x)=\points (y) \\
\parallel (x,y)     &\Leftrightarrow     &\points (x)\cap\points (y)=\emptyset \\
\parallelity (x,y)  &\Leftrightarrow     &\coincides (x,y)\vee\parallel (x,y)\label{parallelity}
\end{eqnarray}
The first three relations are symmetric, in the sense that for all
$r\in\{\cuts ,\coincides ,$ $\parallel\}$, and for all $x,y\in\dlines$, if
$r(x,y)$ then $r(y,x)$. They define a partition of $\dlines\times\dlines$; in other
words, using a terminology now common in Qualitative Spatial Reasoning (QSR), the three
relations $\cuts$, $\coincides$ and $\parallel$ are Jointly Exhaustive and Pairwise Disjoint
(JEPD).

We use the relations $\cuts$ and $\parallelity$ to define four ternary relations,
$\cc$, $\cp$, $\pc$ and $\pp$, over $\dlines$. For all $x,y,z\in\dlines$:
\begin{eqnarray}
\cc (x,y,z) &\Leftrightarrow &\cuts (y,x)\wedge\cuts (z,x) \nonumber \\
\cp (x,y,z) &\Leftrightarrow &\cuts (y,x)\wedge\parallelity (z,x) \nonumber \\
\pc (x,y,z) &\Leftrightarrow &\parallelity (y,x)\wedge\cuts (z,x) \nonumber \\
\pp (x,y,z) &\Leftrightarrow &\parallelity (y,x)\wedge\parallelity (z,x) \nonumber
\end{eqnarray}
The relations $\cp$ and $\pc$ are the converses of each other: $\cp ^\smile =\pc$ and
$\pc ^\smile =\cp$; each of the other two relations, $\cc$ and $\pp$, is its own converse:
$\cc ^\smile =\cc$ and $\pp ^\smile =\pp$.
The relations $\cc$, $\cp$, $\pc$
and $\pp$ provide for each of their last two arguments the knowledge of whether it cuts, or is
parallel to, the first argument.

In order for the translational component of our algebra to be
expressively interesting, we want it to express as well knowledge such as the following:
\begin{enumerate}
  \item when the last two arguments both cut the first, which of them comes first
    when we walk along the first argument heading the positive direction;
  \item when one of the last two arguments is parallel to the first, which side of the
    first argument (the left half-plane, the $\dligne$ itself, or the right half-plane) it
    belongs to; and
  \item when all three arguments are parallel to each other, in what order do they appear
    when we walk perpendicularly to, from the left half-plane and
    heading towards the right half-plane bounded by, the first argument.
\end{enumerate}
\begin{df}\label{iesdlines}
Let $\ell\in\dlines$. The relations $<_{\ell }$, $=_{\ell }$ and $>_{\ell }$ are defined as follows.
For all $x,y\in\BBR ^2$:
\begin{eqnarray}
x<_{\ell}y &\Leftrightarrow &x\in\points (\ell )\wedge
                                                                          y\in\points (\ell )\wedge
                                                                          x\not = y\wedge
                                                                          \angleg\ell ,\overrightarrow{xy}\angled =0 \nonumber \\
x=_{\ell}y &\Leftrightarrow &x\in\points (\ell )\wedge
                                                                          y\in\points (\ell )\wedge
                                                                          x= y \nonumber \\
x>_{\ell}y &\Leftrightarrow &y<_{\ell}x \nonumber
\end{eqnarray}
\end{df}
Readers familiar with Vilain and Kautz's \cite{Vilain86a} linear time point algebra, $\ltpa$, can
easily notice a similarity between the relations in Definition \ref{iesdlines}, $<_{\ell}$, $=_{\ell}$ and $>_{\ell}$, and the
$\ltpa$ atoms, $<$, $=$ and $>$; the latter uses the time line as the reference directed
line, which, because it is a global reference line, does not need to appear as a subscript in the
relations. As argued in Appendix \ref{appendixb}, the fact that $\dlalg$ is an RA is a direct consequence
of the conjunction of the two facts that (1) $\ltpa$ is an RA \cite{Ladkin94a}, and (2) $\atra$ is an RA \cite{Isli00b}.

\begin{figure*}
\begin{center}
\setlength{\unitlength}{4144sp}%
\begingroup\makeatletter\ifx\SetFigFont\undefined%
\gdef\SetFigFont#1#2#3#4#5{%
  \reset@font\fontsize{#1}{#2pt}%
  \fontfamily{#3}\fontseries{#4}\fontshape{#5}%
  \selectfont}%
\fi\endgroup%
\begin{picture}(5367,5682)(271,-5281)
\thinlines
\special{ps: gsave 0 0 0 setrgbcolor}\put(4501,-4831){\vector( 0, 1){1575}}
\special{ps: grestore}\special{ps: gsave 0 0 0 setrgbcolor}\put(4051,-5056){\framebox(1575,2385){}}
\special{ps: gsave 0 0 0 setrgbcolor}\put(5176,-4831){\line( 0, 1){1575}}
\special{ps: grestore}\special{ps: gsave 0 0 0 setrgbcolor}\put(3151,-4831){\vector( 0, 1){1575}}
\special{ps: grestore}\special{ps: gsave 0 0 0 setrgbcolor}\put(2701,-5056){\framebox(900,2385){}}
\special{ps: gsave 0 0 0 setrgbcolor}\put(1801,-4831){\vector( 0, 1){1575}}
\special{ps: grestore}\special{ps: gsave 0 0 0 setrgbcolor}\put(676,-5056){\framebox(1575,2385){}}
\special{ps: gsave 0 0 0 setrgbcolor}\put(1126,-4831){\line( 0, 1){1575}}
\special{ps: grestore}\put(4141,-4381){\makebox(0,0)[lb]{\smash{\SetFigFont{17}{20.4}{\rmdefault}{\mddefault}{\updefault}\special{ps: gsave 0 0 0 setrgbcolor}0\special{ps: grestore}}}}
\put(4456,-4381){\makebox(0,0)[lb]{\smash{\SetFigFont{17}{20.4}{\rmdefault}{\mddefault}{\updefault}\special{ps: gsave 0 0 0 setrgbcolor}1\special{ps: grestore}}}}
\put(4771,-4381){\makebox(0,0)[lb]{\smash{\SetFigFont{17}{20.4}{\rmdefault}{\mddefault}{\updefault}\special{ps: gsave 0 0 0 setrgbcolor}2\special{ps: grestore}}}}
\put(5131,-4381){\makebox(0,0)[lb]{\smash{\SetFigFont{17}{20.4}{\rmdefault}{\mddefault}{\updefault}\special{ps: gsave 0 0 0 setrgbcolor}3\special{ps: grestore}}}}
\put(5446,-4381){\makebox(0,0)[lb]{\smash{\SetFigFont{17}{20.4}{\rmdefault}{\mddefault}{\updefault}\special{ps: gsave 0 0 0 setrgbcolor}4\special{ps: grestore}}}}
\put(4391,-3211){\makebox(0,0)[lb]{\smash{\SetFigFont{17}{20.4}{\rmdefault}{\mddefault}{\updefault}\special{ps: gsave 0 0 0 setrgbcolor}$\ell _1$\special{ps: grestore}}}}
\put(5111,-3211){\makebox(0,0)[lb]{\smash{\SetFigFont{17}{20.4}{\rmdefault}{\mddefault}{\updefault}\special{ps: gsave 0 0 0 setrgbcolor}$\ell _2$\special{ps: grestore}}}}
\put(4836,-5281){\makebox(0,0)[lb]{\smash{\SetFigFont{12}{14.4}{\rmdefault}{\mddefault}{\updefault}\special{ps: gsave 0 0 0 setrgbcolor}$\mbox{(e)}$\special{ps: grestore}}}}
\put(2791,-4381){\makebox(0,0)[lb]{\smash{\SetFigFont{17}{20.4}{\rmdefault}{\mddefault}{\updefault}\special{ps: gsave 0 0 0 setrgbcolor}0\special{ps: grestore}}}}
\put(3106,-4381){\makebox(0,0)[lb]{\smash{\SetFigFont{17}{20.4}{\rmdefault}{\mddefault}{\updefault}\special{ps: gsave 0 0 0 setrgbcolor}1\special{ps: grestore}}}}
\put(3421,-4381){\makebox(0,0)[lb]{\smash{\SetFigFont{17}{20.4}{\rmdefault}{\mddefault}{\updefault}\special{ps: gsave 0 0 0 setrgbcolor}2\special{ps: grestore}}}}
\put(3041,-3211){\makebox(0,0)[lb]{\smash{\SetFigFont{17}{20.4}{\rmdefault}{\mddefault}{\updefault}\special{ps: gsave 0 0 0 setrgbcolor}$\ell _1$\special{ps: grestore}}}}
\put(3041,-2986){\makebox(0,0)[lb]{\smash{\SetFigFont{17}{20.4}{\rmdefault}{\mddefault}{\updefault}\special{ps: gsave 0 0 0 setrgbcolor}$\ell _2$\special{ps: grestore}}}}
\put(3171,-5281){\makebox(0,0)[lb]{\smash{\SetFigFont{12}{14.4}{\rmdefault}{\mddefault}{\updefault}\special{ps: gsave 0 0 0 setrgbcolor}$\mbox{(d)}$\special{ps: grestore}}}}
\put(766,-4381){\makebox(0,0)[lb]{\smash{\SetFigFont{17}{20.4}{\rmdefault}{\mddefault}{\updefault}\special{ps: gsave 0 0 0 setrgbcolor}0\special{ps: grestore}}}}
\put(1396,-4381){\makebox(0,0)[lb]{\smash{\SetFigFont{17}{20.4}{\rmdefault}{\mddefault}{\updefault}\special{ps: gsave 0 0 0 setrgbcolor}2\special{ps: grestore}}}}
\put(1756,-4381){\makebox(0,0)[lb]{\smash{\SetFigFont{17}{20.4}{\rmdefault}{\mddefault}{\updefault}\special{ps: gsave 0 0 0 setrgbcolor}3\special{ps: grestore}}}}
\put(2071,-4381){\makebox(0,0)[lb]{\smash{\SetFigFont{17}{20.4}{\rmdefault}{\mddefault}{\updefault}\special{ps: gsave 0 0 0 setrgbcolor}4\special{ps: grestore}}}}
\put(1081,-4381){\makebox(0,0)[lb]{\smash{\SetFigFont{17}{20.4}{\rmdefault}{\mddefault}{\updefault}\special{ps: gsave 0 0 0 setrgbcolor}1\special{ps: grestore}}}}
\put(1016,-3211){\makebox(0,0)[lb]{\smash{\SetFigFont{17}{20.4}{\rmdefault}{\mddefault}{\updefault}\special{ps: gsave 0 0 0 setrgbcolor}$\ell _2$\special{ps: grestore}}}}
\put(1691,-3211){\makebox(0,0)[lb]{\smash{\SetFigFont{17}{20.4}{\rmdefault}{\mddefault}{\updefault}\special{ps: gsave 0 0 0 setrgbcolor}$\ell _1$\special{ps: grestore}}}}
\put(1461,-5281){\makebox(0,0)[lb]{\smash{\SetFigFont{12}{14.4}{\rmdefault}{\mddefault}{\updefault}\special{ps: gsave 0 0 0 setrgbcolor}$\mbox{(c)}$\special{ps: grestore}}}}
\special{ps: gsave 0 0 0 setrgbcolor}\put(4051,-1996){\framebox(1350,2385){}}
\special{ps: gsave 0 0 0 setrgbcolor}\put(5176,-1771){\line(-3, 5){945}}
\special{ps: grestore}\special{ps: gsave 0 0 0 setrgbcolor}\put(4726,-1771){\vector( 0, 1){1575}}
\special{ps: grestore}\put(4616,-151){\makebox(0,0)[lb]{\smash{\SetFigFont{17}{20.4}{\rmdefault}{\mddefault}{\updefault}\special{ps: gsave 0 0 0 setrgbcolor}$\ell _1$\special{ps: grestore}}}}
\put(4746,-2221){\makebox(0,0)[lb]{\smash{\SetFigFont{12}{14.4}{\rmdefault}{\mddefault}{\updefault}\special{ps: gsave 0 0 0 setrgbcolor}$\mbox{(b)}$\special{ps: grestore}}}}
\put(4146,-106){\makebox(0,0)[lb]{\smash{\SetFigFont{17}{20.4}{\rmdefault}{\mddefault}{\updefault}\special{ps: gsave 0 0 0 setrgbcolor}$\ell _2$\special{ps: grestore}}}}
\put(4681,-1141){\makebox(0,0)[lb]{\smash{\SetFigFont{17}{20.4}{\rmdefault}{\mddefault}{\updefault}\special{ps: gsave 0 0 0 setrgbcolor}$=$\special{ps: grestore}}}}
\put(4681,-1546){\makebox(0,0)[lb]{\smash{\SetFigFont{17}{20.4}{\rmdefault}{\mddefault}{\updefault}\special{ps: gsave 0 0 0 setrgbcolor}$<$\special{ps: grestore}}}}
\put(4681,-781){\makebox(0,0)[lb]{\smash{\SetFigFont{17}{20.4}{\rmdefault}{\mddefault}{\updefault}\special{ps: gsave 0 0 0 setrgbcolor}$>$\special{ps: grestore}}}}
\special{ps: gsave 0 0 0 setrgbcolor}\put(676,-1996){\framebox(1350,2385){}}
\special{ps: gsave 0 0 0 setrgbcolor}\put(1351,-1771){\vector( 0, 1){1575}}
\special{ps: grestore}\put(1306,-1321){\makebox(0,0)[lb]{\smash{\SetFigFont{17}{20.4}{\rmdefault}{\mddefault}{\updefault}\special{ps: gsave 0 0 0 setrgbcolor}$c$\special{ps: grestore}}}}
\put(1621,-1321){\makebox(0,0)[lb]{\smash{\SetFigFont{17}{20.4}{\rmdefault}{\mddefault}{\updefault}\special{ps: gsave 0 0 0 setrgbcolor}$r$\special{ps: grestore}}}}
\put(1241,-151){\makebox(0,0)[lb]{\smash{\SetFigFont{17}{20.4}{\rmdefault}{\mddefault}{\updefault}\special{ps: gsave 0 0 0 setrgbcolor}$\ell$\special{ps: grestore}}}}
\put(1371,-2221){\makebox(0,0)[lb]{\smash{\SetFigFont{12}{14.4}{\rmdefault}{\mddefault}{\updefault}\special{ps: gsave 0 0 0 setrgbcolor}$\mbox{(a)}$\special{ps: grestore}}}}
\put(991,-1321){\makebox(0,0)[lb]{\smash{\SetFigFont{17}{20.4}{\rmdefault}{\mddefault}{\updefault}\special{ps: gsave 0 0 0 setrgbcolor}$l$\special{ps: grestore}}}}
\end{picture}
\end{center}
\caption{(a) The plane partition determined by a $\dligne$;
  (b) the line partition determined by a $\dligne$ $\ell _2$ on a $\dligne$ $\ell _1$;
  (c) the plane partition determined by two $\dlignes$ $\ell _1$ and $\ell _2$ verifying $\leftpar (\ell _2,\ell _1)$;
  (d) the plane partition determined by two $\dlignes$ $\ell _1$ and $\ell _2$ verifying $\coincides (\ell _2,\ell _1)$;
  (e) the plane partition determined by two $\dlignes$ $\ell _1$ and $\ell _2$ verifying $\rightpar(\ell _2,\ell _1)$.}\label{regions}
\end{figure*}
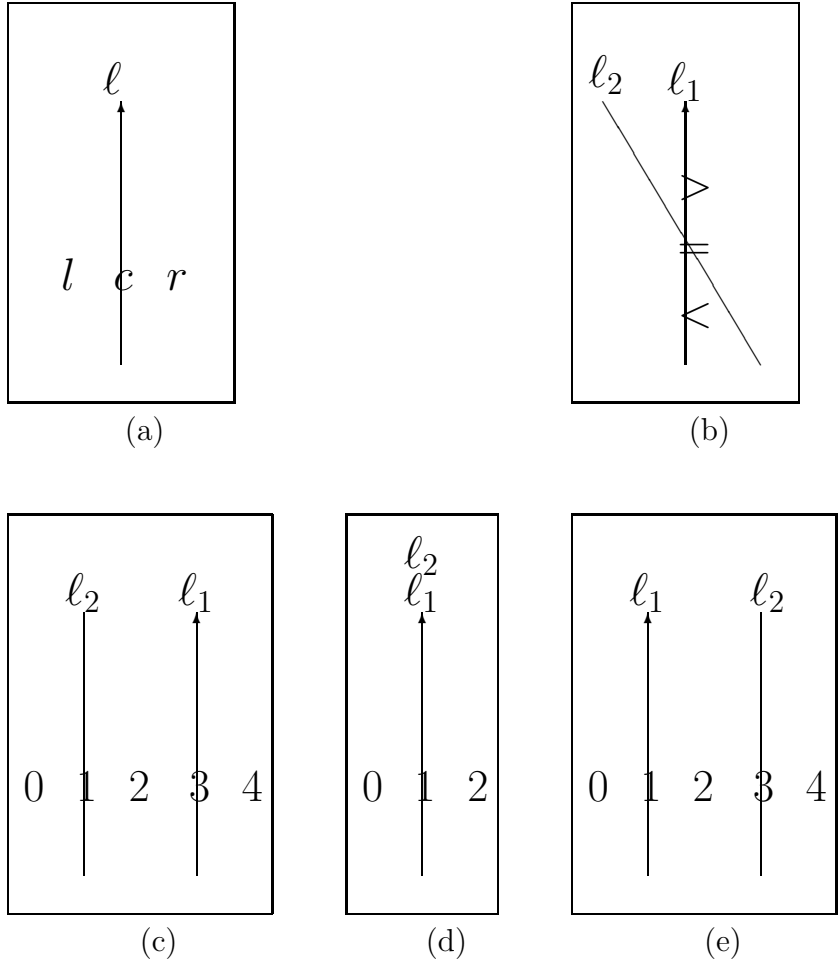

We make use of the relations $<_{\ell}$, $=_{\ell}$ and $>_{\ell}$ of Definition \ref{iesdlines} to refine the relation
$\cc$ into three relations $\cc _<$, $\cc _=$ and $\cc _>$, which add to the knowledge already
contained in $\cc$, the order in which the last two arguments are met in the walk along the
first argument heading the positive direction.
For all $\ell _1,\ell _2,\ell _3\in\dlines$:
\begin{eqnarray}
\cc _< (\ell _1,\ell _2,\ell _3 ) &\Leftrightarrow &\cc (\ell _1,\ell _2,\ell _3 )\wedge \nonumber \\
&&                                                                           (\forall x\in\points (\ell _2)\cap\points (\ell _1))
                                                                             (\forall y\in\points (\ell _3)\cap\points (\ell _1))
                                                                             (x<_{\ell _1}y) \nonumber \\
\cc _= (\ell _1,\ell _2,\ell _3 ) &\Leftrightarrow &\cc (\ell _1,\ell _2,\ell _3 )\wedge
                                                                             (\points (\ell _2)\cap\points (\ell _1)=
                                                                              \points (\ell _3)\cap\points (\ell _1)) \nonumber \\
\cc _> (\ell _1,\ell _2,\ell _3 ) &\Leftrightarrow &\cc _< (\ell _1,\ell _3,\ell _2) \nonumber
\end{eqnarray}
\begin{df}[plane partition determined by a $\dligne$]\label{ppartition-1l}
A $\dligne$ $\ell$ defines the obvious partition of the plane illustrated in Figure \ref{regions}(a). We refer
to the set of all regions of the partition as $\ppartition (\ell )$, and to each region in
$\ppartition (\ell )$ as $\ppregion _x(\ell )$, where $x$ is the label associated with the region in
Figure \ref{regions}(a).
\end{df}
Given a $\dligne$ $\ell$, we will also refer to $\ppregion _l(\ell )$, $\ppregion _c(\ell )$ and
$\ppregion _r(\ell )$ as $\lhp (\ell )$ (the open left half-plane bounded by $\ell $), $\points (\ell )$
(the set of points of $\ell$)
and $\rhp (\ell )$ (the open right half-plane bounded by $\ell $), respectively.

We now split the relation $\parallel$ into two obvious (finer) relations, $\leftpar$
($l$ for left) and $\rightpar$ ($r$ for right). For all $\ell ,\ell '\in\dlines$:
\begin{eqnarray}
\leftpar (\ell ',\ell ) &\Leftrightarrow &\parallel (\ell ',\ell )\wedge
                                                     (\forall x\in\points (\ell '))
                                                       (x\in\lhp (\ell )) \nonumber \\
\rightpar (\ell ',\ell ) &\Leftrightarrow &\parallel (\ell ',\ell )\wedge
                                                                             \neg\leftpar (\ell ',\ell ) \nonumber
\end{eqnarray}
In other words, we have the following, for all $\dlignes$
$\ell$ and $\ell '$:
\begin{eqnarray}
\leftpar (\ell ',\ell )   &\Leftrightarrow &\points (\ell ')\subset\lhp (\ell ) \nonumber \\
\coincides (\ell ',\ell ) &\Leftrightarrow &\points (\ell ')=\points (\ell ) \nonumber \\
\rightpar (\ell ',\ell )  &\Leftrightarrow &\points (\ell ')\subset\rhp (\ell ) \nonumber
\end{eqnarray}
Readers familiar with Vilain and Kautz's point algebra $\ltpa$ \cite{Vilain86a} can, again,
easily notice a similarity between the relations $\leftpar$, $\coincides$ and $\rightpar$, on the
one hand, and the $\ltpa$ atoms $<$, $=$ and $>$, on the other hand.

We make use of the relations $\leftpar$, $\coincides$ and $\rightpar$ to refine
the relation $\cp$ into three relations, $\cp _l$, $\cp _c$ and $\cp _r$;
the relation $\pc$ into three relations, $\pc _l$, $\pc _c$ and $\pc _r$;
and the relation $\pp$ into three relations, $\pp _l$, $\pp _c$ and $\pp _r$. For all
$\ell _1,\ell _2,\ell _3\in\dlines$:
\begin{eqnarray}
\cp _l (\ell _1,\ell _2,\ell _3 ) &\Leftrightarrow &\cp (\ell _1,\ell _2,\ell _3 )\wedge
                                                                             \leftpar (\ell _3,\ell _1) \nonumber \\
\cp _c (\ell _1,\ell _2,\ell _3 ) &\Leftrightarrow &\cp (\ell _1,\ell _2,\ell _3 )\wedge
                                                                             \coincides (\ell _3,\ell _1) \nonumber \\
\cp _r (\ell _1,\ell _2,\ell _3 ) &\Leftrightarrow &\cp (\ell _1,\ell _2,\ell _3 )\wedge
                                                                             \rightpar (\ell _3,\ell _1) \nonumber \\
\pc _l (\ell _1,\ell _2,\ell _3 ) &\Leftrightarrow &\pc (\ell _1,\ell _2,\ell _3 )\wedge
                                                                             \leftpar (\ell _2,\ell _1) \nonumber \\
\pc _c (\ell _1,\ell _2,\ell _3 ) &\Leftrightarrow &\pc (\ell _1,\ell _2,\ell _3 )\wedge
                                                                             \coincides (\ell _2,\ell _1) \nonumber \\
\pc _r (\ell _1,\ell _2,\ell _3 ) &\Leftrightarrow &\pc (\ell _1,\ell _2,\ell _3 )\wedge
                                                                             \rightpar (\ell _2,\ell _1) \nonumber \\
\pp _l (\ell _1,\ell _2,\ell _3 ) &\Leftrightarrow &\pp (\ell _1,\ell _2,\ell _3 )\wedge
                                                                             \leftpar (\ell _2,\ell _1) \nonumber \\
\pp _c (\ell _1,\ell _2,\ell _3 ) &\Leftrightarrow &\pp (\ell _1,\ell _2,\ell _3 )\wedge
                                                                             \coincides (\ell _2,\ell _1) \nonumber \\
\pp _r (\ell _1,\ell _2,\ell _3 ) &\Leftrightarrow &\pp (\ell _1,\ell _2,\ell _3 )\wedge
                                                                             \rightpar (\ell _2,\ell _1) \nonumber
\end{eqnarray}
Again, readers familiar with Vilain and Kautz's algebra $\ltpa$ \cite{Vilain86a} can
easily notice a similarity between the relations $\cp _l$, $\cp _c$ and $\cp _r$ and the
$\ltpa$ atoms, $<$, $=$ and $>$; between the relations $\pc _l$, $\pc _c$ and $\pc _r$ and the
$\ltpa$ atoms; and between the relations $\pp _l$, $\pp _c$ and $\pp _r$ and the
$\ltpa$ atoms.
\begin{df}[line partition]\label{lpartition}
Let $\ell _1$ and $\ell _2$ be two cutting $\dlignes$ ---i.e., such that $\cuts (\ell _1,\ell _2)$.
$\ell _2$ defines a partition of $\ell _1$ as illustrated in Figure \ref{regions}(b). The three
regions of the partition, labelled $<$, $=$ and $>$ in Figure \ref{regions}(b), correspond,
respectively, to the open left half-line bounded by the intersecting point of $\ell _1$ and
$\ell _2$, the intersecting point of $\ell _1$ and $\ell _2$, and the open right half-line bounded
by the intersecting point of $\ell _1$ and $\ell _2$. We refer to the set of all regions of the
partition as $\lpartition (\ell _1,\ell _2)$, and to each region in $\lpartition (\ell _1,\ell _2)$
as $\lpregion _x(\ell _1,\ell _2)$, where $x$ is the label associated with the region in Figure
\ref{regions}(b).
\end{df}
Using the line partition of Definition \ref{lpartition}, we have the following, for all $\dlignes$
$\ell _1$, $\ell _2$ and $\ell _3$ verifying $\cc (\ell _1,\ell _2,\ell _3)$:
$\cc _<(\ell _1,\ell _2,\ell _3 )$ iff $\points (\ell _2)\cap\points (\ell _1)\subset\lpregion _<(\ell _1,\ell _3)$;
$\cc _=(\ell _1,\ell _2,\ell _3 )$ iff $\points (\ell _2)\cap\points (\ell _1)=\lpregion _=(\ell _1,\ell _3)$; and
$\cc _>(\ell _1,\ell _2,\ell _3 )$ iff $\points (\ell _2)\cap\points (\ell _1)\subset\lpregion _>(\ell _1,\ell _3)$.
\begin{df}[plane partition determined by two parallel $\dlignes$]\label{ppartition}
Two parallel $\dlignes$ $\ell _1$ and $\ell _2$ define a partition of the plane as illustrated in
Figure \ref{regions}(c) for the case $\leftpar (\ell _2,\ell _1)$, in
Figure \ref{regions}(d) for the case $\coincides (\ell _2,\ell _1)$, and in
Figure \ref{regions}(e) for the case $\rightpar (\ell _2,\ell _1)$. Each region of the partition is
an open half-plane bounded by either $\ell _1$ or $\ell _2$, a line ($\ell _1$ or $\ell _2$), or
the intersection of two open half-planes bounded by $\ell _1$ and $\ell _2$. We refer to the set
of all regions of the partition as $\ppartition (\ell _1,\ell _2)$, and to each region in
$\ppartition (\ell _1,\ell _2)$ as $\ppregion _x(\ell _1,\ell _2)$, where $x$ is the
label associated with the region in Figures \ref{regions}(c-d-e).
\end{df}
The partition of the plane determined by two parallel $\dlignes$ ---Definition \ref{ppartition}--- is now used to
refine the relation $\pp _l$ into $\pp _{l0}$, $\pp _{l1}$, $\pp _{l2}$, $\pp _{l3}$ and $\pp _{l4}$;
the relation $\pp _c$ into $\pp _{c0}$, $\pp _{c1}$ and $\pp _{c2}$; and the relation $\pp _r$ into
$\pp _{r0}$, $\pp _{r1}$, $\pp _{r2}$, $\pp _{r3}$ and $\pp _{r4}$. For all
$\ell _1,\ell _2,\ell _3\in\dlines$:
\begin{eqnarray}
(\forall i\leq 4)(\pp _{li} (\ell _1,\ell _2,\ell _3 ) &\Leftrightarrow &\pp _l(\ell _1,\ell _2,\ell _3 )\wedge
                                                                             \points (\ell _3)\subseteq\ppregion _i(\ell _1,\ell _2))
                                                                             \nonumber \\
(\forall i\leq 2)(\pp _{ci} (\ell _1,\ell _2,\ell _3 ) &\Leftrightarrow &\pp _c(\ell _1,\ell _2,\ell _3 )\wedge
                                                                             \points (\ell _3)\subseteq\ppregion _i(\ell _1,\ell _2))
                                                                             \nonumber \\
(\forall i\leq 4)(\pp _{ri} (\ell _1,\ell _2,\ell _3 ) &\Leftrightarrow &\pp _r(\ell _1,\ell _2,\ell _3 )\wedge
                                                                             \points (\ell _3)\subseteq\ppregion _i(\ell _1,\ell _2))
                                                                             \nonumber
\end{eqnarray}
Readers familiar with Ligozat's $(p,q)$-relations \cite{Ligozat91a} can easily notice a
similarity between $(1,2)$-relations and the relations $\pp _{li},i\in\{0,\ldots ,4\}$, on
the one hand, and between $(1,2)$-relations and the relations
$\pp _{ri},i\in\{0,\ldots ,4\}$, on the other hand. Ligozat's $(1,2)$-relations are called
point-interval relations in \cite{Vilain82a}. Again, readers familiar with Vilain and Kautz's algebra $\ltpa$ \cite{Vilain86a} can
easily notice a similarity between the relations $\pp _{c0}$, $\pp _{c1}$ and $\pp _{c2}$ and the
$\ltpa$ atoms, $<$, $=$ and $>$.

From now on, we refer to the translational component as $\dltalg$ (Translational Algebra of ternary relations ---over
$\dlines$); to the set of all $\dltalg$ atoms as $\dltats$:
\begin{eqnarray}
\cc     &=&\{\cc _<,\cc _=,\cc _>\} \nonumber \\
\cp     &=&\{\cp _l,\cp _c,\cp _r\} \nonumber \\
\pc     &=&\{\pc _l,\pc _c,\pc _r\} \nonumber \\
\pp _l  &=&\{\pp _{l0},\pp _{l1},\pp _{l2},\pp _{l3},\pp _{l4}\} \nonumber \\
\pp _c  &=&\{\pp _{c0},\pp _{c1},\pp _{c2}\} \nonumber \\
\pp _r  &=&\{\pp _{r0},\pp _{r1},\pp _{r2},\pp _{r3},\pp _{r4}\} \nonumber \\
\pp     &=&\pp _l\cup\pp _c\cup\pp _r \nonumber \\
\dltats &=&\cc\cup
                  \cp\cup
                  \pc\cup
                  \pp\nonumber
\end{eqnarray}
{\bf The $\dltalg$ composition tables.}
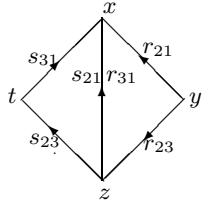
\begin{figure}[t]
\begin{center}
\setlength{\unitlength}{1500sp}%
\begingroup\makeatletter\ifx\SetFigFont\undefined%
\gdef\SetFigFont#1#2#3#4#5{%
  \reset@font\fontsize{#1}{#2pt}%
  \fontfamily{#3}\fontseries{#4}\fontshape{#5}%
  \selectfont}%
\fi\endgroup%
\begin{picture}(3000,3255)(2176,-3361)
\thinlines
\special{ps: gsave 0 0 0 setrgbcolor}\put(4471,-2341){\vector(-1,-1){ 60}}
\special{ps: grestore}\special{ps: gsave 0 0 0 setrgbcolor}\put(4413,-1049){\vector(-4, 3){ 76.800}}
\special{ps: grestore}\special{ps: gsave 0 0 0 setrgbcolor}\put(2926,-2221){\vector(-1, 1){ 75}}
\special{ps: grestore}\special{ps: gsave 0 0 0 setrgbcolor}\put(3736,-1636){\vector( 0, 1){150}}
\special{ps: grestore}\special{ps: gsave 0 0 0 setrgbcolor}\put(2903,-1208){\vector( 1, 1){135}}
\special{ps: grestore}\put(2956,-2611){\makebox(2.6458,18.5208){\SetFigFont{5}{6}{\rmdefault}{\mddefault}{\updefault}.}}
\put(2431,-1741){\line( 1,-1){1335}}
\put(3736,-361){\line( 0,-1){2700}}
\put(3732,-365){\line( 1,-1){1342.500}}
\put(5101,-1711){\line(-1,-1){1350}}
\put(2398,-1723){\line( 1, 1){1342.500}}
\put(4421,-2551){\makebox(0,0)[lb]{\smash{\SetFigFont{9}{10.8}{\rmdefault}{\mddefault}{\updefault}$r_{23}$}}}
\put(2526,-2461){\makebox(0,0)[lb]{\smash{\SetFigFont{9}{10.8}{\rmdefault}{\mddefault}{\updefault}$s_{23}$}}}
\put(3676,-3361){\makebox(0,0)[lb]{\smash{\SetFigFont{9}{10.8}{\rmdefault}{\mddefault}{\updefault}$z$}}}
\put(3751,-286){\makebox(0,0)[lb]{\smash{\SetFigFont{9}{10.8}{\rmdefault}{\mddefault}{\updefault}$x$}}}
\put(4401,-961){\makebox(0,0)[lb]{\smash{\SetFigFont{9}{10.8}{\rmdefault}{\mddefault}{\updefault}$r_{21}$}}}
\put(3226,-1411){\makebox(0,0)[lb]{\smash{\SetFigFont{9}{10.8}{\rmdefault}{\mddefault}{\updefault}$s_{21}$}}}
\put(3801,-1411){\makebox(0,0)[lb]{\smash{\SetFigFont{9}{10.8}{\rmdefault}{\mddefault}{\updefault}$r_{31}$}}}
\put(5176,-1786){\makebox(0,0)[lb]{\smash{\SetFigFont{9}{10.8}{\rmdefault}{\mddefault}{\updefault}$y$}}}
\put(2176,-1786){\makebox(0,0)[lb]{\smash{\SetFigFont{9}{10.8}{\rmdefault}{\mddefault}{\updefault}$t$}}}
\put(2501,-1111){\makebox(0,0)[lb]{\smash{\SetFigFont{9}{10.8}{\rmdefault}{\mddefault}{\updefault}$s_{31}$}}}
\end{picture}
\end{center}
\caption{The conjunction $r(x,y,z)\wedge s(x,z,t)$ is inconsistent
if       $r_{31}\not = s_{21}$.}\label{TernaryComposition}
\end{figure}
Given four $\dlignes$ $x,y,z,t$ and two $\dltalg$ atoms $r$ and $s$,
the conjunction $r(x,y,z)\wedge s(x,z,t)$ is inconsistent if the most
specific binary relation, $r_{31}(z,x)$, implied by $r(x,y,z)$ on the
pair $(z,x)$, is different from the most specific binary relation,
$s_{21}(z,x)$, on the same pair $(z,x)$, implied by $s(x,z,t)$ (see
Figure \ref{TernaryComposition} for illustration).
Each of $r_{31}$ and $s_{21}$ can be either of the four binary
relations $\cuts$, $\leftpar$, $\coincides$ or $\rightpar$; these
four binary relations are Jointly Exhaustive and Pairwise Disjoint (JEPD),
which means that any two $\dlignes$ are related by one and only one of the
four relations. Stated otherwise, when $r_{31}\not = s_{21}$, we have
$r\circ s=\emptyset$. Thus composition splits into four composition tables,
corresponding to the following four cases:
\begin{enumerate}
  \item \underline{Case 1:} $r_{31}= s_{21}=\cuts$. This corresponds to
    $r\in\cuts _{31}$ and $s\in\cuts _{21}$, with
    $\cuts _{31}=\{\cc _<,\cc _=,\cc _>,\pc _l,\pc _c,\pc _r\}$ and
    $\cuts _{21}=\{\cc _<,\cc _=,\cc _>,\cp _l,\cp _c,\cp _r\}$;
  \item \underline{Case 2:} $r_{31}= s_{21}=\leftpar$. This corresponds to
    $r\in\leftpar _{31}$ and $s\in\leftpar _{21}$,
    with $\leftpar _{31}=\{\cp _l,
                 \pp _{l0},\pp _{l1},\pp _{l2},
                 \pp _{c0},
                 \pp _{r0}\}$ and
    $\leftpar _{21}=\{\pc _l,
                 \pp _{l0},\pp _{l1},\pp _{l2},\pp _{l3},\pp _{l4}\}$;
  \item \underline{Case 3:} $r_{31}= s_{21}=\coincides$. This corresponds to
    $r\in\coincides _{31}$ and $s\in\coincides _{21}$,
    with $\coincides _{31}=\{\cp _c,
                  \pp _{l3},
                  \pp _{c1},
                  \pp _{r1}\}$ and
    $\coincides _{21}=\{\pc _c,
                   \pp _{c0},\pp _{c1},\pp _{c2}\}$; and
  \item \underline{Case 4:} $r_{31}= s_{21}=\rightpar$. This corresponds to
    $r\in\rightpar _{31}$ and $s\in\rightpar _{21}$,
    with $\rightpar _{31}=\{\cp _r,
                  \pp _{l4},
                  \pp _{c2},
                  \pp _{r2},\pp _{r3},\pp _{r4}\}$ and
  $\rightpar _{21}=\{\pc _r,
                  \pp _{r0},\pp _{r1},\pp _{r2},\pp _{r3},\pp _{r4}\}$.
\end{enumerate}
Figure \ref{TernaryComp1} presents the four composition tables.
\footnote{Alternatively,
one could define one single composition table for $\dltalg$. Such a table would have
$22\times 22$ entries, most of which (i.e.,
$22\times 22-(6\times 6+6\times 6+4\times 4+6\times 6))$ would be the empty relation.}
\begin{figure*}
\begin{footnotesize}
\begin{center}
$
\begin{array}{|l|l|l|}  \hline
t        &t^\smile &t^\frown\\  \hline\hline
\cc _<   &\cc _>   &\{\cc _<,\cc _>,\pc _l,\pc _r\}\\  \hline
\cc _=   &\cc _=   &\{\cc _=,\pc _c\}\\  \hline
\cc _>   &\cc _<   &\{\cc _<,\cc _>,\pc _l,\pc _r\}\\  \hline
\cp _l   &\pc _l   &\{\cc _<,\cc _>\}\\  \hline
\cp _c   &\pc _c   &\{\cc _=\}\\  \hline
\cp _r   &\pc _r   &\{\cc _<,\cc _>\}\\  \hline
\pc _l   &\cp _l   &\{\cp _l,\cp _r\}\\  \hline
\pc _c   &\cp _c   &\{\cp _c\}\\  \hline
\pc _r   &\cp _r   &\{\cp _l,\cp _r\}\\  \hline
\pp _{l0}&\pp _{l2}&\{\pp _{l4},\pp _{r0}\}\\  \hline
\pp _{l1}&\pp _{l1}&\{\pp _{c0},\pp _{c2}\}\\  \hline
\end{array}
\hskip 0.2cm
\begin{array}{|l|l|l|}  \hline
t        &t^\smile  &t^\frown\\  \hline\hline
\pp _{l2}&\pp _{l0} &\{\pp _{l0},\pp _{r4}\}\\  \hline
\pp _{l3}&\pp _{c0} &\{\pp _{l1},\pp _{r3}\}\\  \hline
\pp _{l4}&\pp _{r0} &\{\pp _{l2},\pp _{r2}\}\\  \hline
\pp _{c0}&\pp _{l3} &\{\pp _{l3},\pp _{r1}\}\\  \hline
\pp _{c1}&\pp _{c1} &\{\pp _{c1}\}\\  \hline
\pp _{c2}&\pp _{r1} &\{\pp _{l3},\pp _{r1}\}\\  \hline
\pp _{r0}&\pp _{l4} &\{\pp _{l2},\pp _{r2}\}\\  \hline
\pp _{r1}&\pp _{c2} &\{\pp _{l1},\pp _{r3}\}\\  \hline
\pp _{r2}&\pp _{r4} &\{\pp _{l0},\pp _{r4}\}\\  \hline
\pp _{r3}&\pp _{r3} &\{\pp _{c0},\pp _{c2}\}\\  \hline
\pp _{r4}&\pp _{r2} &\{\pp _{l4},\pp _{r0}\}\\  \hline
\end{array}
$
\vskip 0.1cm
$
\begin{array}{|l||l|l|l|l|l|l|l|l|} \hline
\circ       &\cc _<    &\cc _=    &\cc _>
            &\cp _l    &\cp _c    &\cp _r    \\      \hline\hline
\cc _<      &\cc _<    &\cc _<    &\cc
            &\cp _l    &\cp _c    &\cp _r                 \\      \hline
\cc _=      &\cc _<    &\cc _=    &\cc _>
            &\cp _l    &\cp _c    &\cp _r                 \\      \hline
\cc _>      &\cc       &\cc _<    &\cc _>
            &\cp _l    &\cp _c    &\cp _r                 \\      \hline
\pc _l      &\pc _l    &\pc _l    &\pc _l
            &\pp _{ll}          &\pp _{l3}    &\pp _{l4}                 \\      \hline
\pc _c      &\pc _c    &\pc _c    &\pc _c
            &\pp _{c0}    &\pp _{c1}    &\pp _{c2}                 \\      \hline
\pc _r      &\pc _r    &\pc _r    &\pc _r
            &\pp _{r0}    &\pp _{r1}    &\pp _{rr}                 \\      \hline
\end{array}
\hskip 0.2cm
\begin{array}{|l||l|l|l|l|l|l|l|l|} \hline
\circ      &\pc _l    &\pp _{l0}    &\pp _{l1}    &\pp _{l2}    &\pp _{l3}    &\pp _{l4}    \\      \hline\hline
\cp _l     &\cc    &\cp _l    &\cp _l   
           &\cp _l    &\cp _c    &\cp _r                 \\      \hline
\pp _{l0}  &\pc _l    &\pp _{l0}    &\pp _{l0}
           &\pp _{ll}    &\pp _{l3}    &\pp _{l4}                 \\      \hline
\pp _{l1}  &\pc _l    &\pp _{l0}    &\pp _{l1}
           &\pp _{l2}    &\pp _{l3}    &\pp _{l4}                 \\      \hline
\pp _{l2}  &\pc _l    &\pp _{ll}      &\pp _{l2}
           &\pp _{l2}    &\pp _{l3}    &\pp _{l4}                 \\      \hline
\pp _{c0}  &\pc _c    &\pp _{c0}    &\pp _{c0}
               &\pp _{c0}    &\pp _{c1}    &\pp _{c2}                 \\      \hline
\pp _{r0}   &\pc _r    &\pp _{r0}    &\pp _{r0}
            &\pp _{r0}    &\pp _{r1}    &\pp _{rr}                 \\      \hline
\end{array}
$
\vskip 0.1cm
$
\begin{array}{|l||l|l|l|l|} \hline
\circ      &\pc _c       &\pp _{c0}                          &\pp _{c1}    &\pp _{c2}                               \\      \hline\hline
\cp _c     &\cc          &\cp _l                             &\cp _c             &\cp _r                            \\      \hline
\pp _{l3}  &\pc _l       &\pp _{ll}  &\pp _{l3}          &\pp _{l4}                         \\      \hline
\pp _{c1}  &\pc _c       &\pp _{c0}                          &\pp _{c1}          &\pp _{c2}                         \\      \hline
\pp _{r1}  &\pc _r       &\pp _{r0}                          &\pp _{r1}          &\pp _{rr} \\      \hline
\end{array}
\hskip 0.1cm
\begin{array}{|l||l|l|l|l|l|l|l|l|} \hline
\circ      &\pc _r    &\pp _{r0}                         &\pp _{r1} &\pp _{r2}
           &\pp _{r3} &\pp _{r4}  \\      \hline\hline
\cp _r     &\cc       &\cp _l                            &\cp _c    &\cp _r
           &\cp _r    &\cp _r  \\      \hline
\pp _{l4}  &\pc _l    &\pp _{ll} &\pp _{l3} &\pp _{l4}
           &\pp _{l4} &\pp _{l4}  \\      \hline
\pp _{c2}  &\pc _c    &\pp _{c0}                         &\pp _{c1} &\pp _{c2}
           &\pp _{c2} &\pp _{c2}  \\      \hline
\pp _{r2}  &\pc _r    &\pp _{r0}                         &\pp _{r1} &\pp _{r2}
           &\pp _{r2} &\pp _{rr}  \\      \hline
\pp _{r3}  &\pc _r    &\pp _{r0}                         &\pp _{r1} &\pp _{r2}
           &\pp _{r3} &\pp _{r4}  \\      \hline
\pp _{r4}  &\pc _r    &\pp _{r0}                         &\pp _{r1} &\pp _{rr}
           &\pp _{r4} &\pp _{r4}  \\      \hline
\end{array}
$
\end{center}
\caption{(Top) the converse $t^\smile$ and the rotation $t^\frown$ for each $\dltalg$ atom $t$;
         (Middle and Bottom) the $\dltalg$ composition tables (case
         $1$, case $2$, case $3$ and case $4$, respectively).
         $\pp _{ll}$ and $\pp _{rr}$ stand, respectively, for
         $\{\pp _{l0},\pp _{l1},\pp _{l2}\}$ and
         $\{\pp _{r2},\pp _{r3},\pp _{r4}\}$.}\label{TernaryComp1}
\end{footnotesize}
\end{figure*}
\subsection{The rotational component}
It is important to insist at this point on the importance, for the translational component $\dltalg$, of the oriented-ness of $\dlignes$:
if the objects we are dealing with were simple $\ulignes$, we would not be able, when two
lines both cut a third line, to say more than whether they cut it at the same point or at
distinct points (specifically, when the cutting points are distinct, saying that one of
the lines comes before the other would make no sense); similarly, we would only be able
to say, when two lines are parallel, whether they coincide or not.

If we consider the rotational knowledge present in the $\dltalg$ relations,
i.e., the knowledge on the relative angles of the three arguments, we realise that this consists, for
pairs $(x,y)$ of the three arguments, of knowledge of the form $\langle x,y\rangle\in (0,\pi )\cup (\pi ,2\pi )$,
inferrable from $x$ and $y$ being cutting $\dlignes$, or of the form $\langle x,y\rangle\in\{\pi ,2\pi\}$, inferrable
from $x$ and $y$ being parallel $\dlignes$. However, so restricting the rotational expressiveness would
mean that we are not exploiting the oriented-ness of the $\dligne$ arguments. In other words, this would
mean that we are using $\dlignes$ as if they were simple $\ulignes$.
The oriented-ness of $\dlignes$, again, makes them much richer than $\ulignes$, so that we can, for
instance, say that a $\dligne$ is to the left of, or opposite to, another $\dligne$; a level of relation
granularity which cannot be reached using the universe of $\ulignes$.

It should be clear that the relations $\cuts$ and $\parallelity$ relate to the $\apra$ relations
in the following way. For all $x,y\in\dlines$:
\begin{eqnarray}
\cuts (x,y)        &\Leftrightarrow &\{l,r\}(\orient (x),\orient (y)) \nonumber \\
\parallelity (x,y) &\Leftrightarrow &\{e,o\}(\orient (x),\orient (y)) \nonumber
\end{eqnarray}
On the other hand, it is easy to see that
the rotational information recorded by the four relations $\cc$, $\cp$, $\pc$ and $\pp$ can be
expressed using the RA $\atra$. Namely, for all $x,y,z\in\dlines$:
\begin{eqnarray}
\cc (x,y,z) &\Leftrightarrow &\phi _1 (\orient (x),\orient (y),\orient (z)) \nonumber \\
\cp (x,y,z) &\Leftrightarrow &\phi _2 (\orient (x),\orient (y),\orient (z)) \nonumber \\
\pc (x,y,z) &\Leftrightarrow &\phi _3 (\orient (x),\orient (y),\orient (z)) \nonumber \\
\pp (x,y,z) &\Leftrightarrow &\phi _4 (\orient (x),\orient (y),\orient (z)) \nonumber
\end{eqnarray}
where $\phi _1$, $\phi _2$, $\phi _3$ and $\phi _4$ are the following $\atra$ relations, defining a
partition of the set $\atraats$ of all $\atra$ atoms:
\begin{eqnarray}
\phi _1 &=&\{\lrl ,\lel ,\lll ,\llr ,\lor ,\lrr ,\rll ,\rol ,\rrl ,\rrr ,\rer ,\rlr\} \nonumber \\
\phi _2 &=&\{\lre ,\llo ,\rle ,\rro \} \nonumber \\
\phi _3 &=&\{\elll ,\err ,\orl ,\olr \} \nonumber \\
\phi _4 &=&\{\eee ,\eoo ,\ooe ,\oeo\} \nonumber
\end{eqnarray}
The first two rows in Figure \ref{CycordsRelation}(a) illustrate the $\atra$ atoms in $\phi _1$, the third row
illustrates the atoms in $\phi _2$, the fourth row illustrates the atoms in $\phi _3$,
and the bottom row illustrates the atoms in $\phi _4$.

In other words, the rotational expressiveness of what we have defined so far reduces to the four
$\atra$ relations $\phi _1$, $\phi _2$, $\phi _3$ and $\phi _4$ above.
We augment the rotational component by using the whole RA $\atra$. From now on, given a $\atra$
relation $R$ and three $\dlignes$ $x$, $y$ and $z$, we use the notation $R(x,y,z)$ as a synonym to
$R({\cal O}(x),{\cal O}(y),{\cal O}(z))$:
\begin{eqnarray}
(\forall R\in\atra )(\forall x,y,z\in\dlines )(R(x,y,z) &\Leftrightarrow &R({\cal O}(x),{\cal O}(y),{\cal O}(z))) \nonumber
\end{eqnarray}
\subsection{The final algebra}
From now on, we refer to
the final algebra as $\dlalg$ (Positional Algebra of ternary relations ---over $\dlines$).
\begin{figure}
\input{figures-directory/all_atoms3.latex}
\caption{Each $\dltalg$ atom in $\cc =\{\cc _<,\cc _=,\cc _>\}$ is compatible with each $\atra$ atom $r$ in
                                        $\phi _1=\{\lrl ,\lel ,\lll ,\llr ,\lor ,\lrr ,\rll ,\rol ,\rrl ,\rrr ,\rer ,\rlr\}$
                                        (see the top row for $r=\lrl$);
         each $\dltalg$ atom in $\cp =\{\cp _l,\cp _c,\cp _r\}$ is compatible with each $\atra$ atom $r$ in
                                        $\phi _2=\{\lre ,\llo ,\rle ,\rro \}$
                                        (see the second row from the top for $r=\llo$);
         each $\dltalg$ atom in $\pc =\{\pc _l,\pc _c,\pc _r\}$ is compatible with each $\atra$ atom $r$ in
                                        $\phi _3=\{\elll ,\err ,\orl ,\olr \}$
                                        (see the third row from the top for $r=\elll$); and
         each $\dltalg$ atom in $\pp =\{\pp _{l0},\pp _{l1},\pp _{l2},\pp _{l3},\pp _{l4},\pp _{c0},\pp _{c1},
                                     \pp _{c2},\pp _{r0},\pp _{r1},\pp _{r2},\pp _{r3},\pp _{r4}\}$ is compatible
                                     with each $\atra$ atom $r$ in
                                        $\phi _4=\{\eee ,\eoo ,\ooe ,\oeo\}$
                                        (see the last three rows from the top for $r=\eoo$).}\label{TranslationIllustration}
\end{figure}
\begin{figure}
\input{figures-directory/indep.latex}
\caption{Each $\dltalg$ atom in $\cc$ is compatible with each $\atra$ atom $r$ in
                                        $\phi _1$
                                        (see the top pair of boxes for $t=\cc _<$);
         each $\dltalg$ atom in $\cp$ is compatible with each $\atra$ atom $r$ in
                                        $\phi _2$
                                        (see the second pair of boxes from the top for $t=\cp _l$);
         each $\dltalg$ atom in $\pc$ is compatible with each $\atra$ atom $r$ in
                                        $\phi _3$
                                        (see the third pair of boxes from the top for $t=\pc _l$); and
         each $\dltalg$ atom in $\pp$ is compatible
                                     with each $\atra$ atom $r$ in
                                        $\phi _4$
                                        (see the last pair of boxes from the top for $t=\pp _{l0}$).}\label{independance}
\end{figure}
\begin{df}[the $\dlalg$ atoms](1) A $\dltalg$ atom $t$ is compatible with a $\atra$ atom
$r$, denoted by $\compatible (t,r)$, if and only if there exists a configuration of
three $\dlignes$ $x$, $y$ and $z$ such that both $t(x,y,z)$ and $r(x,y,z)$ hold;
(2) a $\dltalg$ atom $t$ and a $\atra$ atom $r$ such that $\compatible (t,r)$ define a
$\dlalg$ atom, which we refer to as $\langle t,r\rangle$.
\end{df}
Figure \ref{TranslationIllustration} considers one atom $r$ for each of the four $\atra$ disjunctive relations $\phi _1$,
$\phi _2$, $\phi _3$ and $\phi _4$, and illustrates all $\dlalg$ atoms $\langle t,r\rangle$ by
considering all $\dltalg$ atoms $t$ that are compatible with $r$. For each such $r$:
\begin{enumerate}
  \item the figure provides a spatial scene of three $\dlignes$ $x$, $y$ and $z$ satisfying $r$; i.e., such that
    $r(x,y,z)$; and
  \item for each $\dltalg$ atom $t$ that is compatible with $r$ ---$\langle t,r\rangle$ being
    therefore a $\dlalg$ atom--- the figure provides a spatial scene of three $\dlignes$ $x$,
    $y$ and $z$ satifying $\langle t,r\rangle$; i.e., such that $\langle t,r\rangle (x,y,z)$.
\end{enumerate}
More generally, each $\dltalg$ atom in $\cc =\{\cc _<,\cc _=,\cc _>\}$ (resp.
               $\cp =\{\cp _l,\cp _c,\cp _r\}$,
               $\pc = \{\pc _l,\pc _c,\pc _r\}$,
               $\pp =\{\pp _{l0},\pp _{l1},\pp _{l2},\pp _{l3},\pp _{l4},\pp _{c0},\pp _{c1},
                \pp _{c2},\pp _{r0},\pp _{r1},\pp _{r2},$\\
               $\pp _{r3},\pp _{r4}\}$)
is compatible with each $\atra$ atom in $\phi _1$ (resp. $\phi _2$, $\phi _3$, $\phi _4$).
Thus the set of all $\dlalg$ atoms is
\begin{footnotesize}
\begin{eqnarray}
\dlats &=&\{\langle t,r\rangle |(t\in\cc\wedge r\in\phi _1)\vee
                                        (t\in\cp\wedge r\in\phi _2)\vee
                                        (t\in\pc\wedge r\in\phi _3)\vee
                                        (t\in\pp\wedge r\in\phi _4)\} \label{dlalg-ats}\nonumber
\end{eqnarray}
\end{footnotesize}
The total number of $\dlalg$ atoms is $3\times 12+3\times 4+3\times 4+13\times 4=112$.
\begin{df}[projection and cross product]Let $T$ be a $\dltalg$ relation, $R$ a $\atra$ relation, 
and $S$ a $\dlalg$ relation:
\begin{enumerate}
  \item The translational projection, $\proj _t(S)$, and the rotational projection,
    $\proj _r(S)$, of $S$ are the $\dltalg$ relation and the $\atra$ relation, respectively,
    defined as follows:
    \begin{eqnarray}
      \proj _t(S) &=&\{t\in\dltats |(\exists r\in\atraats )(\langle t,r\rangle\in S)\} \nonumber \\
     \proj _r(S)  &=&\{r\in\atraats |(\exists t\in\dltats )(\langle t,r\rangle\in S)\} \nonumber
    \end{eqnarray}
  \item The cross product, $\Pi (T,R)$, of $T$ and $R$ is the $\dlalg$ relation defined as follows:
    \begin{eqnarray}
      \Pi (T,S) &=&\{\langle t,r\rangle\in\dlats |(t\in T)\wedge (r\in R)\} \nonumber
    \end{eqnarray}
    The notation $\langle T,S\rangle$ will be used synonymously to $\Pi (T,S)$.
  \item $S$ is projectable if it is equal to the cross product of its translational projection and its
    rotational projection; i.e., if $S=\Pi (\proj _t(S),\proj _r(S))$.
\end{enumerate}
\end{df}

\subsection{RDFs and TDFs of $\dlignes$: their independance}\label{rdftdfindependance}
Consider a $\atra$ atom, say $r$, and three $\dlignes$ $X$, $Y$ and $Z$ such that $r(X,Y,Y)$. For all $\dltalg$ atoms $t$ that are compatible with $r$, one can find $\dlignes$ $X'$, $Y'$ and $Z'$ that are translations of $X$, $Y$ and $Z$, respectively, thus verifying $r(X',Y',Y')$, such that $t(X',Y',Y')$:
\begin{eqnarray}
r(X',Y',Y')\wedge t(X',Y',Y')\nonumber
\end{eqnarray}
This is illustrated in Figure \ref{TranslationIllustration} for each $\atra$ atom $r$ in $\{\lrl ,\llo ,\elll ,\eoo\}$.

In a similar way, consider a $\dltalg$ atom, say $t$, and three $\dlignes$ $X$, $Y$ and $Z$ such that $t(X,Y,Y)$. For all $\atra$ atoms $r$ that are compatible with $t$, one can find $\dlignes$ $Y'$ and $Z'$ that are rotations of $Y$ and $Z$, respectively, each about its intersecting point with $X$ thus verifying $r(X',Y',Y')$, such that $t(X',Y',Y')$:
\begin{eqnarray}
r(X',Y',Y')\wedge t(X',Y',Y')\nonumber
\end{eqnarray}
This is illustrated in Figure \ref{independance} for each $\atra$ atom $t$ in $\{\cc _< ,\cp _l ,\pc _l ,\pp _{l0}\}$.

\subsection{The operations applied to the $\dlalg$ atoms}
The converse table, the rotation table and the composition tables of $\atra$ can be found in
\cite{Isli98a,Isli00b} (the converse table and the rotation table are reproduced in Figre
\ref{con-rot-atra}).
For $\dltalg$,
Figure \ref{TernaryComp1} provides the converse table and the composition tables.
Thus, thanks to the independence property discussed above, the converse and the
composition of $\dlalg$ atoms can be obtained from the converse
and the composition of the atoms of the translational component, $\dltalg$, and the
converse and the composition of the rotational component, $\atra$. Namely, if
$s_1=\langle t_1,r_1\rangle$ and $s_2=\langle t_2,r_2\rangle$ are two $\dlalg$ atoms, then:
\begin{eqnarray}
(s_1)^\smile &=&\Pi ((t_1)^\smile ,(r_1)^\smile )=\langle (t_1)^\smile ,(r_1)^\smile\rangle \nonumber \\
s_1\circ s_2 &=&\Pi (t_1\circ t_2,r_1\circ r_2)=\langle t_1\circ t_2,r_1\circ r_2\rangle \nonumber
\end{eqnarray}
As an example:
\begin{enumerate}
  \item from $(\pp _{l0})^\smile =\pp _{l2}$ and $(\ooe )^\smile =\eoo$, we get
$\langle\pp _{l0},\ooe\rangle ^\smile =\langle\pp _{l2},\eoo\rangle$;
  \item from $\cc _<\circ\cc _==\cc _<$ and $\rrl\circ\lrr =\rlr$, we get
$\langle\cc _<,\rrl\rangle\circ\langle\cc _=,\lrr\rangle =\langle\cc _<,\rlr\rangle$; and
  \item from $\cc _<\circ\cp _l=\cp _l$ and $\rrl\circ\llo =\rro$, we get
$\langle\cc _<,\rrl\rangle\circ\langle\cp _l,\llo\rangle =\langle\cp _l,\rro\rangle$.
\end{enumerate}
The $\dltalg$ relations take into account only the orientation of the first argument,
which is sufficient, given the knowledge, summarised below, the algebra is supposed
to represent:
\begin{enumerate}
  \item If the last two arguments both cut the first, the algebra is supposed to
    represent the order of the cutting points, in the walk along the first argument
    heading the positive direction.
  \item If one of the last two arguments is parallel to the first, the algebra is
    supposed to represent the side of the first argument (the left half-plane, the
    argument itself, or the right half-plane) the parallel $\dligne$ lies in.
  \item If the last two arguments are both parallel to the first, the algebra is
    supposed to represent the order in which the three arguments are encountered,
    when we walk perpendicularly to, from the left half-plane
    and heading towards the right half-plane bounded by, the first argument.
\end{enumerate}
The orientations of the last two arguments are ignored by the $\dltalg$ relations.
This has an effect on the computation of the rotations of the $\dltalg$ atoms, as
explained below.

Composition records the relation one can infer on the triple $(x,y,t)$, given a
relation $r_1$ on a triple $(x,y,z)$ and a relation $r_2$ on a triple $(x,z,t)$. For the particular case of $\dltalg$,
since $r_1$ and $r_2$ hold on the triples $(x,y,z)$ and $(x,z,t)$, respectively, this means that the
only orientation that is taken into consideration in the two relations is that of the common first
argument, $x$; and since the relation $R$ we want to infer is on the triple $(x,y,t)$, which also has
$x$ as the first argument, we can, just from the way $y$ and $z$, on the one hand, and $z$ and $t$,
on the other hand, compare relative to $x$, easily infer how the extreme variables, $y$ and $t$,
compare relative to the same variable $x$ (again, this is very similar to work done so far on temporal
relations, such as point-point relations \cite{Vilain86a}, interval-interval relations
\cite{Allen83b}, and point-interval and interval-point relations \cite{Ligozat91a,Vilain82a}).

Similarly to composition, from a $\dltalg$ atom $r$ on a triple $(x,y,z)$, which, again, takes
into consideration only the orientation of the first argument, $x$, the converse operation needs to
find a relation $r^\smile$ on the triple $(x,z,y)$, which needs to take into consideration only the
orientation of the first argument, which happens to be also $x$ (i.e., the same argument as the one
$r$ takes into consideration).

Computing the composition and the converse for the $\dltalg$ atoms poses thus no problem.
This is however not the case when considering the rotation operation. From a $\dltalg$
atom $r$ on $(x,y,z)$, which takes the orientation of $x$ into account,  the operation needs to
infer the relation $r^\frown$ on $(y,z,x)$, which needs, but cannot get from what is known, the
orientation of the first argument, $y$. 
Instead of showing how to get the rotation of the $\dlalg$
atoms from the rotation of the $\dltalg$ atoms and the rotation of the $\atra$
atoms, which is possible but not as straightforward as for composition and converse, we
preferred to dress a 112-element rotation table recording for each $\dlalg$ atom
$\langle t,r\rangle$ its rotation $\langle t,r\rangle ^\frown$ (see Figure \ref{dla-rot}).
\begin{figure}[t]
\begin{scriptsize}
\begin{center}
$
\begin{array}{|l||l|l|l|l|l|l|}  \hline
           &\lrl                       &\lel                       &\lll                       &\llr
           &\lor                       &\lrr   \\  \hline\hline
\cc _<     &\langle\cc _<,\rrr\rangle  &\langle\pc _r,\err\rangle  &\langle\cc _>,\lrr\rangle  &\langle\cc _<,\llr\rangle
           &\langle\pc _r,\olr\rangle  &\langle\cc _>,\rlr\rangle  \\  \hline
\cc _=     &\langle\cc _=,\rrr\rangle  &\langle\pc _c,\err\rangle  &\langle\cc _=,\lrr\rangle  &\langle\cc _=,\llr\rangle
           &\langle\pc _c,\olr\rangle  &\langle\cc _=,\rlr\rangle  \\  \hline
\cc _>     &\langle\cc _>,\rrr\rangle  &\langle\pc _l,\err\rangle  &\langle\cc _<,\lrr\rangle  &\langle\cc _>,\llr\rangle
           &\langle\pc _l,\olr\rangle  &\langle\cc _<,\rlr\rangle
                              \\  \hline
\end{array}
$\vskip 0.2cm
$
\begin{array}{|l||l|l|l|l|l|l|}  \hline
           &\rll                       &\rol
           &\rrl                       &\rrr                       &\rer                       &\rlr   \\  \hline\hline
\cc _<     &\langle\cc _>,\lrl\rangle  &\langle\pc _l,\orl\rangle
           &\langle\cc _<,\rrl\rangle  &\langle\cc _>,\rll\rangle  &\langle\pc _l,\elll\rangle  &\langle\cc _<,\lll\rangle  \\  \hline
\cc _=     &\langle\cc _=,\lrl\rangle  &\langle\pc _c,\orl\rangle
           &\langle\cc _=,\rrl\rangle  &\langle\cc _=,\rll\rangle  &\langle\pc _c,\elll\rangle  &\langle\cc _=,\lll\rangle  \\  \hline
\cc _>     &\langle\cc _<,\lrl\rangle  &\langle\pc _r,\orl\rangle
           &\langle\cc _<,\rrl\rangle  &\langle\cc _<,\rll\rangle  &\langle\pc _r,\elll\rangle  &\langle\cc _>,\lll\rangle
                              \\  \hline
\end{array}
$\vskip 0.2cm
$
\begin{array}{|l||l|l|l|l|}  \hline
           &\lre                       &\llo                       &\rle                       &\rro   \\  \hline\hline
\cp _l     &\langle\cc _>,\rer\rangle  &\langle\cc _>,\lor\rangle  &\langle\cc _<,\lel\rangle  &\langle\cc _<,\rol\rangle  \\  \hline
\cp _c     &\langle\cc _=,\rer\rangle  &\langle\cc _=,\lor\rangle  &\langle\cc _=,\lel\rangle  &\langle\cc _=,\rol\rangle  \\  \hline
\cp _r     &\langle\cc _<,\rer\rangle  &\langle\cc _<,\lor\rangle  &\langle\cc _>,\lel\rangle  &\langle\cc _>,\rol\rangle  \\  \hline
\end{array}
$\hskip 0.5cm
$
\begin{array}{|l||l|l|l|l|}  \hline
           &\elll                      &\err                       &\orl                       &\olr  \\  \hline
\pc _l     &\langle\cp _r,\lre\rangle  &\langle\cp _r,\rle\rangle  &\langle\cp _l,\rro\rangle  &\langle\cp _l,\llo\rangle  \\  \hline
\pc _c     &\langle\cp _c,\lre\rangle  &\langle\cp _c,\rle\rangle  &\langle\cp _c,\rro\rangle  &\langle\cp _c,\llo\rangle  \\  \hline
\pc _r     &\langle\cp _l,\lre\rangle  &\langle\cp _l,\rle\rangle  &\langle\cp _r,\rro\rangle  &\langle\cp _r,\llo\rangle
                               \\  \hline
\end{array}
$\vskip 0.2cm
$
\begin{array}{|l||l|l|l|l|}  \hline
           &\eee                       &\eoo                       &\ooe                       &\oeo  \\  \hline\hline
\pp _{l0}  &\langle\pp _{l4},\eee\rangle  &\langle\pp _{l4},\ooe\rangle  &\langle\pp _{r0},\oeo\rangle  &\langle\pp _{r0},\eoo\rangle
  \\  \hline
\pp _{l1}  &\langle\pp _{c2},\eee\rangle  &\langle\pp _{c2},\ooe\rangle  &\langle\pp _{c0},\oeo\rangle  &\langle\pp _{c0},\eoo\rangle
  \\  \hline
\pp _{l2}  &\langle\pp _{r4},\eee\rangle  &\langle\pp _{r4},\ooe\rangle  &\langle\pp _{l0},\oeo\rangle  &\langle\pp _{l0},\eoo\rangle
  \\  \hline
\pp _{l3}  &\langle\pp _{r3},\eee\rangle  &\langle\pp _{r3},\ooe\rangle  &\langle\pp _{l1},\oeo\rangle  &\langle\pp _{l1},\eoo\rangle
  \\  \hline
\pp _{l4}  &\langle\pp _{r2},\eee\rangle  &\langle\pp _{r2},\ooe\rangle  &\langle\pp _{l2},\oeo\rangle  &\langle\pp _{l2},\eoo\rangle
  \\  \hline
\pp _{c0}  &\langle\pp _{l3},\eee\rangle  &\langle\pp _{l3},\ooe\rangle  &\langle\pp _{r1},\oeo\rangle  &\langle\pp _{r1},\eoo\rangle
  \\  \hline
\pp _{c1}  &\langle\pp _{c1},\eee\rangle  &\langle\pp _{c1},\ooe\rangle  &\langle\pp _{c1},\oeo\rangle  &\langle\pp _{c1},\eoo\rangle
  \\  \hline
\pp _{c2}  &\langle\pp _{r1},\eee\rangle  &\langle\pp _{r1},\ooe\rangle  &\langle\pp _{l3},\oeo\rangle  &\langle\pp _{l3},\eoo\rangle
  \\  \hline
\pp _{r0}  &\langle\pp _{l2},\eee\rangle  &\langle\pp _{l2},\ooe\rangle  &\langle\pp _{r2},\oeo\rangle  &\langle\pp _{r2},\eoo\rangle
  \\  \hline
\pp _{r1}  &\langle\pp _{l1},\eee\rangle  &\langle\pp _{l1},\ooe\rangle  &\langle\pp _{r3},\oeo\rangle  &\langle\pp _{r3},\eoo\rangle
  \\  \hline
\pp _{r2}  &\langle\pp _{l0},\eee\rangle  &\langle\pp _{l0},\ooe\rangle  &\langle\pp _{r4},\oeo\rangle  &\langle\pp _{r4},\eoo\rangle
  \\  \hline
\pp _{r3}  &\langle\pp _{c0},\eee\rangle  &\langle\pp _{c0},\ooe\rangle  &\langle\pp _{c2},\oeo\rangle  &\langle\pp _{c2},\eoo\rangle
  \\  \hline
\pp _{r4}  &\langle\pp _{r0},\eee\rangle  &\langle\pp _{r0},\ooe\rangle  &\langle\pp _{l4},\oeo\rangle  &\langle\pp _{l4},\eoo\rangle
  \\  \hline
\end{array}
$
\end{center}
\end{scriptsize}
\caption{The rotations of the $\dlalg$ atoms: for each of the five tables, the leftmost element of a line is a $\dltalg$
atom, $t$, and the top element of a column is a $\atra$ atom, $r$, and the entry at the intersection of the line and the
column records the rotation of the $\dlalg$ atom $\langle t,r\rangle$.}\label{dla-rot}
\end{figure}
\begin{remk}
A $\atra$ relation $R$ is equivalent to the $\dlalg$ relation consisting of all $\dlalg$ atoms $\langle r,t\rangle$ verifying $r\in R$:
\begin{eqnarray}
R&\equiv&\{\langle r,t\rangle\in\dlats |r\in R\}\nonumber
\end{eqnarray}
In a similar way, a $\dltalg$ relation $T$ is equivalent to the $\dlalg$ relation consisting of all $\dlalg$ atoms $\langle r,t\rangle$ verifying $t\in T$:
\begin{eqnarray}
T&\equiv&\{\langle r,t\rangle\in\dlats |t\in T\}\nonumber
\end{eqnarray}
\end{remk}
\section{Reasoning about $\dlalg$ relations: $\dlalg$-CSPs}\label{dlacsps}
A $\dlalg$-CSP (resp. $\atra$-CSP, $\dltalg$-CSP) is a CSP
\cite{Mackworth77a,Montanari74a} of ternary constraints
(ternary CSP), of which
\begin{enumerate}
  \item the variables range over the set $\dlines$ of $\dlignes$; and
  \item the constraints consist of $\dlalg$ (resp. $\atra$, $\dltalg$) relations on
(triples of) the variables.
\end{enumerate}
A CSP of either of the three forms is said to be
atomic if the entries of its constraint matrix all consist of atomic relations. A
scenario is a refinement which is atomic.
The translational (resp. rotational) projection,
$\proj _t(P)$ (resp. $\proj _r(P)$), of a $\dlalg$-CSP $P$ is the $\dltalg$-CSP
(resp. $\atra$-CSP) defined as follows:
\begin{enumerate}
  \item the variables are the same as the ones of $P$; and
  \item the constraint matrix of the projection is obtained by projecting the
    constraint matrix of $P$ onto $\dltalg$ (resp. $\atra$):
    \begin{eqnarray}
    (\forall i,j,k)[(\termat ^{\proj _t(P)})_{ijk}=\proj _t([\termat ^P]_{ijk})]\nonumber
    \end{eqnarray}
    \begin{eqnarray}
    \mbox{(resp. }
    (\forall i,j,k)([\termat ^{\proj _r(P)}]_{ijk}=\proj _r([\termat ^P]_{ijk})))\nonumber
    \end{eqnarray}
\end{enumerate}
The solution search algorithm in \cite{Isli98a,Isli00b} for $\atra$-CSPs, which we refer to as
$\icsa$ algorithm,
can be easily adapted so that it searches for a
$4$-consistent scenario of a $\dlalg$-CSP, if any, or otherwise reports
inconsistency. The $\icsa$ algorithm differs from Ladkin and Reinefeld's
 \cite{Ladkin92a} in that:
\begin{enumerate}
  \item it refines the relation on a triple of variables at each node of the
    search tree, instead of the relation on a pair of variables; and
  \item it makes use of a procedure achieving $4$-consistency, in the
    preprocessing step and as the filtering method during the search, instead
    of a path consistency procedure.
\end{enumerate}
On the other hand, the $4$-consistency procedure in \cite{Isli98a,Isli00b} for $\atra$-CSPs,
which we refer to as $\icpa$ algorithm, can be adapted so that it achieves
$4$-consistency for a $\dlalg$-CSP. Such an
adaptation would repeat the following steps until either the empty relation is
detected (indicating inconsistency), or a fixed point is reached, indicating
that the CSP has been made $4$-consistent:
\begin{enumerate}
  \item consider a quadruple $(X_i,X_j,X_k,X_l)$ of variables verifying
        $(\termat ^P)_{ijl}\not\subseteq ((\termat ^P)_{ijk}\circ (\termat ^P)_{ikl})$;
  \item $(\termat ^P)_{ijl}\leftarrow (\termat ^P)_{ijl}\cap (\termat ^P)_{ijk}\circ (\binmat ^P)_{ikl}$;
  \item if $((\termat ^P)_{ijl}=\emptyset)$ then exit (the CSP is inconsistent).
\end{enumerate}
The reader is referred to \cite{Isli98a,Isli00b} for more details on the
$\icsa$ and $\icpa$ algorithms. $\icpa$ achieves $4$-consistency for
$\dlalg$-CSPs.
\begin{thr}\label{sfcthm2}
Let $P$ be a $\dlalg$-CSP. Applied to $P$, the $\icpa$ algorithm either
detects that $P$ is inconsistent, or achieves
strong $4$-consistency for $P$.
\end{thr}
{\bf Proof.} Suppose that $\icpa$ \cite{Isli98a,Isli00b} applied
to $P$ does not detect any inconsistency: we show that $P$ has then been made strongly
$4$-consistent. The definition of composition for ternary relations implies that, if $P$ is
closed under the $4$-consistency operation,
$(\termat ^P)_{ijl}\leftarrow (\termat ^P)_{ijl}\cap (\termat ^P)_{ijk}\circ (\binmat ^P)_{ikl}$,
which is the case if $P$ is $4$-consistent, then any solution to any $3$-variable sub-CSP
extends to any fourth variable, as long as the composition computed from the composition
tables matches the exact composition;
i.e., as long as, given any two $\dlalg$ atoms, say $r$ and $s$, it is the case that
$r\circ s=T[r,s]$, where $T[r,s]$ is the computed composition of $r$ and $s$ (computed, as we
have seen, as the cross product of the composition of the translational projections of $r$ and
$s$, on the one
hand, and the composition of the rotational projections of $r$ and $s$, on the other hand).
But this is the case since $\dlalg$ is an RA ---see Appendix \ref{appendixb}. \cqfd

The important question now is whether the $\icpa$ algorithm in
\cite{Isli98a,Isli00b} is complete for atomic $\dlalg$-CSPs. A positive answer would imply,
on the one hand, that we can check complete knowledge, expressed in $\dlalg$ as an atomic CSP, in
polynomial time; and,
on the other hand, that a general CSP expressed in $\dlalg$ can be checked for consistency using the
$\icsa$ solution search algorithm in \cite{Isli98a,Isli00b}. We show that the answer is almost
in the affirmative: completeness holds for a set $\tractablesubset$, defined below, of $\dlalg$
JEPD relations including almost all of the $\dlalg$ atomic relations.
\begin{df}[$\tractablesubset =\tractablesubset _1\cup\tractablesubset _2$]\label{def-sone-stwo}
The set $\tractablesubset$ of $\dlalg$ relations is defined as
$\tractablesubset =\tractablesubset _1\cup\tractablesubset _2$, with the subsets
$\tractablesubset _1$ and $\tractablesubset _2$ defined as follows:
\begin{enumerate}
  \item $\tractablesubset _1$ is the set of all $\dlalg$ atomic relations
    $\{\langle t,s\rangle\}$ holding on triples of $\dlignes$ involving at least two
    arguments that are parallel to each other ---strictly parallel or coincide.
  \item $\tractablesubset _2$ is the set of all $\dlalg$ relations of the form
    $\{\langle\cc _<,s\rangle ,\langle\cc _=,s\rangle ,\langle\cc _>,s\rangle\}$,
    where $s$ is any $\atra$ atom from the set
    $\pairwisecutting =\{\lll ,\llr ,\lrl ,\lrr ,\rll ,\rlr ,\rrl ,\rrr\}$. Each
    element of $\pairwisecutting$ has the property that it is compatible with and
    only with each $\dltalg$ atom from the set
    $\twoandthreecutone =\{\cc _<,\cc _=,\cc _>\}$. Therefore the set
    $\tractablesubset _2$ can be written either as
    $\{\{\langle\cc _<,s\rangle ,\langle\cc _=,s\rangle ,\langle\cc _>,s\rangle\}|
       s\in\pairwisecutting\}$, or as a set of $\atra$ atomic relations,
    $\{\{s\}|
       s\in\pairwisecutting\}$.
\end{enumerate}
\end{df}
The relations in $\tractablesubset$ are JEPD; their particularity is that they can
represent the knowledge consisting of the order in which two $\dlignes$ cut a
third $\dligne$ only if the first two $\dlignes$ are parallel to each other. For
triples of $\dlignes$ that are pairwise cutting (no two arguments are parallel to
each other), the relations in $\tractablesubset$ can represent only their
rotational knowledge ---namely, they cannot represent the order in which any two
of the three arguments cut the third.
To summarise, we have $\tractablesubset=\tractablesubset _1\cup\tractablesubset _2$,
with
$\tractablesubset _1=\{\{\langle t,s\rangle\}|\langle t,s\rangle\in\dlats
                                     \wedge s\notin\pairwisecutting\}$ and
$\tractablesubset _2=\{\{\langle\cc _<,s\rangle ,\langle\cc _=,s\rangle ,\langle\cc _>,s\rangle\}|
                                     s\in\pairwisecutting\}$.

We refer to the subalgebra of $\dlalg$ generated by $\tractablesubset$ as $\cdlalg$ (coarse
$\dlalg$).
Each $\dlalg$ atomic relation $\{\langle t,s\rangle\}$ from the set $\tractablesubset _1$ gives
rise to an atom of the RA $\cdlalg$: the atom $\langle t,s\rangle$, which is also an
atom of $\dlalg$. Each $\dlalg$ relation
$\{\langle\cc _<,s\rangle ,\langle\cc _=,s\rangle ,\langle\cc _>,s\rangle\}$ from the
set $\tractablesubset _2$ ($s\in\pairwisecutting$) gives rise to an atom of the RA
$\cdlalg$: the atom $\langle *,s\rangle$, which is semantically the same as the $\atra$
atom $s$ (intuitively, the * symbol in the translational part of the atom says that the
atom records no translational information ---to say it differently, the information
recorded by the atom is the same as that recorded by the $\atra$ atom appearing in the
rotational part). The set $\cdlats$ of $\cdlalg$ atoms is thus
$\cdlats =\{\langle t,s\rangle |\langle t,s\rangle\in\dlats\wedge s\notin\pairwisecutting\}\cup
             \{\langle *,s\rangle |s\in\pairwisecutting\}$.

The next theorem states tractability of the set $\cdlar =\{\{r\}|r\in\cdlats\}$ of $\cdlalg$ atomic relations.
\begin{thr}\label{sfcthm}
Let $P$ be a $\cdlalg$-CSP expressed in the set $\cdlar$ of atomic relations. If $P$ is $4$-consistent then it is
globally consistent.
\end{thr}
The proof uses Helly's convexity theorem \cite{Chvatal83a}.
\begin{thr}[Helly's theorem \cite{Chvatal83a}]\label{hellystheorem}
Let $\Gamma$ be a set of convex regions of the $m$-dimensional space
$\BBR ^m$. If every $m+1$ elements of $\Gamma$ have a non empty
intersection then the intersection of all elements of $\Gamma$ is
non empty.
\end{thr}
We now prove Theorem \ref{sfcthm}.

{\bf Proof.} Let $P$ be a $4$-consistent atomic $\cdlalg$-CSP, $P'=\proj _t(P)$ and $P''=\proj _r(P)$.
From Theorem \ref{sfcthm2}, we get strong $4$-consistency of $P$.
From strong $4$-consistency of $P$, we get strong $4$-consistency of $P'$ and strong $4$-consistency of
$P''$. $P''$ being atomic and strongly $4$-consistent, it is globally consistent \cite{Isli98a,Isli00b}.
Let $S=(d_1,\ldots ,d_i,\ldots ,d_n)$ be an instantiation of the variable \mbox{$n$-tuple}
$(X_1,\ldots ,X_i,\ldots ,X_n)$ that is solution to $P''$. We can, and do, suppose
that the $d_i$'s are $\dlignes$ through $O$ (see Definition
\ref{isos}). We now show that we can move the $d_i$'s around, so
that the new instantiation of the variables is still solution to $P''$, and is
solution to $P'$ (thus to $P$): one way, which is the way followed, to make
the solution to $P''$ remain solution to $P''$ is to use only $\dlignes$
translation: translating a $\dligne$ does not modify its orientation. So all
that is needed, is to find the right translation, one that makes the solution
to $P''$ also solution to $P'$, thus to $P$.
Thanks to the property of independance of the translational component and
the rotational component of a $\dlalg$ atom (see Subsection \ref{rdftdfindependance}), $\langle t,r\rangle$, it is the
case that for all configurations $(\ell _1,\ell _2,\ell _3)$ of three
$\dlignes$, such that $r(\ell _1,\ell _2,\ell _3)$ holds, we can translate
the $\ell _i$'s, $i=1\ldots 3$, relative to one another (again, a translation
does not alter the rotational knowledge on the triple), so that the $\dltalg$
atom $t$ holds on the triple $(\ell _1',\ell _2',\ell _3')$, where $\ell _1'$,
$\ell _2'$ and $\ell _3'$ are, respectively, the translations of $\ell _1$, $\ell _2$ and $\ell _3$.

We go back to our rotational solution $S=(d_1,\ldots ,d_i,\ldots ,d_n)$. We suppose that we have successfully translated
$d_1,\ldots ,d_i$ ($i\geq 3$), so that the new instantiation
$(d_1',\ldots ,d_i')$ of the variable \mbox{$i$-tuple} $(X_1,\ldots ,X_i)$ is
solution to the sub-CSP $P_{|\{X_1,\ldots ,X_i\}}'$, thus to
$P_{|\{X_1,\ldots ,X_i\}}$. We show that we can translate $d_{i+1}$, so that
$(d_1',\ldots ,d_i',d_{i+1}')$, where $d_{i+1}'$ is
the new instantiation of $X_{i+1}$ resulting from the translation of $d_{i+1}$,  is
solution to $P_{|\{X_1,\ldots ,X_i,X_{i+1}\}}'$, thus to
$P_{|\{X_1,\ldots ,X_i,X_{i+1}\}}$.
For this purpose, we suppose that the 2D space is
associated with a system $(x,O,y)$ of coordinates. Without loss of generality, we
assume that the $x$-axis, $\overrightarrow{Ox}$, (is parallel to, and) has the same
orientation as the $\dligne$ $d_{i+1}$. As a consequence, all $\dlignes$ $d_j'$,
$j\in\{1,\cdots ,i\}$, that are parallel to $d_{i+1}$, are curves of equation of the
form $y=q_j$, where $q_j$ is a constant; furthermore, the equation of the $\dligne$
$d_{i+1}'$ we are looking for, should be of the form $y=q_{i+1}$, where $q_{i+1}$ is
a constant. The problem thus is to show that such a constant $q_{i+1}$ can be found.

Given this, we can write as follows the constraints relating the $\dligne$
$d_{i+1}'$ we are looking for, to the $\dlignes$ $d_1',\ldots ,d_i'$, constituting the
assignments of the variables already consistently instantiated, both rotationally
and translationally. Initialise $S$ to the empty set; then for all
$j,k\in\{1,\ldots ,i\}$, with $j\leq k$:
\begin{figure}
\input{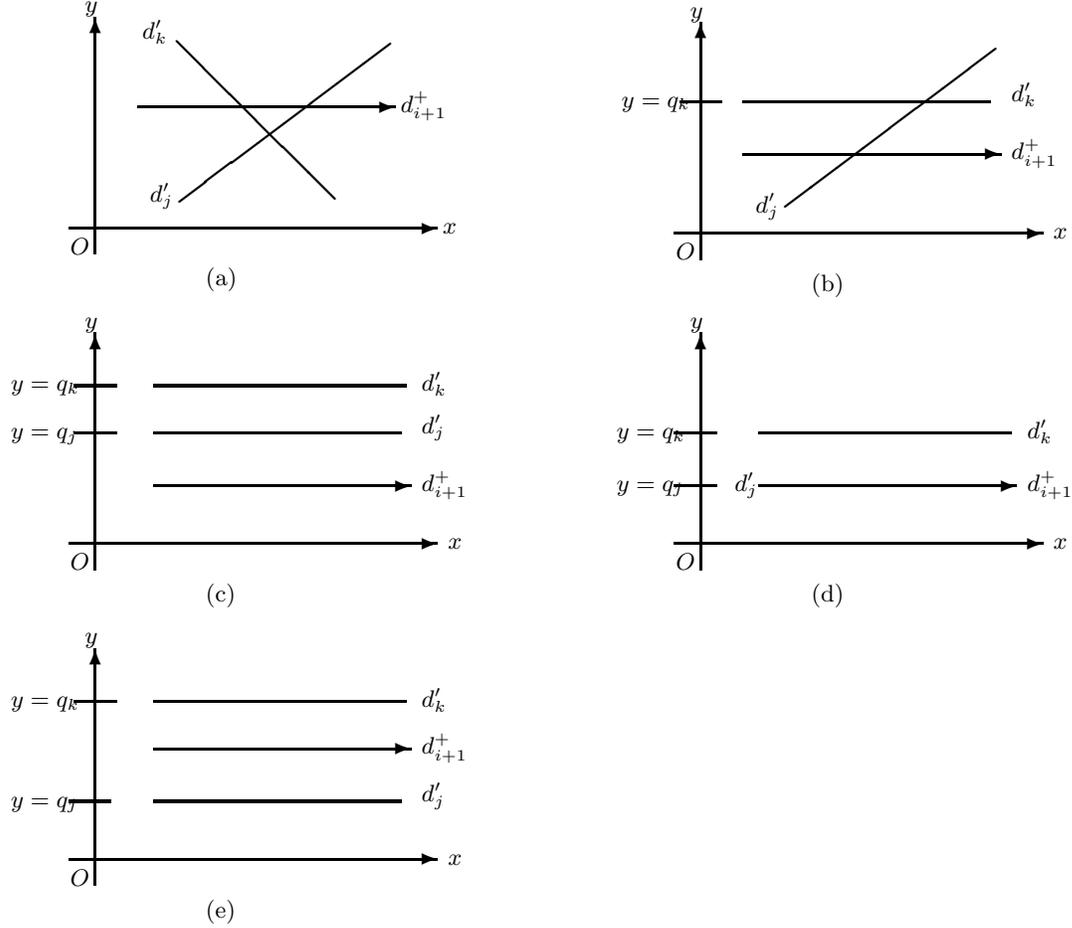}
\caption{Illustration of the proof of Theorem \ref{sfcthm}:
(a) all translations $d_{i+1}^+$ of $d_{i+1}$ satisfy the constraint
        $\cc (d_{i+1}^+,d_j',d_k')$;
(b) the constraint $\cp_l (d_{i+1}^+,d_j',d_k')$ is equivalent to the constraint
        that the equation of $d_{i+1}^+$ should be of the form $y<q_k$, $y=q_k$
        or $y>q_k$, depending on whether $m=l$, $m=c$ or $m=r$,
        respectively.}\label{proof-illustration}
\end{figure}
\begin{enumerate}
      \item if $(\termat ^P)_{(i+1)jk}=\{\langle *,s\rangle\}$ (with $s\in\pairwisecutting$).
        According to the definition of the $\cdlalg$ atoms, the * symbol refers to the
        $\dltalg$ relation $\cc =\{\cc _<,\cc _=,\cc _>\}$, consisting of all $\dltalg$ atoms that
        are compatible with each of the $\atra$ atoms from the set $\pairwisecutting$:
        $\cc (d_{i+1}',d_j',d_k')$ (see the illustration of Figure
        \ref{proof-illustration}(a)).
        All translations $d_{i+1}^+$ of $d_{i+1}$ satisfy the constraint
        $\cc (d_{i+1}^+,d_j',d_k')$, thus
        $(\termat ^P)_{(i+1)jk}(d_{i+1}^+,d_j',d_k')$, since they already satisfy the rotational
        constraint $s (d_{i+1}^+,d_j',d_k')$ ---nothing is added to $S$;
      \item if $(\termat ^P)_{(i+1)jk}=\{\langle\cp _m,s\rangle\}$,
        with $m\in\{l,c,r\}$, $s\notin\pairwisecutting$ (see the illustration of Figure
        \ref{proof-illustration}(b) for $m=l$), then the $\dligne$ $d_{i+1}^+$ we are looking for,
        should be so that $d_k'$ is parallel to $d_{i+1}^+$, and lies within the left open
        half-plane bounded by $d_{i+1}^+$ if $m=l$, coincides with $d_{i+1}^+$ if $m=c$, or lies
        within the right open half-plane bounded by $d_{i+1}^+$ if $m=r$.
        The translational sub-constraint $\cp_m (d_{i+1}^+,d_j',d_k')$ is equivalent to the
        constraint that the equation of $d_{i+1}^+$ should be of the form $y=q_{i+1}$, with
        $q_{i+1}$ being a constant, and $q_{i+1}<q_k$, $q_{i+1}=q_k$ or $q_{i+1}>q_k$, depending
        on whether $m=l$, $m=c$ or $m=r$, respectively. Add to $S$ the corresponding equivalent
        linear inequality:
\begin{center}
$
S\leftarrow
  \left\{
                   \begin{array}{ll}
                         S\cup\{q_{i+1}\in (-\infty ,q_k)\}
                               &\mbox{ if }m=l,  \\
                         S\cup\{q_{i+1}\in\{q_k\}\}
                               &\mbox{ if }m=c,  \\
                         S\cup\{q_{i+1}\in (q_k,+\infty )\}
                               &\mbox{ if }m=r .  \\
                   \end{array}
  \right.
$
\end{center}
      \item if $(\termat ^P)_{(i+1)jk}=\{\langle\pc _m,s\rangle\}$,
        with $m\in\{l,c,r\}$, $s\notin\pairwisecutting$.
        The translational sub-constraint $\pc_m (d_{i+1}',d_j',d_k')$ is equivalent to
        $\cp_m (d_{i+1}',d_k',d_j')$, obtained by swapping $d_j'$ and $d_k'$ and
        replacing the atom $\pc _m$ by its converse, $\cp _m$. In a similar way as in
        the previous point, we add to $S$ the equivalent linear inequality:
\begin{center}
$
S\leftarrow
  \left\{
                   \begin{array}{ll}
                         S\cup\{q_{i+1}\in (-\infty ,q_j)\}
                               &\mbox{ if }m=l,  \\
                         S\cup\{q_{i+1}\in\{q_j\}\}
                               &\mbox{ if }m=c,  \\
                         S\cup\{q_{i+1}\in (q_j,+\infty )\}
                               &\mbox{ if }m=r .  \\
                   \end{array}
  \right.
$
\end{center}
      \item  if $(\termat ^P)_{(i+1)jk}=\{\langle\pp _{mn},s\rangle\}$, with $m=l$ and $n\in\{0,\ldots ,4\}$
        (see the illustration of Figure \ref{proof-illustration}(c) for $m=l$ and $n=0$),
        $m=c$ and $n\in\{0,1,2\}$
        (see the illustration of Figure \ref{proof-illustration}(d) for $m=c$ and $n=0$),
        or $m=r$ and $n\in\{0,\ldots ,4\}$
        (see the illustration of Figure \ref{proof-illustration}(e) for $m=r$ and $n=0$),
        then the $\dligne$ $d_{i+1}'$ we are looking for, should be such that $d_j'$ and $d_k'$ are both
        parallel to it. $d_j'$ and $d_k'$ are thus curves of
        equations of the form $y=q_j$ and $y=q_k$, respectively, $q_j$ and $q_k$ being constants.
        The translational constraint $\pp _{mn} (d_{i+1}^+,d_j',d_k')$, which has to be satisfied, can be equivalently replaced by either
        of the two linear inequalities $q_{i+1}\alpha q_j$ or $q_{i+1}\beta q_k$, or by their conjunction
        $q_{i+1}\alpha q_j\wedge q_{i+1}\beta q_k$, with $\alpha ,\beta\in\{<,=,>\}$. Add to $S$ the
        corresponding equivalent linear inequality or conjunction of linear inequalities, in the
        following way:
\begin{center}
$
S\leftarrow
  \left\{
                   \begin{array}{ll}
                         S\cup\{q_{i+1}\in (-\infty ,q_j)\}
                               &\mbox{ if }m=l\mbox{ and }n\in\{0,1\},  \\
                         S\cup\{q_{i+1}\in (-\infty ,q_k)\}
                               &\mbox{ if }m=l\mbox{ and }n=2,  \\
                         S\cup\{q_{i+1}\in\{q_k\}\}
                               &\mbox{ if }m=l\mbox{ and }n=3,  \\
                         S\cup\{q_{i+1}\in (-\infty ,q_j),q_{i+1}\in (q_k,+\infty )\}
                               &\mbox{ if }m=l\mbox{ and }n=4,  \\
                         S\cup\{q_{i+1}\in\{q_j\}\}
                               &\mbox{ if }m=c,  \\
                         S\cup\{q_{i+1}\in (q_j,+\infty ),q_{i+1}\in (-\infty ,q_k)\}
                               &\mbox{ if }m=r\mbox{ and }n=0,  \\
                         S\cup\{q_{i+1}\in\{q_k\}\}
                               &\mbox{ if }m=r\mbox{ and }n=1,  \\
                         S\cup\{q_{i+1}\in (q_k,+\infty )\}
                               &\mbox{ if }m=r\mbox{ and }n\{2,3\},  \\
                         S\cup\{q_{i+1}\in (q_j,+\infty )\}
                               &\mbox{ if }m=r\mbox{ and }n=4  \\
                   \end{array}
  \right.
$
\end{center}
\end{enumerate}
Consider now two elements $q_{i+1}\in S_1$ and $q_{i+1}\in S_2$ of $S$, and
suppose that the sets $S_1$ and $S_2$ have an empty intersection. There would then exist
$j,k\in\{1,\ldots ,i\}$ such that $S_1\in\{(-\infty ,q_j),\{q_j\},(q_j,+\infty )\}$,
$S_2\in\{(-\infty ,q_k),\{q_k\},(q_k,+\infty )\}$ and $S_1\cap S_2=\emptyset$.
From the construction of $S$, we get that there exist $j_1,k_1\in\{1,\ldots ,i\}$ such
that $(\termat ^{P'})_{(i+1)jj_1}(d_{i+1}^+,d_j',d_{j_1}')\Rightarrow q_{i+1}\in S_1$
and $(\termat ^{P'})_{(i+1)kk_1}(d_{i+1}^+,d_k',d_{k_1}')\Rightarrow q_{i+1}\in S_2$.
Because the elements of $\{(-\infty ,q_j),\{q_j\},(q_j,+\infty )\}$, on the one
hand, and the elements of $\{(-\infty ,q_k),\{q_k\},(q_k,+\infty )\}$, on the other
hand, are jointly exhaustive (their union gives the whole set of real numbers) and
pairwise disjoint, and the CSP $P'$ strongly $4$-consistent, it must be the
case that for all $l\in\{1,\ldots ,i\}$,
$(\termat ^{P'})_{(i+1)jl}(d_{i+1}^+,d_j',d_l')\Rightarrow q_{i+1}\in S_1$ and
$(\termat ^{P'})_{(i+1)kl}(d_{i+1}^+,d_k',d_l')\Rightarrow q_{i+1}\in S_2$. Strong
$4$-consistency of the CSP implies that it must be the case that $S_1$ and $S_2$
have a nonempty intersection.

Let $S'$ be the set of all elements of the form $A$ such that $q_{i+1}\in A$ is
element of $S$:
\begin{center}
$S'=\{A|\{q_{i+1}\in A\}\subseteq S\}$
\end{center}
The point now is that the elements of $S'$ are convex subsets of the set $\BBR$ of real
numbers. The elements of $S'$ being pairwise intersecting, Helly's theorem \cite{Chvatal83a}
specialised to $m=2$ (see Theorem \ref{hellystheorem}) implies that the intersection
of all elements in $S'$ is non empty. Any translation $d_{i+1}'$ of $d_{i+1}$ of equation
$y=q_{i+1}$, with $q_{i+1}$ being a constant from $S'$, would make the
tuple $(x_1',\ldots ,x_i',x_{i+1}')$ solution to $P_{|\{X_1,\ldots ,X_i,X_{i+1}\}}'$, thus
to $P_{|\{X_1,\ldots ,X_i,X_{i+1}\}}$. \cqfd

Given that the $\cdlalg$ atomic relations are tractable, and that an atomic $\cdlalg$-CSP
can be be solved using the $\icpa$ propagation algorithm in \cite{Isli98a,Isli00b}, it
follows that a general $\cdlalg$-CSP can be solved using the $\icsa$ search algorithm also
in \cite{Isli98a,Isli00b}, alluded to before.
\begin{corol}
Let $P$ be a $\cdlalg$-CSP. The consistency problem of $P$ can be solved using the $\icsa$
search algorithm in \cite{Isli98a,Isli00b}.
\end{corol}
\section{Use of the RA $\dlalg$}\label{dlause}
The RA $\dlalg$ is clearly well suited for applications such as robot
localisation and navigation, Geographical Information Systems ($\gis$), and
shape description. First, thanks to the
objects the RA deals with, namely, $\dlignes$: such an object is much
richer than a simple $\uligne$, because it does not consist only of a
support (e.g., ``the support of the current robot's motion is the line
University-TrainStation''), but also of the important feature of orientation,
which allows, in the particular case of robot navigation, of representing, in
addition of the motion's support, the motion's direction (e.g.,
``the current robot's motion is supported by, and is of the same orientation
as, the $\dligne$ University-TrainStation''). Then, thanks to the kind of
relations on the handled objects; the strength of the relations comes from
their two features, a rotational feature and a translational feature:
the former handles the RDFs of the represented objects, the latter their TDFs:
\begin{enumerate}
  \item The rotational feature allows for the representation of statements
    such as
    ``parallel to, and of same/opposite orientation as'', or
    ``cuts, and to the left/right of''.
    As we saw, this feature corresponds to what Isli and Cohn's RA $\atra$
    \cite{Isli98a,Isli00b} can express.
  \item The translational feature
    allows for the representation of statements such as
    ``$\ell _1$ is parallel to, and lies strictly to the left of'', or
    ``$\ell _2$ cuts $\ell _1$ before $\ell _3$ does'', or (and this is an
    important disjunctive relation!)
    ``$\ell _3$ is parallel to both, and does not lie between, $\ell _1$ and
    $\ell _2$''. This last statement is represented as follows:\footnote{The
    representation of the other statements is left to the reader.}$^,$\footnote{$\ell _1$
    and $\ell _2$ may coincide, in which case the statement
    ``$\ell _3$ is parallel to both, and does not lie between, $\ell _1$ and $\ell _2$''
    is synonymous of
    ``$\ell _3$ is parallel to both, and does not coincide with, $\ell _1$ and $\ell _2$'',
    represented by the subformula
    $\{\pp _{c0},\pp _{c2}\}(\ell _1,\ell _2,\ell _3)$.}
    \begin{eqnarray}
    \{\pp _{l0},\pp _{l4},\pp _{c0},\pp _{c2},\pp _{r0},\pp _{r4}\}(\ell _1,\ell _2,\ell _3)\nonumber
    \end{eqnarray}
\end{enumerate}
We discuss below some of the potential application areas of the presented work.
\begin{figure*}[h]
\begin{center}
\setlength{\unitlength}{2486sp}%
\begingroup\makeatletter\ifx\SetFigFont\undefined%
\gdef\SetFigFont#1#2#3#4#5{%
  \reset@font\fontsize{#1}{#2pt}%
  \fontfamily{#3}\fontseries{#4}\fontshape{#5}%
  \selectfont}%
\fi\endgroup%
\begin{picture}(9249,6360)(889,-6181)
\thinlines
\special{ps: gsave 0 0 0 setrgbcolor}\put(1126,-2086){\vector( 1, 1){1912.500}}%here ...
\special{ps: grestore}\special{ps: gsave 0 0 0 setrgbcolor}\put(2926,-2086){\vector(-3, 4){1539}}%here ...
\special{ps: grestore}\special{ps: gsave 0 0 0 setrgbcolor}\put(901,-5686){\line( 2, 1){5130}}%here ...
\special{ps: grestore}\special{ps: gsave 0 0 0 setrgbcolor}\put(5176,-5686){\line( 0, 1){2700}}%here ...
\special{ps: grestore}\special{ps: gsave 0 0 0 setrgbcolor}\put(4051,-5686){\line( 0, 1){2700}}%here ...
\special{ps: grestore}\special{ps: gsave 0 0 0 setrgbcolor}\put(1801,-5686){\line( 0, 1){2700}}%here ...
\special{ps: grestore}\special{ps: gsave 0 0 0 setrgbcolor}\put(6526,-5686){\line( 3, 2){3634.615}}%here ...
\special{ps: grestore}\special{ps: gsave 0 0 0 setrgbcolor}\put(6526,-5461){\line( 1, 0){3600}}%here ...
\special{ps: grestore}\special{ps: gsave 0 0 0 setrgbcolor}\put(9901,-5686){\line(-1, 1){2452.500}}%here ...
\special{ps: grestore}\put(1846,-2356){\makebox(0,0)[lb]{\smash{\SetFigFont{8}{9.6}{\rmdefault}{\mddefault}{\updefault}\special{ps: gsave 0 0 0 setrgbcolor}(a)\special{ps: grestore}}}}
\put(2926,-6181){\makebox(0,0)[lb]{\smash{\SetFigFont{8}{9.6}{\rmdefault}{\mddefault}{\updefault}\special{ps: gsave 0 0 0 setrgbcolor}(b)\special{ps: grestore}}}}
\put(8011,-6181){\makebox(0,0)[lb]{\smash{\SetFigFont{8}{9.6}{\rmdefault}{\mddefault}{\updefault}\special{ps: gsave 0 0 0 setrgbcolor}(c)\special{ps: grestore}}}}
\put(2926,-16){\makebox(0,0)[lb]{\smash{\SetFigFont{8}{9.6}{\rmdefault}{\mddefault}{\updefault}\special{ps: gsave 0 0 0 setrgbcolor}$\ell _1$\special{ps: grestore}}}}
\put(1351,-61){\makebox(0,0)[lb]{\smash{\SetFigFont{8}{9.6}{\rmdefault}{\mddefault}{\updefault}\special{ps: gsave 0 0 0 setrgbcolor}$\ell _2$\special{ps: grestore}}}}
\put(2161,-1096){\makebox(0,0)[lb]{\smash{\SetFigFont{8}{9.6}{\rmdefault}{\mddefault}{\updefault}\special{ps: gsave 0 0 0 setrgbcolor}$P$\special{ps: grestore}}}}
\put(1801,-2986){\makebox(0,0)[lb]{\smash{\SetFigFont{8}{9.6}{\rmdefault}{\mddefault}{\updefault}\special{ps: gsave 0 0 0 setrgbcolor}$\ell _c$\special{ps: grestore}}}}
\put(1846,-5371){\makebox(0,0)[lb]{\smash{\SetFigFont{8}{9.6}{\rmdefault}{\mddefault}{\updefault}\special{ps: gsave 0 0 0 setrgbcolor}$P_3$\special{ps: grestore}}}}
\put(4141,-4291){\makebox(0,0)[lb]{\smash{\SetFigFont{8}{9.6}{\rmdefault}{\mddefault}{\updefault}\special{ps: gsave 0 0 0 setrgbcolor}$P_2$\special{ps: grestore}}}}
\put(4051,-2986){\makebox(0,0)[lb]{\smash{\SetFigFont{8}{9.6}{\rmdefault}{\mddefault}{\updefault}\special{ps: gsave 0 0 0 setrgbcolor}$\ell _b$\special{ps: grestore}}}}
\put(5176,-2986){\makebox(0,0)[lb]{\smash{\SetFigFont{8}{9.6}{\rmdefault}{\mddefault}{\updefault}\special{ps: gsave 0 0 0 setrgbcolor}$\ell _a$\special{ps: grestore}}}}
\put(6031,-3166){\makebox(0,0)[lb]{\smash{\SetFigFont{8}{9.6}{\rmdefault}{\mddefault}{\updefault}\special{ps: gsave 0 0 0 setrgbcolor}$\ell _d$\special{ps: grestore}}}}
\put(7516,-3211){\makebox(0,0)[lb]{\smash{\SetFigFont{8}{9.6}{\rmdefault}{\mddefault}{\updefault}\special{ps: gsave 0 0 0 setrgbcolor}$\ell _c$\special{ps: grestore}}}}
\put(10081,-3166){\makebox(0,0)[lb]{\smash{\SetFigFont{8}{9.6}{\rmdefault}{\mddefault}{\updefault}\special{ps: gsave 0 0 0 setrgbcolor}$\ell _b$\special{ps: grestore}}}}
\put(8641,-4381){\makebox(0,0)[lb]{\smash{\SetFigFont{8}{9.6}{\rmdefault}{\mddefault}{\updefault}\special{ps: gsave 0 0 0 setrgbcolor}$P_3$\special{ps: grestore}}}}
\put(7021,-5461){\makebox(0,0)[lb]{\smash{\SetFigFont{8}{9.6}{\rmdefault}{\mddefault}{\updefault}\special{ps: gsave 0 0 0 setrgbcolor}$P_1$\special{ps: grestore}}}}
\put(9631,-5461){\makebox(0,0)[lb]{\smash{\SetFigFont{8}{9.6}{\rmdefault}{\mddefault}{\updefault}\special{ps: gsave 0 0 0 setrgbcolor}$P_2$\special{ps: grestore}}}}
\put(10081,-5461){\makebox(0,0)[lb]{\smash{\SetFigFont{8}{9.6}{\rmdefault}{\mddefault}{\updefault}\special{ps: gsave 0 0 0 setrgbcolor}$\ell _a$\special{ps: grestore}}}}
\put(5221,-3751){\makebox(0,0)[lb]{\smash{\SetFigFont{8}{9.6}{\rmdefault}{\mddefault}{\updefault}\special{ps: gsave 0 0 0 setrgbcolor}$P_1$\special{ps: grestore}}}}
\end{picture}
\end{center}
\caption{Illustration of the use of the RA $\dlalg$ ---incidence geometry.}\label{illust-in-geo}
\end{figure*}
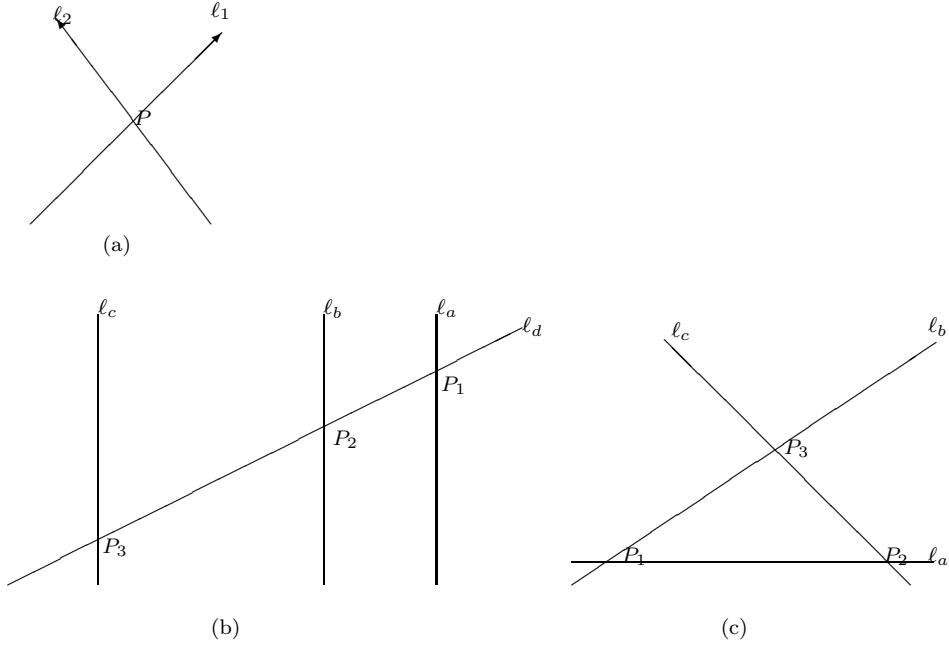
\subsection{Incidence geometry}\label{IncGeometry}
2D incidence geometry \cite{Balbiani94a} deals with the universe of points and (directed) lines.
Incidence (of a point with a line), betweenness (of three points, but also betweenness of three
parallel $\dlignes$\footnote{A $\dligne$ $\ell _2$ is between $\dlignes$ $\ell _1$ and
$\ell _3$ if and only if $\ell _2$ is parallel to both, and lies between, $\ell _1$ and
$\ell _3$.}), and non-collinearity (of three points) are easily representable in the RA
$\dlalg$:
\begin{enumerate}
  \item\label{IncGeometryItem1} A point $P$ will be considered as the intersection of two $\dlignes$ $\ell _1$ and
    $\ell _2$, such that $\cuts (\ell _2,\ell _1)$ and $l(\ell _2,\ell _1)$
    ---$\ell _1$ and $\ell _2$ are cutting $\dlignes$ and $\ell _2$ is
    to the left of $\ell _1$ (see Figure \ref{illust-in-geo}(a)). Transforming the conjunction
    $\cuts (\ell _2,\ell _1)\wedge l(\ell _2,\ell _1)$ into the RA $\dlalg$,
    we get $\langle\cp _c,\lre\rangle (\ell _1,\ell _2,\ell _1)$. We refer
    to the pair $(\ell _1,\ell _2)$ as the $\dlalg$ representation of $P$, and
    denote it by $\psi (P)$: $\psi (P)=(\ell _1,\ell _2)$.
  \item Let $P$ be a point such that $\psi (P)=(\ell _1,\ell _2)$, and $\ell$ a
    $\dligne$. Incidence of $P$ with $\ell$, $\incidence (P,\ell )$, is represented in  $\dlalg$ as
    $\{\cc _=,\cp _c,\pc _c\}(\ell ,\ell _1,\ell _2)$, saying that the three
    $\dlignes$ $\ell$, $\ell _1$ and $\ell _2$ are concurrent.
  \item Let $P_1$, $P_2$ and $P_3$ be three points such that
    $\psi (P_1)=(\ell _1,\ell _2)$, $\psi (P_2)=(\ell _3,\ell _4)$ and
    $\psi (P_3)=(\ell _5,\ell _6)$. $P_2$ is between $P_1$ and $P_3$
    can be represented using four $\dlignes$
    $\ell _a,\ell _b,\ell _c,\ell _d$ on which we impose the constraints that
    (see Figure \ref{illust-in-geo}(b) for illustration):\footnote{We
      represent here large betweenness of $P_1$, $P_2$ and $P_3$, in the sense that $P_2$
      coincides with $P_1$, lies strictly between $P_1$ and $P_3$, or coincides with $P_3$.}
    \begin{enumerate}
      \item\label{statement1} $\ell _b$ is parallel to both, and lies between, $\ell _a$ and $\ell _c$; and
      \item\label{statement2} $\ell _d$ cuts $\ell _a$ at $P_1$, $\ell _b$ at $P_2$ and $\ell _c$ at $P_3$.
    \end{enumerate}
    Statement \ref{statement1} defines betweenness of parallel $\dlignes$, and is represented as
    $\btwdl (\ell _a,\ell _b,\ell _c)\equiv\{\pp _{l0},\pp _{l1},\pp _{c0},\pp _{c1},\pp _{c2},\pp _{r3},
    \pp _{r4}\}(\ell _a,\ell _b,\ell _c)$,
    which splits into:
    \begin{enumerate}
      \item[$\bullet$] $\{\pp _{l0},\pp _{l1}\}(\ell _a,\ell _b,\ell _c)$, corresponding to $\ell _b$ being
        strictly to the left of $\ell _a$;
      \item[$\bullet$] $\{\pp _{c0},\pp _{c1},\pp _{c2}\}(\ell _a,\ell _b,\ell _c)$, corresponding to
        $\ell _b$ coinciding with $\ell _a$; and
      \item[$\bullet$] $\{\pp _{r3},\pp _{r4}\}(\ell _a,\ell _b,\ell _c)$, corresponding to $\ell _b$ being strictly to the
        right of $\ell _a$.
    \end{enumerate}
    Statement \ref{statement2} is represented using the previous point on incidence of a point with a line.
    Namely:
    \begin{enumerate}
      \item[$\bullet$] the substatement ``$\ell _d$ cuts $\ell _a$ at $P_1$'' is represented as
        \begin{eqnarray}
        \incidence (P_1,\ell _a)\wedge\incidence (P_1,\ell _d)&\equiv &
                         \{\cc _=,\cp _c,\pc _c\}(\ell _a,\ell _1,\ell _2)\wedge\nonumber  \\
                         &&\{\cc _=,\cp _c,\pc _c\}(\ell _d,\ell _1,\ell _2);\nonumber
        \end{eqnarray}
      \item[$\bullet$] the substatement ``$\ell _d$ cuts $\ell _b$ at $P_2$'' as
        \begin{eqnarray}
        \incidence (P_2,\ell _b)\wedge\incidence (P_2,\ell _d)&\equiv &
                         \{\cc _=,\cp _c,\pc _c\}(\ell _b,\ell _3,\ell _4)\wedge\nonumber  \\
                         &&\{\cc _=,\cp _c,\pc _c\}(\ell _d,\ell _3,\ell _4);\nonumber
        \end{eqnarray} and
      \item[$\bullet$] the substatement ``$\ell _d$ cuts $\ell _c$ at $P_3$'' as
        \begin{eqnarray}
        \incidence (P_3,\ell _c)\wedge\incidence (P_3,\ell _d)&\equiv &
                         \{\cc _=,\cp _c,\pc _c\}(\ell _c,\ell _5,\ell _6)\wedge\nonumber  \\
                         &&\{\cc _=,\cp _c,\pc _c\}(\ell _d,\ell _5,\ell _6).\nonumber
        \end{eqnarray}
    \end{enumerate}
    Putting everything together, betweenness of $P_1$, $P_2$ and $P_3$, $\btwp (P_1,P_2,P_3)$, is represented
    as follows:
    \begin{eqnarray}
    \btwp (P_1,P_2,P_3)&\equiv &
    \{\pp _{l0},\pp _{l1},\pp _{c0},\pp _{c1},\pp _{c2},\pp _{r3},\pp _{r4}\}(\ell _a,\ell _b,\ell _c)\wedge\nonumber  \\
    &&\{\cc _=,\cp _c,\pc _c\}(\ell _a,\ell _1,\ell _2)\wedge
                         \{\cc _=,\cp _c,\pc _c\}(\ell _d,\ell _1,\ell _2)\wedge\nonumber  \\
    &&\{\cc _=,\cp _c,\pc _c\}(\ell _b,\ell _3,\ell _4)\wedge
                         \{\cc _=,\cp _c,\pc _c\}(\ell _d,\ell _3,\ell _4)\wedge\nonumber  \\
    &&\{\cc _=,\cp _c,\pc _c\}(\ell _c,\ell _5,\ell _6)\wedge
                         \{\cc _=,\cp _c,\pc _c\}(\ell _d,\ell _5,\ell _6)\nonumber
    \end{eqnarray}
  \item Let $P_1$, $P_2$ and $P_3$ as in the previous point:
    $\psi (P_1)=(\ell _1,\ell _2)$, $\psi (P_2)=(\ell _3,\ell _4)$ and
    $\psi (P_3)=(\ell _5,\ell _6)$. Non-collinearity of the three points,
    $\noncoll (P_1,P_2,P_3)$,
    can be represented using three $\dlignes$
    $\ell _a,\ell _b,\ell _c$ on which we impose the constraints that
    (see Figure \ref{illust-in-geo}(c) for illustration):
    \begin{enumerate}
      \item $\ell _b$ and $\ell _c$ both cut $\ell _a$, but at distinct points;
      \item $P_1$ is incident with each of $\ell _a$ and $\ell _b$, $P_2$ with each of
        $\ell _a$ and $\ell _c$, and $P_3$ with each of $\ell _b$ and $\ell _c$.
    \end{enumerate}
    We get:
    \begin{eqnarray}
    \noncoll (P_1,P_2,P_3)&\equiv &\{\cc _<,\cc _>\}(\ell _a,\ell _b,\ell _c)\wedge\nonumber  \\
    &&\incidence (P_1,\ell _a)\wedge\incidence (P_1,\ell _b)\wedge\nonumber  \\
    &&\incidence (P_2,\ell _a)\wedge\incidence (P_2,\ell _c)\wedge\nonumber  \\
    &&\incidence (P_3,\ell _b)\wedge\incidence (P_3,\ell _c)\nonumber
    \end{eqnarray}
    Translating the incidence relation into the RA $\dlalg$, we get:
    \begin{eqnarray}
    \noncoll (P_1,P_2,P_3)&\equiv &\{\cc _<,\cc _>\}(\ell _a,\ell _b,\ell _c)\wedge\nonumber  \\
    &&\{\cc _=,\cp _c,\pc _c\}(\ell _a,\ell _1,\ell _2)\wedge
                         \{\cc _=,\cp _c,\pc _c\}(\ell _b,\ell _1,\ell _2)\wedge\nonumber  \\
    &&\{\cc _=,\cp _c,\pc _c\}(\ell _a,\ell _3,\ell _4)\wedge
                         \{\cc _=,\cp _c,\pc _c\}(\ell _c,\ell _3,\ell _4)\wedge\nonumber  \\
    &&\{\cc _=,\cp _c,\pc _c\}(\ell _b,\ell _5,\ell _6)\wedge
                         \{\cc _=,\cp _c,\pc _c\}(\ell _c,\ell _5,\ell _6)\nonumber
    \end{eqnarray}
\end{enumerate}
\subsection{Geographical Information Systems}\label{gissubsection}
The objects manipulated by $\gis$ applications are mainly points,
segments and proper polygons (more than two sides) of the 2-dimensional space.
A general (for instance, concave) polygon can always be decomposed into a union
of conex polygons ---see, for instance, the work in \cite{Bennett98a}, where a
system answering queries on the RCC-8 \cite{Randell92a} relation between two
input (polygonal) regions of a geographical database is defined.

In order to use the RA $\dlalg$ to reason about polygons, we need to
provide a representation of convex polygons, not in terms of an ordered, say
anticlockwise, list of vertices, of the form $(X_1,\ldots ,X_n)$,
$n\geq 1$, but in terms of an ordered, anticlockwise, list of $\dlignes$, of
the form $(\ell _1,\ldots ,\ell _n)$, $n\geq 1$.\footnote{A ordered,
anticlockwise, list of $\dlignes$ is a list
$(\ell _1,\ldots ,\ell _n)$, $n\geq 1$, of $\dlignes$, such that the
list $(\ell _1',\ldots ,\ell _n')$, obtained by translating each of the
$\ell _i$'s so that it contains a fixed point $O$, verifies the following: the
positive half-lines of $\ell _1',\ldots ,\ell _n'$ bounded by $O$ are met in
that order when scanning a circle, say ${\cal C}$, centered at $O$, starting from the
intersecting point of ${\cal C}$ with the positive half-line of $\ell _1'$
bounded by $O$.}
Reasoning about a collection of convex polygons transfoms then into reasoning about the
$\dlignes$ in the representations of the different convex polygons in the collection.

\begin{figure*}
\begin{center}
\centerline{\epsffile{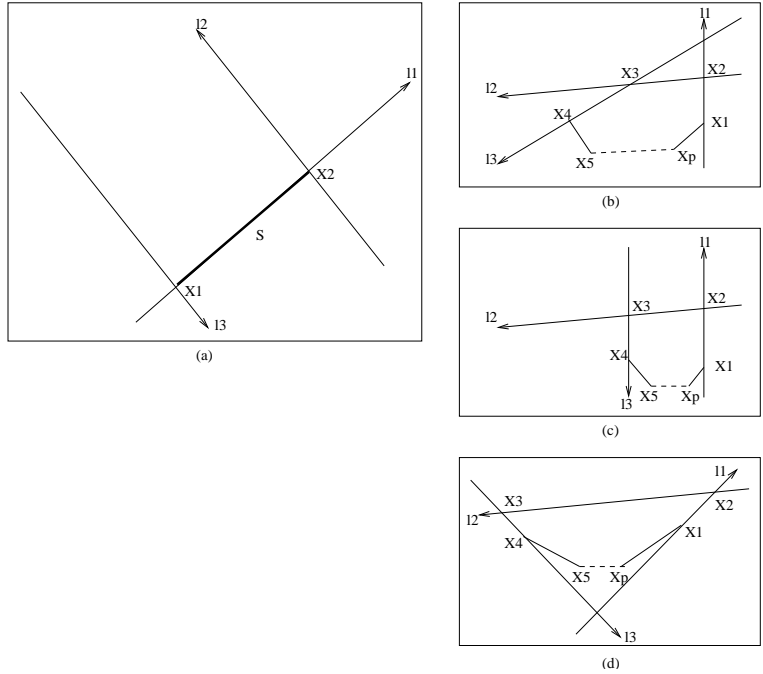}}
\end{center}
\caption{(Left) a (directed) line segment $S=(X_1,X_2)$ is represented in the RA
$\dlalg$ using three $\dlignes$ $\ell _1$, $\ell _2$ and $\ell _3$, such that
$\ell _2$ is to the left of, and $\ell _3$ to the right of, $\ell _1$; and
$\ell _3$ is parallel to, and lies within the left half-plane bounded by, $\ell _2$
($X_1$ is then the intersecting point of $\ell _1$ with $\ell _3$; and $X_2$ the
intersecting point of $\ell _1$ with $\ell _2$):
    $\{\langle\cc _>,\lor\rangle\}(\ell _1,\ell _2,\ell _3)$.
  (Right) a $p$-vertex convex polygon, given as an ordered, anticlockwise, list
  $(X_1,X_2,X_3,\ldots ,X_p)$ of $p$ vertices, $p\geq 3$, is represented in the RA $\dlalg$ as
  an ordered, anticlockwise, list $(\ell _1,\ell _2,\ell _3,\ldots ,\ell _p)$ of $p$
  $\dlignes$, such that every three consecutive $\dlignes$ $\ell _i$, $\ell _{i+_p1}$ and
  $\ell _{i+_p2}$ verify
  $\{\langle\cc _<,\rll\rangle (\ell _{i+_p1},\ell _i,\ell _{i+_p2})$ (see illustration (b) for $i=1$),
  $\langle\cc _<,\rol\rangle (\ell _{i+_p1},\ell _i,\ell _{i+_p2})$  (see illustration (c) for $i=1$) or
  $\langle\cc _<,\rrl\rangle (\ell _{i+_p1},\ell _i,\ell _{i+_p2})$ (see illustration (d) for $i=1$):
  $\{\langle\cc _<,\rll\rangle ,
   \langle\cc _<,\rol\rangle ,
   \langle\cc _<,\rrl\rangle\}(\ell _{i+_p1},\ell _i,\ell _{i+_p2})$.}\label{illust-polyg}
\end{figure*}

Points and (directed) line segments are special cases of convex polygons:
\begin{enumerate}
  \item We have already seen how to represent a point $P$ as the intersection
    of two $\dlignes$ $\ell _1$ and $\ell _2$, such that $\cuts (\ell _2,\ell _1)$
    and $l(\ell _2,\ell _1)$, which transforms into the RA $\dlalg$ as
    $\{\langle\cp _c,\lre\rangle\}(\ell _1,\ell _2,\ell _1)$. The pair $(\ell _1,\ell _2)$
    is referred to as the $\dlalg$ representation of $P$, denoted $\psi (P)$:
    $\psi (P)=(\ell _1,\ell _2)$ ---see Figure \ref{illust-in-geo}(a).
  \item\label{gisitem2} A segment $S=(X_1,X_2)$ will be represented using three $\dlignes$
    $\ell _1$, $\ell _2$ and $\ell _3$, such that $\ell _2$ and $\ell _3$ are parallel
    to each other; $\ell _3$ lies within the left half-plane bounded by $\ell _2$; and
    $\ell _2$ is to the left of, and $\ell _3$ to the right of, $\ell _1$  (the
    segment is then the part of $\ell _1$ between the intersecting points with the
    other two $\dlignes$, oriented from the intersecting point with $\ell _3$ to the
    intersecting point with $\ell _2$). An illustration is provided in Figure
    \ref{illust-polyg}(a). We get the following:
    $\{\langle\cc _>,\lor\rangle\} (\ell _1,\ell _2,\ell _3)$. We refer to
    the triple $(\ell _1,\ell _2,\ell _3)$ as the $\dlalg$ representation of $S$, which
    we denote by $\psi (S)$:
    $\psi (S)=(\ell _1,\ell _2,\ell _3)$.
\end{enumerate}
A convex polygon with $p$ vertices, with $p\geq 3$, will be represented as the $p$-tuple
    $(\ell _1,\ell _2,\ldots ,\ell _p)$ of $\dlignes$, such that, for all
    $i=1\ldots p$, the $\dlignes$ $\ell _i$, $\ell _{i+_p1}$ and $\ell _{i+_p2}$ verify the following:
    \begin{enumerate}
      \item\label{point1} $\ell _i$ and $\ell _{i+_p2}$ both cut $\ell _{i+_p1}$, but $\ell _i$ does
        it before $\ell _{i+_p2}$;
      \item\label{point2} $\ell _{i+_p1}$ is to the left of $\ell _i$; and
      \item\label{point3} $\ell _{i+_p2}$ is to the left of $\ell _{i+_p1}$,
    \end{enumerate}
    where $+_p$ is cyclic addition over the set $\{1,\ldots ,p\}$; i.e.
$$i+_p1=\left\{\begin{array}{l}
                            i+1\mbox{ if }i\leq p-1,\\
                            1\mbox{ if }i=p
                          \end{array}
                   \right.\hskip 1cm
        i+_p2=\left\{\begin{array}{l}
                            i+2\mbox{ if }i\leq p-2,\\
                            1\mbox{ if }i=p-1,\\
                            2\mbox{ if }i=p
                          \end{array}
                   \right.
        $$
The polygon is then the contour of the surface consisting of the intersection of the $p$ left
half-planes delimited by the $\dlignes$ $\ell _1, \ldots ,\ell _p$.
The conjunction of the three points \ref{point1}, \ref{point2} and \ref{point3} just above
translates into the RA $\dlalg$ as follows:
\begin{eqnarray}
\{\langle\cc _<,\rll\rangle ,
   \langle\cc _<,\rol\rangle ,
   \langle\cc _<,\rrl\rangle\}(\ell _{i+_p1},\ell _i,\ell _{i+_p2}) \nonumber
\end{eqnarray}
---see the illustration of Figure \ref{illust-polyg}(b-c-d).
\subsection{(Polygonal) shape representation}
In shape
representation, the shapes dealt with are mostly polygonal; when they are not, they
are generally given polygonal approximations (a circle, for instance, can
be so approximated).
\subsubsection*{Example 1}
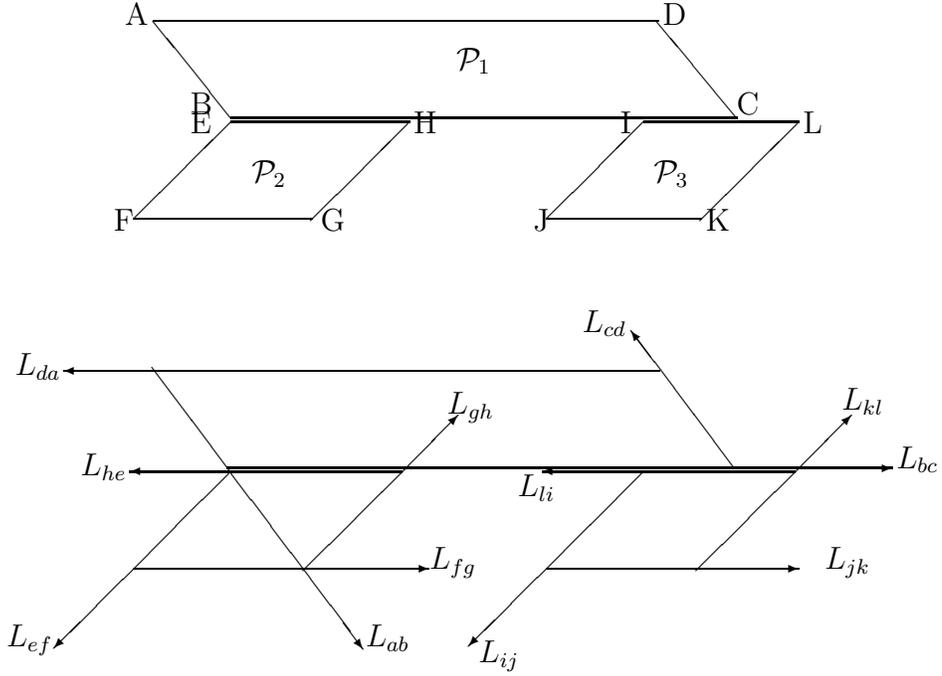
\begin{figure*}
\begin{center}
\setlength{\unitlength}{2144sp}%
\begingroup\makeatletter\ifx\SetFigFont\undefined%
\gdef\SetFigFont#1#2#3#4#5{%
  \reset@font\fontsize{#1}{#2pt}%
  \fontfamily{#3}\fontseries{#4}\fontshape{#5}%
  \selectfont}%
\fi\endgroup%
\begin{picture}(9990,7569)(856,-7270)
\thinlines
\special{ps: gsave 0 0 0 setrgbcolor}\put(6796,-2086){\line( 1, 1){1125}}
\special{ps: grestore}\special{ps: gsave 0 0 0 setrgbcolor}\put(8573,-2108){\line( 1, 1){1125}}
\special{ps: grestore}\special{ps: gsave 0 0 0 setrgbcolor}\put(7921,-961){\line( 1, 0){1800}}
\special{ps: grestore}\special{ps: gsave 0 0 0 setrgbcolor}\put(6796,-2086){\line( 1, 0){1800}}
\special{ps: grestore}\special{ps: gsave 0 0 0 setrgbcolor}\put(2026,-2086){\line( 1, 1){1125}}
\special{ps: grestore}\special{ps: gsave 0 0 0 setrgbcolor}\put(4073,-2108){\line( 1, 1){1125}}
\special{ps: grestore}\special{ps: gsave 0 0 0 setrgbcolor}\put(3151,-961){\line( 1, 0){2070}}
\special{ps: grestore}\special{ps: gsave 0 0 0 setrgbcolor}\put(2026,-2086){\line( 1, 0){2070}}
\special{ps: grestore}\special{ps: gsave 0 0 0 setrgbcolor}\multiput(4096,-2086)(-75.24590,90.29508){13}{\line(-5, 6){ 37.623}}
\special{ps: grestore}\special{ps: gsave 0 0 0 setrgbcolor}\put(3151,-916){\line( 1, 0){5850}}
\special{ps: grestore}\special{ps: gsave 0 0 0 setrgbcolor}\put(2251,209){\line( 1, 0){5850}}
\special{ps: grestore}\special{ps: gsave 0 0 0 setrgbcolor}\put(9001,-916){\line(-5, 6){940.574}}
\special{ps: grestore}\special{ps: gsave 0 0 0 setrgbcolor}\put(3151,-916){\line(-4, 5){900}}
\special{ps: grestore}\put(7651,-1096){\makebox(0,0)[lb]{\smash{\SetFigFont{12}{14.4}{\rmdefault}{\mddefault}{\updefault}\special{ps: gsave 0 0 0 setrgbcolor}I\special{ps: grestore}}}}
\put(1926,164){\makebox(0,0)[lb]{\smash{\SetFigFont{12}{14.4}{\rmdefault}{\mddefault}{\updefault}\special{ps: gsave 0 0 0 setrgbcolor}A\special{ps: grestore}}}}
\put(8146,164){\makebox(0,0)[lb]{\smash{\SetFigFont{12}{14.4}{\rmdefault}{\mddefault}{\updefault}\special{ps: gsave 0 0 0 setrgbcolor}D\special{ps: grestore}}}}
\put(9001,-871){\makebox(0,0)[lb]{\smash{\SetFigFont{12}{14.4}{\rmdefault}{\mddefault}{\updefault}\special{ps: gsave 0 0 0 setrgbcolor}C\special{ps: grestore}}}}
\put(2681,-871){\makebox(0,0)[lb]{\smash{\SetFigFont{12}{14.4}{\rmdefault}{\mddefault}{\updefault}\special{ps: gsave 0 0 0 setrgbcolor}B\special{ps: grestore}}}}
\put(5641,-353.50){\makebox(0,0)[lb]{\smash{\SetFigFont{12}{14.4}{\rmdefault}{\mddefault}{\updefault}
	\special{ps: gsave 0 0 0 setrgbcolor}${\cal P}_1$\special{ps: grestore}}}}
\put(2681,-1096){\makebox(0,0)[lb]{\smash{\SetFigFont{12}{14.4}{\rmdefault}{\mddefault}{\updefault}\special{ps: gsave 0 0 0 setrgbcolor}E\special{ps: grestore}}}}
\put(5266,-1096){\makebox(0,0)[lb]{\smash{\SetFigFont{12}{14.4}{\rmdefault}{\mddefault}{\updefault}\special{ps: gsave 0 0 0 setrgbcolor}H\special{ps: grestore}}}}
\put(4191,-2221){\makebox(0,0)[lb]{\smash{\SetFigFont{12}{14.4}{\rmdefault}{\mddefault}{\updefault}\special{ps: gsave 0 0 0 setrgbcolor}G\special{ps: grestore}}}}
\put(3393.50,-1658.50){\makebox(0,0)[lb]{\smash{\SetFigFont{12}{14.4}{\rmdefault}{\mddefault}{\updefault}\special{ps: gsave 0 0 0 setrgbcolor}${\cal P}_2$\special{ps: grestore}}}}
\put(8041,-1658.50){\makebox(0,0)[lb]{\smash{\SetFigFont{12}{14.4}{\rmdefault}{\mddefault}{\updefault}\special{ps: gsave 0 0 0 setrgbcolor}${\cal P}_3$\special{ps: grestore}}}}
\put(1796,-2221){\makebox(0,0)[lb]{\smash{\SetFigFont{12}{14.4}{\rmdefault}{\mddefault}{\updefault}\special{ps: gsave 0 0 0 setrgbcolor}F\special{ps: grestore}}}}
\put(6651,-2221){\makebox(0,0)[lb]{\smash{\SetFigFont{12}{14.4}{\rmdefault}{\mddefault}{\updefault}\special{ps: gsave 0 0 0 setrgbcolor}J\special{ps: grestore}}}}
\put(8641,-2221){\makebox(0,0)[lb]{\smash{\SetFigFont{12}{14.4}{\rmdefault}{\mddefault}{\updefault}\special{ps: gsave 0 0 0 setrgbcolor}K\special{ps: grestore}}}}
\put(9766,-1096){\makebox(0,0)[lb]{\smash{\SetFigFont{12}{14.4}{\rmdefault}{\mddefault}{\updefault}\special{ps: gsave 0 0 0 setrgbcolor}L\special{ps: grestore}}}}
\special{ps: gsave 0 0 0 setrgbcolor}\put(8965,-4958){\vector(-3, 4){1188}}
\special{ps: grestore}\special{ps: gsave 0 0 0 setrgbcolor}\put(3106,-4966){\vector( 1, 0){7695}}
\special{ps: grestore}\special{ps: gsave 0 0 0 setrgbcolor}\put(9676,-5011){\vector(-1, 0){2925}}
\special{ps: grestore}\special{ps: gsave 0 0 0 setrgbcolor}\put(6796,-6136){\vector( 1, 0){2925}}
\special{ps: grestore}\special{ps: gsave 0 0 0 setrgbcolor}\put(3128,-5033){\vector(-1,-1){2025}}
\special{ps: grestore}\special{ps: gsave 0 0 0 setrgbcolor}\put(2026,-6136){\vector( 1, 0){3420}}
\special{ps: grestore}\special{ps: gsave 0 0 0 setrgbcolor}\put(5131,-5011){\vector(-1, 0){3150}}
\special{ps: grestore}\special{ps: gsave 0 0 0 setrgbcolor}\put(8101,-3841){\vector(-1, 0){6885}}
\special{ps: grestore}\special{ps: gsave 0 0 0 setrgbcolor}\put(3983,-6158){\vector( 1, 1){1800}}
\special{ps: grestore}\special{ps: gsave 0 0 0 setrgbcolor}\put(2241,-3805){\vector( 3,-4){2440.800}}
\special{ps: grestore}\special{ps: gsave 0 0 0 setrgbcolor}\put(7921,-5011){\vector(-1,-1){2025}}
\special{ps: grestore}\special{ps: gsave 0 0 0 setrgbcolor}\put(8528,-6158){\vector( 1, 1){1800}}
\special{ps: grestore}\put(656,-3886){\makebox(0,0)[lb]{\smash{\SetFigFont{12}{14.4}{\rmdefault}{\mddefault}{\updefault}\special{ps: gsave 0 0 0 setrgbcolor}$L_{da}$\special{ps: grestore}}}}
\put(1416,-5056){\makebox(0,0)[lb]{\smash{\SetFigFont{12}{14.4}{\rmdefault}{\mddefault}{\updefault}\special{ps: gsave 0 0 0 setrgbcolor}$L_{he}$\special{ps: grestore}}}}
\put(5651,-4336){\makebox(0,0)[lb]{\smash{\SetFigFont{12}{14.4}{\rmdefault}{\mddefault}{\updefault}\special{ps: gsave 0 0 0 setrgbcolor}$L_{gh}$\special{ps: grestore}}}}
\put(5446,-6136){\makebox(0,0)[lb]{\smash{\SetFigFont{12}{14.4}{\rmdefault}{\mddefault}{\updefault}\special{ps: gsave 0 0 0 setrgbcolor}$L_{fg}$\special{ps: grestore}}}}
\put(6016,-7216){\makebox(0,0)[lb]{\smash{\SetFigFont{12}{14.4}{\rmdefault}{\mddefault}{\updefault}\special{ps: gsave 0 0 0 setrgbcolor}$L_{ij}$\special{ps: grestore}}}}
\put(6471,-5291){\makebox(0,0)[lb]{\smash{\SetFigFont{12}{14.4}{\rmdefault}{\mddefault}{\updefault}\special{ps: gsave 0 0 0 setrgbcolor}$L_{li}$\special{ps: grestore}}}}
\put(7216,-3391){\makebox(0,0)[lb]{\smash{\SetFigFont{12}{14.4}{\rmdefault}{\mddefault}{\updefault}\special{ps: gsave 0 0 0 setrgbcolor}$L_{cd}$\special{ps: grestore}}}}
\put(10216,-4291){\makebox(0,0)[lb]{\smash{\SetFigFont{12}{14.4}{\rmdefault}{\mddefault}{\updefault}\special{ps: gsave 0 0 0 setrgbcolor}$L_{kl}$\special{ps: grestore}}}}
\put(10846,-4966){\makebox(0,0)[lb]{\smash{\SetFigFont{12}{14.4}{\rmdefault}{\mddefault}{\updefault}\special{ps: gsave 0 0 0 setrgbcolor}$L_{bc}$\special{ps: grestore}}}}
\put(10021,-6136){\makebox(0,0)[lb]{\smash{\SetFigFont{12}{14.4}{\rmdefault}{\mddefault}{\updefault}\special{ps: gsave 0 0 0 setrgbcolor}$L_{jk}$\special{ps: grestore}}}}
\put(4711,-7036){\makebox(0,0)[lb]{\smash{\SetFigFont{12}{14.4}{\rmdefault}{\mddefault}{\updefault}\special{ps: gsave 0 0 0 setrgbcolor}$L_{ab}$\special{ps: grestore}}}}
\put(556,-7036){\makebox(0,0)[lb]{\smash{\SetFigFont{12}{14.4}{\rmdefault}{\mddefault}{\updefault}\special{ps: gsave 0 0 0 setrgbcolor}$L_{ef}$\special{ps: grestore}}}}
\end{picture}
\end{center}
\caption{Shape representation ---example 1.}\label{shape-rep-eg}
\end{figure*}

To illustrate the use of the RA $\dlalg$ for polygonal shape
representation, we consider a first example illustrated by the
shape of Figure \ref{shape-rep-eg}(top), representing a table
composed of three parallelogram-like parts, ${\cal P}_1$,
${\cal P}_2$ and ${\cal P}_3$. The parts ${\cal P}_2$ and
${\cal P}_3$ constitute the base of the table, i.e., the part
reposing on the ground, and holding the top part, represented by
${\cal P}_1$. The side AB of the upper part is collinear with the
diagonal EG of part ${\cal P}_2$. The vertex C of part
${\cal P}_1$ comes strictly inside the side IL of part ${\cal P}_3$.
Finally, the non-horizontal sides of parts ${\cal P}_2$ and
${\cal P}_3$ are pairwise parallel.

With each side XY of the three table parts we associate a $\dligne$
$L_{xy}$, as indicated in Figure \ref{shape-rep-eg}(bottom).
Part ${\cal P}_1$ of the table is then the surface consisting of the intersection of the left
half-planes bounded by the $\dlignes$ $\ell _{ab}$, $\ell _{bc}$, $\ell _{cd}$ and $\ell _{da}$;
Part ${\cal P}_2$ is the surface consisting of the intersection of the left half-planes bounded
by the $\dlignes$ $\ell _{ef}$, $\ell _{fg}$, $\ell _{gh}$ and $\ell _{he}$; and
Part ${\cal P}_3$ is the surface consisting of the intersection of the left half-planes bounded
by the $\dlignes$ $\ell _{ij}$, $\ell _{jk}$, $\ell _{kl}$ and $\ell _{li}$.
\begin{enumerate}
  \item Part ${\cal P}_1$ reposes on parts ${\cal P}_2$ and ${\cal P}_3$, which means
    that (the supports of) the three $\dlignes$ $L_{bc}$, $L_{he}$ and $L_{li}$
    coincide:
    \begin{eqnarray}
      \{\langle\pp _{c1},\oeo\rangle\}(L_{bc},L_{he},L_{li}) \nonumber
    \end{eqnarray}
  \item The $\dlignes$ $L_{fg}$ and $L_{jk}$, which constitute the base of the table,
    coincide; and are both parallel to, and lie within the right open half-plane
    bounded by, $L_{bc}$:
    \begin{eqnarray}
      \{\langle\pp _{r3},\eee\rangle\}(L_{bc},L_{fg},L_{jk}) \nonumber
    \end{eqnarray}
  \item The $\dligne$ $L_{da}$ is parallel to, lies within the left open half-plane
    bounded by, and is of opposite orientation than, the $\dligne$ $L_{bc}$:
    \begin{eqnarray}
      \{\langle\pp _{l1},\oeo\rangle\}(L_{bc},L_{da},L_{da}) \nonumber
    \end{eqnarray}
  \item Parallelity of the non-horizontal sides of part ${\cal P}_1$ can be
    expressed thus:
    \begin{eqnarray}
      \{\langle\pp _{l1},\oeo\rangle\}(L_{ab},L_{cd},L_{cd}) \nonumber
    \end{eqnarray}
  \item Pairwise parallelity of the non-horizontal sides of parts ${\cal P}_2$
    and ${\cal P}_3$ can be expressed thus:
    \begin{eqnarray}
      \{\langle\pp _{l0},\ooe\rangle\}(L_{ef},L_{gh},L_{ij})\wedge
      \{\langle\pp _{r4},\ooe\rangle\}(L_{gh},L_{ij},L_{kl}) \nonumber
    \end{eqnarray}
  \item Collinearity of side AB with diagonal EG of part ${\cal P}_2$ can be
    expressed by concurrency of $\dlignes$ $L_{ab}$, $L_{bc}$ and $L_{ef}$, on
    the one hand, and concurrency of $\dlignes$ $L_{ab}$, $L_{fg}$ and
    $L_{gh}$, on the other hand:
    \begin{eqnarray}
      \{\langle\cc _{=},\lrr\rangle\}(L_{ab},L_{bc},L_{ef})\wedge
      \{\langle\cc _{=},\lll\rangle\}(L_{ab},L_{fg},L_{gh}) \nonumber
    \end{eqnarray}
  \item Finally, strict betweenness of vertices $I$, $C$ and $L$ ($C$ strictly between
    $I$ and $L$) is expressed by the conjuction of ``$L_{cd}$ cuts $L_{li}$
    before $L_{ij}$ does'' and ``$L_{cd}$ cuts $L_{li}$ after $L_{kl}$ does''.
    This translates into the RA $\dlalg$ as follows:
    \begin{eqnarray}
      \{\langle\cc _<,\rll\rangle\}(L_{li},L_{cd},L_{ij})\wedge
      \{\langle\cc _>,\rrr\rangle\}(L_{li},L_{cd},L_{kl}) \nonumber
    \end{eqnarray}
\end{enumerate}
\subsubsection*{Example 2}
\begin{figure*}
\begin{center}
\centerline{\epsffile{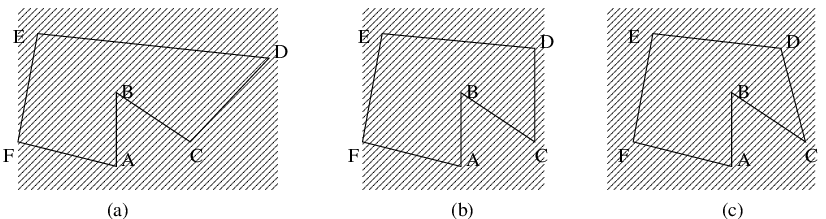}}
\end{center}
\caption{Shape representation ---example 2.}\label{shape-rep-eg2}
\end{figure*}

As a second example, we consider the three polygonal shapes of Figure
\ref{shape-rep-eg2}. Similarly to the previous example, we associate with each
side, say XY, a $\dligne$ $L_{xy}$. The $\dlalg$ algebra is able to
distinguish between the three shapes: for all triples
$(\ell _1,\ell _2,\ell _3)$ of $\dlignes$ not involving the
$\dligne$ $L_{CD}$, the $\dlalg$ relation on $(\ell _1,\ell _2,\ell _3)$ is
the same for all three shapes; the relation on any of the triples involving
the $\dligne$ $L_{CD}$, however, differs from any of the three shapes to any
of the other two. For instance, if we consider the triple
$(L_{ab},L_{bc},L_{cd})$, we get
the relation $\{\langle\cc _>,\rlr\rangle\}$ for the shape of Figure \ref{shape-rep-eg2}(a);
the relation $\{\langle\cp _r,\rle\rangle\}$ for the shape of Figure \ref{shape-rep-eg2}(b); and
the relation $\{\langle\cc _<,\rll\rangle\}$ for the shape of Figure \ref{shape-rep-eg2}(c).
\subsection{Localisation in multi-robot navigation}
Self-localisation of a robot, embedded in an environment with $n$ landmarks,
consists of describing the panorama of the robot w.r.t. the
landmarks; i.e., how the different landmarks are situated relative to one
another, as viewed from the current robot's position (the robot is supposed
equipped with a camera). The standard way of representing such a panorama
is to give the (cyclic) order in which the landmarks appear in a
360-degrees anticlockwise turn, starting, say, from landmark 1 (the landmarks
are supposed numbered from 1 to $n$). If we use the $\dlignes$ relating the
robot to the different landmarks, then the problem can be represented using
the RA $\atra$ \cite{Isli00b}, by providing for each triple
$(\ell _1,\ell _2,\ell _3)$ of the $\dlignes$ the $\atra$ relation it
satisfies. For instance, if the three $\dlignes$ $\ell _1$, $\ell _2$ and $\ell _3$
appear in that order in a 360-degrees anticlockwise turn about the robot's
location, the situation can be described using the $\atra$ relation
$\cyc =\{lrl,orl,rll,rol,$ $rrl,rro,rrr\}$, expressing anticlockwise betweenness:
$\cyc (\ell _1,\ell _2,\ell _3)$.\footnote{This can be easily checked using the
illustrations of the different $\atra$ atoms, given in Figure \ref{CycordsRelation}(a).}
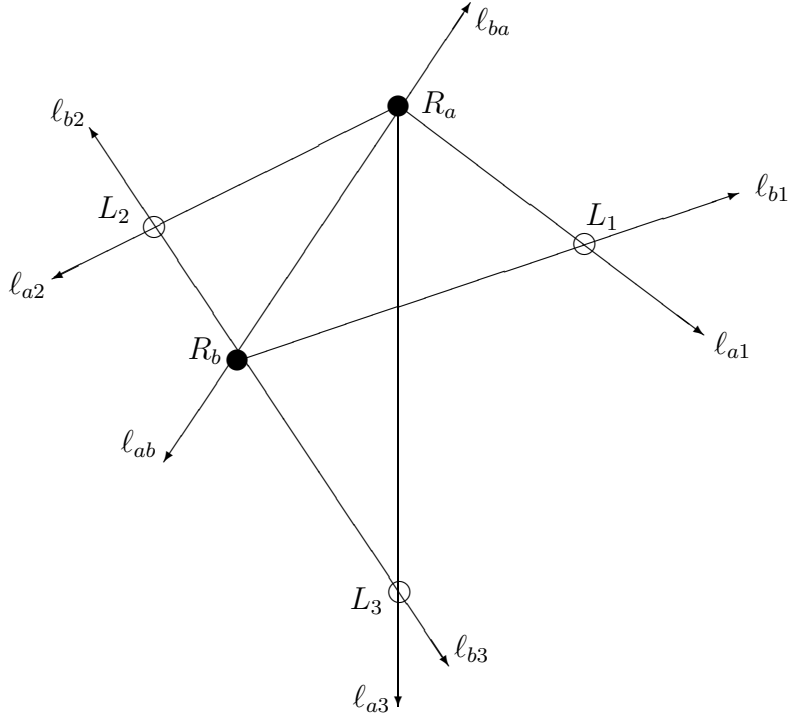
\begin{figure*}
\begin{center}
\setlength{\unitlength}{4144sp}%
\begingroup\makeatletter\ifx\SetFigFont\undefined%
\gdef\SetFigFont#1#2#3#4#5{%
  \reset@font\fontsize{#1}{#2pt}%
  \fontfamily{#3}\fontseries{#4}\fontshape{#5}%
  \selectfont}%
\fi\endgroup%
\begin{picture}(4467,4257)(136,-3943)
\thinlines
\put(1106,-1056){\circle{128}}%don't touch at this line...
\put(3681,-1162){\circle{128}}%don't touch at this line...
\put(2566,-331){\circle*{128}}%don't touch at this line...
\put(1602,-1851){\circle*{128}}%don't touch at this line...
\put(2574,-3244){\circle{128}}
\special{ps: gsave 0 0 0 setrgbcolor}\put(2566,-331){\vector(-2,-1){2070}}%don't touch at this line...
\special{ps: grestore}\special{ps: gsave 0 0 0 setrgbcolor}\put(2566,-331){\vector( 4,-3){1828.800}}%don't touch at this line...
\special{ps: grestore}\special{ps: gsave 0 0 0 setrgbcolor}\put(2566,-331){\vector( 0,-1){3600}}%don't touch at this line...
\special{ps: grestore}\special{ps: gsave 0 0 0 setrgbcolor}\put(1621,-1851){\vector( 3, 1){2983.500}}%don't touch at this line...
\special{ps: grestore}\special{ps: gsave 0 0 0 setrgbcolor}\put(721,-466){\vector(-2, 3){  0}}%don't touch at this line...
\put(721,-466){\vector( 2,-3){2146.154}}%don't touch at this line...
\special{ps: grestore}\special{ps: gsave 0 0 0 setrgbcolor}\put(1164,-2467){\vector(-2,-3){  0}}
\put(1164,-2467){\vector( 2, 3){1834.615}}
\special{ps: grestore}\put(2701,-376){\makebox(0,0)[lb]{\smash{\SetFigFont{12}{14.4}{\rmdefault}{\mddefault}{\updefault}\special{ps: gsave 0 0 0 setrgbcolor}$R_a$\special{ps: grestore}}}}
\put(3686,-1046){\makebox(0,0)[lb]{\smash{\SetFigFont{12}{14.4}{\rmdefault}{\mddefault}{\updefault}\special{ps: gsave 0 0 0 setrgbcolor}$L_1$\special{ps: grestore}}}}
\put(766,-1006){\makebox(0,0)[lb]{\smash{\SetFigFont{12}{14.4}{\rmdefault}{\mddefault}{\updefault}\special{ps: gsave 0 0 0 setrgbcolor}$L_2$\special{ps: grestore}}}}
\put(1306,-1846){\makebox(0,0)[lb]{\smash{\SetFigFont{12}{14.4}{\rmdefault}{\mddefault}{\updefault}\special{ps: gsave 0 0 0 setrgbcolor}$R_b$\special{ps: grestore}}}}
\put(3026,119){\makebox(0,0)[lb]{\smash{\SetFigFont{12}{14.4}{\rmdefault}{\mddefault}{\updefault}\special{ps: gsave 0 0 0 setrgbcolor}$\ell _{ba}$\special{ps: grestore}}}}%don't touch at this line...
\put(906,-2406){\makebox(0,0)[lb]{\smash{\SetFigFont{12}{14.4}{\rmdefault}{\mddefault}{\updefault}\special{ps: gsave 0 0 0 setrgbcolor}$\ell _{ab}$\special{ps: grestore}}}}%don't touch at this line...
\put(236,-1456){\makebox(0,0)[lb]{\smash{\SetFigFont{12}{14.4}{\rmdefault}{\mddefault}{\updefault}\special{ps: gsave 0 0 0 setrgbcolor}$\ell _{a2}$\special{ps: grestore}}}}%don't touch at this line...
\put(486,-421){\makebox(0,0)[lb]{\smash{\SetFigFont{12}{14.4}{\rmdefault}{\mddefault}{\updefault}\special{ps: gsave 0 0 0 setrgbcolor}$\ell _{b2}$\special{ps: grestore}}}}%don't touch at this line...
\put(4701,-871){\makebox(0,0)[lb]{\smash{\SetFigFont{12}{14.4}{\rmdefault}{\mddefault}{\updefault}\special{ps: gsave 0 0 0 setrgbcolor}$\ell _{b1}$\special{ps: grestore}}}}%don't touch at this line...
\put(4456,-1816){\makebox(0,0)[lb]{\smash{\SetFigFont{12}{14.4}{\rmdefault}{\mddefault}{\updefault}\special{ps: gsave 0 0 0 setrgbcolor}$\ell _{a1}$\special{ps: grestore}}}}%don't touch at this line...
\put(2901,-3631){\makebox(0,0)[lb]{\smash{\SetFigFont{12}{14.4}{\rmdefault}{\mddefault}{\updefault}\special{ps: gsave 0 0 0 setrgbcolor}$\ell _{b3}$\special{ps: grestore}}}}%don't touch at this line...
\put(2296,-3931){\makebox(0,0)[lb]{\smash{\SetFigFont{12}{14.4}{\rmdefault}{\mddefault}{\updefault}\special{ps: gsave 0 0 0 setrgbcolor}$\ell _{a3}$\special{ps: grestore}}}}%don't touch at this line...
\put(2276,-3346){\makebox(0,0)[lb]{\smash{\SetFigFont{12}{14.4}{\rmdefault}{\mddefault}{\updefault}\special{ps: gsave 0 0 0 setrgbcolor}$L_3$\special{ps: grestore}}}}%don't touch at this line...
\end{picture}
\end{center}
\caption{A two-robots panorama example.}\label{two-robots-landm}
\end{figure*}

Consider now the situation depicted in Figure \ref{two-robots-landm}, with
two robots, $R_a$ and $R_b$, embedded in a three-landmark environment. As
long as we are only concerned with the panorama of one of the two robots,
say $R_a$, we can use the RA $\atra$ to represent it, by providing for each
triple of the four $\dlignes$ $\ell _{a1}$, $\ell _{a2}$, $\ell _{a3}$ and
$\ell _{ab}$, connecting, respectively, the robot $R_a$ to the three
landmarks $L_1$, $L_2$ and $L_3$ and to
the robot $R_b$, the order in which they
appear in a 360-degrees anticlockwise turn about $R_a$'s location. Because
the two robots are embedded in a same environment, it is clearly unrealistic
to consider only the panorama of one of them. The knowledge should thus
consist of the conjuction of the panoramas of both robots, providing thus the
way each of the two robots sees ``its environment'' (which includes the other
robot). We thus need to consider the $\dlignes$ connecting each of the two
robots to each of the three landmarks, and to the other robot. The involved
$\dlignes$ are thus not concurrent, as is the case with a one-robot panorama.
The RA $\atra$, which handles 2D orientations, which can be viewed as
$\dlignes$ through a fixed point (see Definition \ref{isos}, isomorphism
$\isotrois$), is therefore not sufficient to represent the knowledge at hand:
the RA $\dlalg$ is, however, well-suited for the purpose, as we show below.

\begin{enumerate}
  \item\label{conj1} The panorama of robot $R_a$, which provides the $\atra$ relation on each
    triple of the $\dlignes$ joining $R_a$'s position to the different landmarks
    and to the other robot, is given by the conjunction
    $\rlr (l_{a1},l_{a2},l_{a3})\wedge\rlr (l_{a1},l_{a2},l_{ab})\wedge\rrr (l_{a1},l_{a3},l_{ab})$.
    The $\atra$ relation on the triple $(l_{a2},l_{a3},l_{ab})$ is not provided
    explicitly, but it is implicitly present in the knowledge, and can be
    inferred by propagation as follows:
    \begin{enumerate}
      \item Using the rotation operation, we get
        from $\rlr (l_{a1},l_{a2},l_{a3})$ and $\rlr (l_{a1},l_{a2},l_{ab})$ the
        relations $\lll (l_{a2},l_{a3},l_{a1})$ and $\lll (l_{a2},l_{ab},l_{a1})$,
        respectively. From $\lll (l_{a2},l_{ab},l_{a1})$, we get, using the converse
        operation, the relation $\lrl (l_{a2},l_{a1},l_{ab})$. Finally, from the
        conjunction $\lll (l_{a2},l_{a3},l_{a1})\wedge\lrl (l_{a2},l_{a1},l_{ab})$,
        we get, using the composition operation, a first relation on the triple
        $(l_{a2},l_{a3},l_{ab})$: $\{\lel ,\lll ,\lrl\}(l_{a2},l_{a3},l_{ab})$.
      \item Using rotation and then converse, we get from
        $\lll (l_{a2},l_{a3},l_{a1})$ the relation $\rll (l_{a3},l_{a2},l_{a1})$,
        and from $\rrr (l_{a1},l_{a3},l_{ab})$ the relation
        $\lrr (l_{a3},l_{a1},l_{ab})$. Using composition, we infer from the
        conjunction $\rll (l_{a3},l_{a2},l_{a1})\wedge\lrr (l_{a3},l_{a1},l_{ab})$
        the relation $\{\lre ,\lrl ,\lrr\}$ on the triple $(l_{a3},l_{a2},l_{ab})$:\\
        $\{\rer ,\rlr ,\rrr\}(l_{a3},l_{a2},l_{ab})$. Using, again, rotation and
        then converse, we get a second relation on the triple
        $(l_{a2},l_{a3},l_{ab})$: $\{\lre ,\lrl ,\lrr\}(l_{a2},l_{a3},l_{ab})$.
      \item Intersecting the results of the last two points, we get the final
        relation on the triple $(l_{a2},l_{a3},l_{ab})$:
        $\{\lrl\}(l_{a2},l_{a3},l_{ab})$.
    \end{enumerate}
  \item\label{conj2} Similarly, the panorama of robot $R_b$ is given by the conjunction
    $\{\lor\}(l_{b1},l_{b2},l_{b3})\wedge\{\lrl\}(l_{b1},l_{b2},l_{ba})$.
    The $\atra$ relation on the triple $(l_{b1},l_{b3},l_{ba})$ as well as the
    one on the triple $(l_{b2},l_{b3},l_{ba})$ are not provided explicitly, but
    they are implicitly present in the knowledge.
\end{enumerate}
It is important to note that, if we want to combine the knowledge
consisting of $R_a$'s panorama, on the one hand, and $R_b$'s panorama, on the
other hand, we cannot any longer use orientations, but $\dlignes$: simply
because the $\dligne$ variables involved in $R_a$'s panorama and the ones involved
in $R_b$'s panorama are not all concurrent. As a consequence, we have to
leave the realm of 2D orientations and enter the one of $\dlignes$: we transform
the above knowledge into the RA $\dlalg$. Basically, we need to add to the
relations in the previous enumeration the fact that the arguments of each triple
consist of concurrent $\dlignes$: we need to use the $\dltalg$ relation $\cc _=$.
The main two conjunctions in Items \ref{conj1} and \ref{conj2} become,
respectively, as follows:
$\{\langle\cc _=,\rlr\rangle\}(l_{a1},l_{a2},l_{a3})\wedge
 \{\langle\cc _=,\rlr\rangle\}(l_{a1},l_{a2},l_{ab})\wedge
 \{\langle\cc _=,\rrr\rangle\}(l_{a1},l_{a3},l_{ab})$
and
$\{\langle\cc _=,\lor\rangle\}(l_{b1},l_{b2},l_{b3})\wedge
 \{\langle\cc _=,\lrl\rangle\}(l_{b1},l_{b2},l_{ba})$.

What has been done so far expresses only relations whose arguments consist of
$\dlignes$ that are (1) all incident with $R_a$'s position, or (2) all incident
with $R_b$'s position. Konowledge combining the two kinds of $\dlignes$ needs also
to be expressed; examples include the following:
\begin{enumerate}
  \item The $\dlignes$ $l_{ab}$ and $l_{ba}$ coincide and are of opposite
    orientations; this can be expressed thus:
    $\{\langle\pp _{c1},\oeo\rangle\}(l_{ab},l_{ba},l_{ba})$.
  \item the $\dligne$ $l_{a3}$ cuts the $\dligne$ $l_{ab}$ before $l_{b3}$ does:
    $\{\cc _<\}(l_{ab},l_{a3},l_{b3})$; the orientational knowledge on the same triple
    is $\{\lll\}(l_{ab},l_{a3},l_{b3})$; the positional knowledge is thus the $\dlalg$
    relation $\{\langle\cc _<,\lll\rangle\}(l_{ab},l_{a3},l_{b3})$.
\end{enumerate}
\subsection{Natural language processing: representation of motion prepositions}
According to Herskovits \cite{Herskovits97a}, every motion preposition fits in a syntactic frame
\begin{center}
\begin{tabular}{llll}
NP &[activity verb] &Preposition &NP
\end{tabular}
\end{center}
Examples include:
\begin{enumerate}
  \item The ball rolled across the street.
  \item The ball rolled along the street.
  \item The ball rolled toward the boy.
\end{enumerate}
The moving object is referred to as the {\em Figure}; the referent of the object
of the preposition (the reference object) is referred to as the {\em Ground}
\cite{Talmy83a}. The preposition constrains the trajectory, or path of the Figure.

``In conclusion, a motion preposition defines a field of directed lines w.r.t.
the Ground'' \cite{Herskovits97a}.

The examples above on the use of motion prepositions concern perception.
Herskovits \cite{Herskovits97a} also discusses the use of motion prepositions in
motion planning, as well as in navigation and cognitive maps:
\begin{enumerate}
  \item ``The linguistic representation of objects' paths as lines is fundamental
    to motion planning'' (\cite{Herskovits97a}, page 174).
  \item ``Navigation in large-scale spaces is guided by cognitive maps whose
    major components are landmarks and routes, represented, respectively, as
    points and lines. Moreover, in the context of a cognitive map, a moving
    Figure is conceptualised as a point, and its trajectory as a line''
    (\cite{Herskovits97a}, page 174).
\end{enumerate}
This shows the importance of $\dlignes$, and points, for the representation of
motion prepositions. As we have already seen, points can be represented in our
RA $\dlalg$ as a pair of cutting $\dlignes$. To relate Herskovits work to ours,
we show how to represent in the RA $\dlalg$ Talmy's schema for ``across''
(Figure \ref{talmys-across}(a)), as well as the third sentence in the examples' list above,
``the ball rolled toward the boy''.

Talmy \cite{Talmy83a} has provided a list of conditions defining ``across'' (see also
\cite{Herskovits97a}, page 182). We consider here a more general definition for
``across'', given by the following conditions (F = the Figure object; G = the Ground
object):
\begin{enumerate}
  \item[a.] F is linear and bounded at both ends (a line segment)
  \item[b.] G is ribbonal ---the part of the plane between two
    parallel lines
  \item[d.] The axes of F and G are strictly cutting; i.e., they have a single-point
    intersection
  \item[e.] F and G are coplanar
  \item[g.] F's length is at least as great as G's width
  \item[h.] F touches both of G's edges (G's edges are here the lines bounding it)
\end{enumerate}
The items in the enumeration are numbered alphabetically, and the letters match the ones
of the corresponding condidions in the list given by Herskovits (\cite{Herskovits97a},
page 182).

The Figure F has a directionality and is considered as a directed line segment
$F=(P_1,P_2)$. Therefore there are two possibilities for F to be across G (see Figure
\ref{talmys-across}(b-c)). According to what we have seen in Subsection
\ref{gissubsection} (Item \ref{gisitem2}), the $\dlalg$ representation of F is a
triple $\psi (F)=(\ell _3,\ell _4,\ell _5)$ of $\dlignes$ verifying
    $\{\langle\cc _>,\lor\rangle\} (\ell _3,\ell _4,\ell _5)$.

The Ground G can be represented as a pair $(\ell _1,\ell _2)$ of $\dlignes$ such that
$\ell _2$ coincides with, or is parallel to, and lies within the left half-plane bounded by,
$\ell _1$. In other words, $\ell _1$ and $\ell _2$ are such that
$\leftpar (\ell _2,\ell _1)\vee\coincides (\ell _2,\ell _1)$. We refer to the pair
$(\ell _1,\ell _2)$ as the $\dlalg$ representation of G, and denote it by $\psi (G)$:
$\psi (G)=(\ell _1,\ell _2)$.

Given the representations $\psi (F)$ and $\psi (G)$, of F and G, respectively, the
condition for F to be across G can now be stated in terms of $\dlalg$ relations as
follows:
\begin{enumerate}
  \item $\ell _2$ is parallel to, and lies within the left half-plane bounded by,
    $\ell _1$; and $\ell _3$ cuts $\ell _1$: $\{\pc _l\}(\ell _1,\ell _2,\ell _3)$.
  \item $\ell _5$ cuts $\ell _3$ before $\ell _1$ does;
        $\ell _5$ cuts $\ell _3$ before $\ell _2$ does;
        $\ell _1$ cuts $\ell _3$ before $\ell _4$ does; and
        $\ell _2$ cuts $\ell _3$ before $\ell _4$ does:
        $\{\cc _<\}(\ell _3,\ell _5,\ell _1)\wedge
         \{\cc _<\}(\ell _3,\ell _5,\ell _2)\wedge
         \{\cc _<\}(\ell _3,\ell _1,\ell _4)\wedge
         \{\cc _<\}(\ell _3,\ell _2,\ell _4)$.
\end{enumerate}
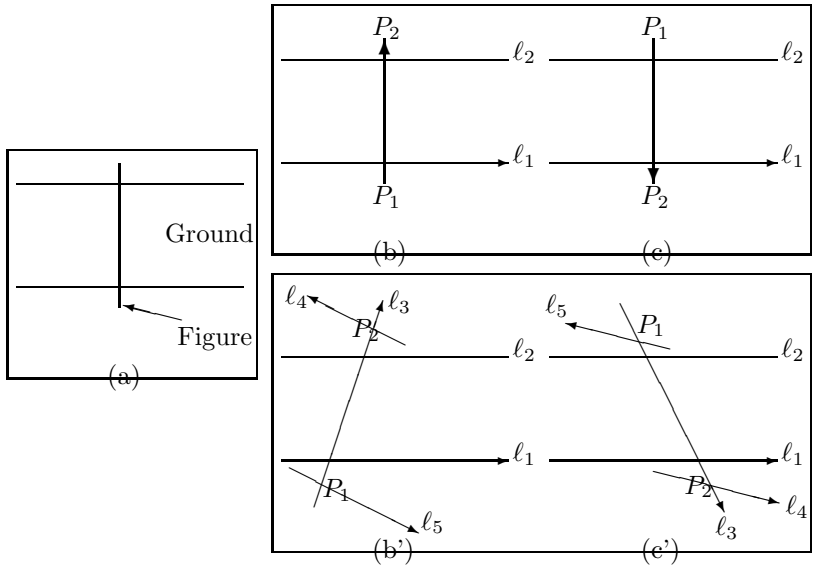
\begin{figure}[t]
\setlength{\unitlength}{2279sp}%
\begingroup\makeatletter\ifx\SetFigFont\undefined%
\gdef\SetFigFont#1#2#3#4#5{%
  \reset@font\fontsize{#1}{#2pt}%
  \fontfamily{#3}\fontseries{#4}\fontshape{#5}%
  \selectfont}%
\fi\endgroup%
\begin{picture}(8754,5964)(2149,-6463)
\thinlines
\special{ps: gsave 0 0 0 setrgbcolor}\put(5041,-3211){\framebox(5850,2700){}}
\special{ps: gsave 0 0 0 setrgbcolor}\put(5131,-1096){\line( 1, 0){2475}}
\special{ps: grestore}\special{ps: gsave 0 0 0 setrgbcolor}\put(5131,-2221){\vector( 1, 0){2475}}%no ...
\special{ps: grestore}\special{ps: gsave 0 0 0 setrgbcolor}\put(8056,-1096){\line( 1, 0){2475}}
\special{ps: grestore}\special{ps: gsave 0 0 0 setrgbcolor}\put(8056,-2221){\vector( 1, 0){2475}}%no ...
\special{ps: grestore}\thicklines
\special{ps: gsave 0 0 0 setrgbcolor}\put(6256,-2446){\vector( 0, 1){1575}}%no ...
\special{ps: grestore}\special{ps: gsave 0 0 0 setrgbcolor}\put(9181,-871){\vector( 0,-1){1575}}%no ...
\special{ps: grestore}\put(10576,-2221){\makebox(0,0)[lb]{\smash{\SetFigFont{10}{12.0}{\rmdefault}{\mddefault}{\updefault}\special{ps: gsave 0 0 0 setrgbcolor}$\ell _1$\special{ps: grestore}}}}
\put(10576,-1096){\makebox(0,0)[lb]{\smash{\SetFigFont{10}{12.0}{\rmdefault}{\mddefault}{\updefault}\special{ps: gsave 0 0 0 setrgbcolor}$\ell _2$\special{ps: grestore}}}}
\put(7651,-2221){\makebox(0,0)[lb]{\smash{\SetFigFont{10}{12.0}{\rmdefault}{\mddefault}{\updefault}\special{ps: gsave 0 0 0 setrgbcolor}$\ell _1$\special{ps: grestore}}}}
\put(7651,-1096){\makebox(0,0)[lb]{\smash{\SetFigFont{10}{12.0}{\rmdefault}{\mddefault}{\updefault}\special{ps: gsave 0 0 0 setrgbcolor}$\ell _2$\special{ps: grestore}}}}
\put(6121,-826){\makebox(0,0)[lb]{\smash{\SetFigFont{10}{12.0}{\rmdefault}{\mddefault}{\updefault}\special{ps: gsave 0 0 0 setrgbcolor}$P_2$\special{ps: grestore}}}}
\put(6121,-3276){\makebox(0,0)[lb]{\smash{\SetFigFont{10}{12.0}{\rmdefault}{\mddefault}{\updefault}\special{ps: gsave 0 0 0 setrgbcolor}(b)\special{ps: grestore}}}}
\put(9046,-3276){\makebox(0,0)[lb]{\smash{\SetFigFont{10}{12.0}{\rmdefault}{\mddefault}{\updefault}\special{ps: gsave 0 0 0 setrgbcolor}(c)\special{ps: grestore}}}}
\put(9046,-826){\makebox(0,0)[lb]{\smash{\SetFigFont{10}{12.0}{\rmdefault}{\mddefault}{\updefault}\special{ps: gsave 0 0 0 setrgbcolor}$P_1$\special{ps: grestore}}}}
\put(6121,-2671){\makebox(0,0)[lb]{\smash{\SetFigFont{10}{12.0}{\rmdefault}{\mddefault}{\updefault}\special{ps: gsave 0 0 0 setrgbcolor}$P_1$\special{ps: grestore}}}}
\put(9046,-2671){\makebox(0,0)[lb]{\smash{\SetFigFont{10}{12.0}{\rmdefault}{\mddefault}{\updefault}\special{ps: gsave 0 0 0 setrgbcolor}$P_2$\special{ps: grestore}}}}
\thinlines
\special{ps: gsave 0 0 0 setrgbcolor}\put(5131,-4336){\line( 1, 0){2475}}
\special{ps: grestore}\special{ps: gsave 0 0 0 setrgbcolor}\put(5131,-5461){\vector( 1, 0){2475}}%no ...
\special{ps: grestore}\special{ps: gsave 0 0 0 setrgbcolor}\put(8056,-4336){\line( 1, 0){2475}}
\special{ps: grestore}\special{ps: gsave 0 0 0 setrgbcolor}\put(8056,-5461){\vector( 1, 0){2475}}%no ...
\special{ps: grestore}\special{ps: gsave 0 0 0 setrgbcolor}\put(5491,-5961){\vector( 1, 3){747}}%no ...
\special{ps: grestore}\special{ps: gsave 0 0 0 setrgbcolor}\put(6481,-4201){\vector(-2, 1){1062}}%l4l
\special{ps: grestore}\special{ps: gsave 0 0 0 setrgbcolor}\put(8821,-3751){\vector( 1,-2){1134}}%no ...
\special{ps: grestore}\special{ps: gsave 0 0 0 setrgbcolor}\put(9181,-5576){\vector( 4,-1){1365.882}}%l4r
\special{ps: grestore}\special{ps: gsave 0 0 0 setrgbcolor}\put(9361,-4246){\vector(-4, 1){1134}}%l5r
\special{ps: grestore}\special{ps: gsave 0 0 0 setrgbcolor}\put(5221,-5541){\vector( 2,-1){1408.235}}%l5l
\special{ps: grestore}\put(6121,-6516){\makebox(0,0)[lb]{\smash{\SetFigFont{10}{12.0}{\rmdefault}{\mddefault}{\updefault}\special{ps: gsave 0 0 0 setrgbcolor}(b')\special{ps: grestore}}}}
\put(9046,-6516){\makebox(0,0)[lb]{\smash{\SetFigFont{10}{12.0}{\rmdefault}{\mddefault}{\updefault}\special{ps: gsave 0 0 0 setrgbcolor}(c')\special{ps: grestore}}}}
\put(10576,-5461){\makebox(0,0)[lb]{\smash{\SetFigFont{10}{12.0}{\rmdefault}{\mddefault}{\updefault}\special{ps: gsave 0 0 0 setrgbcolor}$\ell _1$\special{ps: grestore}}}}
\put(7651,-5461){\makebox(0,0)[lb]{\smash{\SetFigFont{10}{12.0}{\rmdefault}{\mddefault}{\updefault}\special{ps: gsave 0 0 0 setrgbcolor}$\ell _1$\special{ps: grestore}}}}
\put(7651,-4336){\makebox(0,0)[lb]{\smash{\SetFigFont{10}{12.0}{\rmdefault}{\mddefault}{\updefault}\special{ps: gsave 0 0 0 setrgbcolor}$\ell _2$\special{ps: grestore}}}}
\put(5581,-5856){\makebox(0,0)[lb]{\smash{\SetFigFont{10}{12.0}{\rmdefault}{\mddefault}{\updefault}\special{ps: gsave 0 0 0 setrgbcolor}$P_1$\special{ps: grestore}}}}
\put(9521,-5821){\makebox(0,0)[lb]{\smash{\SetFigFont{10}{12.0}{\rmdefault}{\mddefault}{\updefault}\special{ps: gsave 0 0 0 setrgbcolor}$P_2$\special{ps: grestore}}}}
\put(9001,-4066){\makebox(0,0)[lb]{\smash{\SetFigFont{10}{12.0}{\rmdefault}{\mddefault}{\updefault}\special{ps: gsave 0 0 0 setrgbcolor}$P_1$\special{ps: grestore}}}}
\put(5896,-4111){\makebox(0,0)[lb]{\smash{\SetFigFont{10}{12.0}{\rmdefault}{\mddefault}{\updefault}\special{ps: gsave 0 0 0 setrgbcolor}$P_2$\special{ps: grestore}}}}
\put(5176,-3751){\makebox(0,0)[lb]{\smash{\SetFigFont{10}{12.0}{\rmdefault}{\mddefault}{\updefault}\special{ps: gsave 0 0 0 setrgbcolor}$\ell _4$\special{ps: grestore}}}}
\put(6296,-3806){\makebox(0,0)[lb]{\smash{\SetFigFont{10}{12.0}{\rmdefault}{\mddefault}{\updefault}\special{ps: gsave 0 0 0 setrgbcolor}$\ell _3$\special{ps: grestore}}}}
\put(6656,-6211){\makebox(0,0)[lb]{\smash{\SetFigFont{10}{12.0}{\rmdefault}{\mddefault}{\updefault}\special{ps: gsave 0 0 0 setrgbcolor}$\ell _5$\special{ps: grestore}}}}
\put(8006,-3851){\makebox(0,0)[lb]{\smash{\SetFigFont{10}{12.0}{\rmdefault}{\mddefault}{\updefault}\special{ps: gsave 0 0 0 setrgbcolor}$\ell _5$\special{ps: grestore}}}}
\put(10576,-4336){\makebox(0,0)[lb]{\smash{\SetFigFont{10}{12.0}{\rmdefault}{\mddefault}{\updefault}\special{ps: gsave 0 0 0 setrgbcolor}$\ell _2$\special{ps: grestore}}}}
\put(10621,-6046){\makebox(0,0)[lb]{\smash{\SetFigFont{10}{12.0}{\rmdefault}{\mddefault}{\updefault}\special{ps: gsave 0 0 0 setrgbcolor}$\ell _4$\special{ps: grestore}}}}
\put(9856,-6271){\makebox(0,0)[lb]{\smash{\SetFigFont{10}{12.0}{\rmdefault}{\mddefault}{\updefault}\special{ps: gsave 0 0 0 setrgbcolor}$\ell _3$\special{ps: grestore}}}}
\special{ps: gsave 0 0 0 setrgbcolor}\put(5041,-6451){\framebox(5850,3015){}}
\special{ps: gsave 0 0 0 setrgbcolor}\put(2251,-2446){\line( 1, 0){2475}}
\special{ps: grestore}\special{ps: gsave 0 0 0 setrgbcolor}\put(2251,-3571){\line( 1, 0){2475}}
\special{ps: grestore}\special{ps: gsave 0 0 0 setrgbcolor}\put(4051,-3931){\vector(-4, 1){635.294}}%no ...
\special{ps: grestore}\thicklines
\special{ps: gsave 0 0 0 setrgbcolor}\put(3376,-3796){\line( 0, 1){1575}}
\special{ps: grestore}\thinlines
\special{ps: gsave 0 0 0 setrgbcolor}\put(2161,-4561){\framebox(2700,2475){}}
\put(3241,-4626){\makebox(0,0)[lb]{\smash{\SetFigFont{10}{12.0}{\rmdefault}{\mddefault}{\updefault}\special{ps: gsave 0 0 0 setrgbcolor}(a)\special{ps: grestore}}}}
\put(3871,-3076){\makebox(0,0)[lb]{\smash{\SetFigFont{10}{12.0}{\rmdefault}{\mddefault}{\updefault}\special{ps: gsave 0 0 0 setrgbcolor}Ground\special{ps: grestore}}}}
\put(4006,-4201){\makebox(0,0)[lb]{\smash{\SetFigFont{10}{12.0}{\rmdefault}{\mddefault}{\updefault}\special{ps: gsave 0 0 0 setrgbcolor}Figure\special{ps: grestore}}}}
\end{picture}
\caption{Diagrammatic representation of Talmy's schema for ``across''.}\label{talmys-across}
\end{figure}
\section{Related work}\label{relatedwork}
We now discuss the most related work in the literature.
\subsection{Scivos and Nebel's NP-hardness result of Freksa's calculus}\label{SubsectionScivos}
\begin{figure}[t]
\centerline{\epsfxsize=10.5cm\epsfysize=5.1cm\epsffile{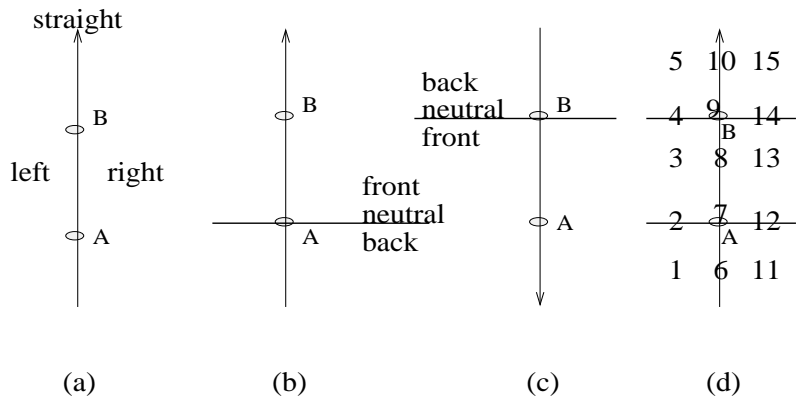}}
\caption{The partition of the plane on which is based
         the relative orientation calculus in \cite{Freksa92b,Zimmermann96a}.}\label{rel-orient}
\end{figure}
A well-known model of relative orientation of 2D points is the
calculus, often referred to as the Double-Cross Calculus, defined in \cite{Freksa92b}, and developed further
in \cite{Zimmermann96a}. The
calculus corresponds to a specific partition, into $15$ regions, of the plane, determined by a
parent object, say $A$, and a reference object, say $B$ (Figure \ref{rel-orient}(d)). The
partition is based on the following:
\begin{enumerate}
  \item the {\it left/straight/right} partition of the plane determined by an observer placed
    at the parent object and looking in the direction of the reference object
    (Figure \ref{rel-orient}(a));
  \item the {\it front/neutral/back} partition of the plane determined by the same observer
    (Figure \ref{rel-orient}(b)); and
  \item the similar {\it front/neutral/back} partition of the plane obtained when we swap the
    roles of the parent object and the reference object (Figure \ref{rel-orient}(c)).
\end{enumerate}
Combining the three partitions (a), (b) and (c) of Figure \ref{rel-orient}
leads to the partition of the plane on which is based the
calculus in \cite{Freksa92b,Zimmermann96a} (Figure \ref{rel-orient}(d)). The
region numbered $n$, $n\in\{1,\ldots ,15\}$, in the partition is referred
to as $\region (A,B,n)$, and gives rise to a basic relation, or atom, of the
calculus, which we refer to as $f_n$:
$$(\forall n\in\{1,\ldots ,15\})(\forall C)(f_n(A,B,C)\Leftrightarrow C\in\region (A,B,n))$$

Scivos and Nebel \cite{Scivos01a} have shown that the subset $\{\{f_{10}\},T\}$, where $T$ is
the universal relation, is NP-hard; the proof uses a reduction of the betweenness problem
(\cite{Garey79a}, page 279). We consider here a coarser version of Freksa's calculus, which
does not distinguish between $f_1$, $f_2$, $f_3$, $f_4$ and $f_5$, on the one hand, and
between $f_{11}$, $f_{12}$, $f_{13}$, $f_{14}$ and $f_{15}$, on the other hand. We show that
a CSP, $P$, expressed in $FC=\{f_{\ell},\{f_6\},\{f_7\},\{f_8\},\{f_9\},\{f_{10}\},f_r,T\}$,
where $f_{\ell}=\{f_1,f_2,f_3,f_4,f_5\}$ and $f_r=\{f_{11},f_{12},f_{13},f_{14},f_{15}\}$,
can be translated into an equivalent $\dlalg$-CSP, $P'$. The idea is to first eliminate the
relations $\{f_7\}$ and $\{f_9\}$ from $P$, which involve necessarily equal variables (see
Definition \ref{necessarilyequal} below). We then show how to translate each constraint of the resulting problem, expressed in
$FC\setminus\{\{f_7\},\{f_9\}\}=\{f_{\ell},\{f_6\},\{f_8\},\{f_{10}\},f_r,T\}$, into the RA
$\dlalg$.
\begin{df}\label{necessarilyequal}
Two variables of a CSP are necessarily equal if they receive the same instantiation in all
models of the CSP.
\end{df}
\begin{enumerate}
  \item\label{item2justabove} Let $X_i$, $X_j$ and $X_k$ be three variables such that
    $\{f_7\}(X_i,X_j,X_k)$. Then $X_i$ and $X_k$ are necessarily equal. In such a case,
    we perform the following, for all variables $X_l$ and $X_m$:
    \begin{enumerate}
      \item $(\termat ^P)_{lmi}\leftarrow (\termat ^P)_{lmi}\cap (\termat ^P)_{lmk}$
      \item $(\termat ^P)_{lim}\leftarrow (\termat ^P)_{lim}\cap (\termat ^P)_{lkm}$
      \item $(\termat ^P)_{ilm}\leftarrow (\termat ^P)_{ilm}\cap (\termat ^P)_{klm}$
    \end{enumerate}
    If in any of the above replacement operations, the empty relation is detected
    then the CSP is clearly inconsistent. Otherwise, the variable $X_k$ can be
    removed from the CSP: replacement the CSP by the sub-CSP
    $P_{|V\setminus\{X_k\}}$, where $V$ is the set of all variables.
  \item If there exist variables $X_i$, $X_j$ and $X_k$ such that
    $\{f_7\}(X_i,X_j,X_k)$ then repeat the process from \ref{item2justabove}.
  \item If there exist three variables $X_i$, $X_j$ and $X_k$ such that
    $\{f_9\}(X_i,X_j,X_k)$ then $X_j$ and $X_k$ are necessarily equal. The
    constraint $\{f_9\}(X_i,X_j,X_k)$ is equivalent to the constraint
    $\{f_7\}(X_j,X_i,X_k)$:
    \begin{enumerate}
      \item $(\termat ^P)_{jik}\leftarrow (\termat ^P)_{jik}\cap\{f_7\}$
      \item If $(\termat ^P)_{jik}=\emptyset$ then exit (the CSP is inconsistent)
      \item Repeat the process from Item \ref{item2justabove}
    \end{enumerate}
  \item If there exist no three variables $X_i$, $X_j$ and $X_k$ such that
    $\{f_9\}(X_i,X_j,X_k)$ then the process has been achieved: the resulting
    CSP has no two variables that are necessarily equal; in other words, it
    is expressed in $FC\setminus\{\{f_7\},\{f_9\}\}$.
\end{enumerate}
The second step is to show how to translate a CSP, $P$, expressed in
$FC\setminus\{\{f_7\},\{f_9\}\}$ into a CSP $P'$ expressed in the RA $\dlalg$.
\begin{enumerate}
  \item\label{etape1} Initialise the set $V'$ of variables and the set $C'$ of constraints of $P'$ to the
    empty set:\\
    $V'\leftarrow\emptyset$,
    $C'\leftarrow\emptyset$
  \item\label{etape2} For each pair $(X_i,X_j)$, $i<j$, of variables from $V$, the set of
    variables of $P$, we create a $\dligne$ variable $X_{ij}$:\\
    $V'\leftarrow V'\cup\{X_{ij}\}$
  \item\label{etape3} For each variable $X_i$ from $V$, we create two $\dligne$ variables $X_{i_1}$ and
    $X_{i_2}$ such that $\psi (X_i)=(X_{i_1},X_{i_2})$; i.e., such that
    $\{<\cp _c,\lre >\}(X_{i_1},X_{i_2},X_{i_1})$:\\
    $V'\leftarrow V'\cup\{X_{i_1},X_{i_2}\}$,
    $C'\leftarrow C'\cup\{\{<\cp _c,\lre >\}(X_{i_1},X_{i_2},X_{i_1})\}$.
  \item\label{etape4} For all distinct variables $X_i$ and $X_j$ of $P$, we have $X_i\in X_{ij}$ and
    $X_j\in X_{ij}$:\\
    $C'\leftarrow C'\cup\{\{\cc _=,\cp _c,\pc _c\}(X_{ij},X_{i_1},X_{i_2}),
                          \{\cc _=,\cp _c,\pc _c\}(X_{ij},X_{j_1},X_{j_2})\}$
  \item\label{etape5} $X_{ij}$ is oriented from $X_i$ to $X_j$; i.e., $X_i$ is met before $X_j$ in the
    positive walk along $X_{ij}$:\footnote{The conditions $\psi (X_i)=(X_{i_1},X_{i_2})$
    and $\psi (X_j)=(X_{j_1},X_{j_2})$ imply that in all solutions to the CSP $P'$ in
    construction, the instantiations
    $\{X_{ij}=\ell _{ij},X_{i_1}=\ell _{i_1},X_{i_2}=\ell _{i_2},X_{j_1}=\ell _{j_1},X_{j_2}=\ell _{j_2}\}$
    of the variables in $\{X_{ij},X_{i_1},X_{i_2},X_{j_1},X_{j_2}\}$ are such that
    $\ell _{i_1}$ and $\ell _{i_2}$, on the one hand, and $\ell _{j_1}$ and
    $\ell _{j_2}$, on the other hand, cannot be both parallel to $\ell _{ij}$.}\\
    $C'\leftarrow C'\cup\{\{\cc _<,\cp _c,\pc _c,\pp _{c1}\}(X_{ij},X_{i_1},X_{j_1}),
                          \{\cc _<,\cp _c,\pc _c,\pp _{c1}\}(X_{ij},X_{i_2},X_{j_2})\}$
  \item\label{etape6} For each constraint of $P$ of the form $\{f_{\ell}\}(X_i,X_j,X_k)$, we add the
    constraint $\{\lel\}(X_{ij},X_{ik},X_{ik})$ to $P'$:\\
    $C'\leftarrow C'\cup\{\{\lel\}(X_{ij},X_{ik},X_{ik})\}$
  \item\label{etape7} For each constraint of $P$ of the form $\{f_6\}(X_i,X_j,X_k)$, we add the
    constraint $\{\oeo\}(X_{ij},X_{ik},X_{ik})$ to $P'$:\\
    $C'\leftarrow C'\cup\{\{\oeo\}(X_{ij},X_{ik},X_{ik})\}$
  \item\label{etape8} For each constraint of $P$ of the form $\{f_8\}(X_i,X_j,X_k)$, we add the
    constraints $\{\eee\}(X_{ij},X_{ik},X_{ik})$ and $\{\oeo\}(X_{ij},X_{jk},X_{jk})$
    to $P'$:\\
    $C'\leftarrow C'\cup\{\{\eee\}(X_{ij},X_{ik},X_{ik}),\{\oeo\}(X_{ij},X_{jk},X_{jk})\}$
  \item\label{etape9} For each constraint of $P$ of the form $\{f_{10}\}(X_i,X_j,X_k)$, transform
    it into the equivalent constraint $\{f_6\}(X_j,X_i,X_k)$ and apply Step \ref{etape7}
  \item\label{etape10} For each constraint of $P$ of the form $\{f_r\}(X_i,X_j,X_k)$, transform
    it into the equivalent constraint $\{f_{\ell}\}(X_j,X_i,X_k)$ and apply Step \ref{etape6}
\end{enumerate}
\subsection{Moratz et al.'s dipole algebra and Renz's spatial Odyssey of Allen's interval algebra}
A dipole is an oriented line segment. We follow here the notation in \cite{Moratz00a}, and denote
dipoles by the letters A, B, C, ..., the starting endpoint and the ending endpoint of a dipole
$A$ by $\dipolese _A$ and $\dipoleee _A$, respectively. The simple version of the dipole algebra,
denoted $\dipolesa$, presents 24 atoms, which are characterised by the fact that they cannot
represent a configuration of two dipoles with at least three of the four endpoints collinear and
pairwise distinct. The reason for presenting a simple version of the algebra is that it has the
advantage of being a relation algebra\footnote{As far as we can say, the authors did not check,
for instance, that the entries of the composition table record the exact composition of the
corresponding atoms; i.e., whether, given any two atoms, say $r$ and $s$, it is the case that
$r\circ s=T[r,s]$, where $T[r,s]$ is the entry at row $r$ and column $s$ of the composition
table. If this is not the case, the calculus would not be an RA.}
\cite{Tarski41b,Ladkin94a,Isli00b}, and of
presenting a relatively small number of atoms, offering thus a manageable composition table, and
a wide application domain. The complete version, denoted $\dipoleca$, contains 69 atoms that are
Jointly Exhaustive and Pairwise Disjoint: any spatial configuration of two dipoles is described
by one and only one of the 69 atoms. We focus on the more general version, $\dipoleca$. The
description of $\dipoleca$ atoms is based on seven dipole-point relations, $l$ (left), $b$
(behind), $s$ (starts), $i$ (inside), $e$ (ends), $f$ (front) and $r$ (right): given a dipole
$A=(\dipolese _A,\dipoleee _A)$, a point $P$ and a dipole-point relation $R\in\{l,b,s,i,e,f,r\}$,
we have $A\mbox{ }R\mbox{ }P$ if and only if the point $P$ belongs to the region labelled $R$ in
the partition of the plane determined by $A$, illustrated in Figure \ref{dipolepartition} (see \cite{Moratz00a}
for details).
\begin{figure}
\begin{center}
  \setlength{\unitlength}{4144sp}%
\begingroup\makeatletter\ifx\SetFigFont\undefined%
\gdef\SetFigFont#1#2#3#4#5{%
  \reset@font\fontsize{#1}{#2pt}%
  \fontfamily{#3}\fontseries{#4}\fontshape{#5}%
  \selectfont}%
\fi\endgroup%
\begin{picture}(1350,3174)(1351,-3673)
\thicklines
\special{ps: gsave 0 0 0 setrgbcolor}\put(2026,-2761){\line( 0, 1){1350}}
\special{ps: grestore}\thinlines
\special{ps: gsave 0 0 0 setrgbcolor}\multiput(2026,-1411)(0.00000,120.00000){8}{\line( 0, 1){ 60.000}}
\special{ps: grestore}\special{ps: gsave 0 0 0 setrgbcolor}\multiput(2026,-3661)(0.00000,120.00000){8}{\line( 0, 1){ 60.000}}
\special{ps: grestore}\put(1151,-2086){\makebox(0,0)[lb]{\smash{\SetFigFont{25}{30.0}{\rmdefault}{\mddefault}{\updefault}\special{ps: gsave 0 0 0 setrgbcolor}l\special{ps: grestore}}}}
\put(2026,-3211){\makebox(0,0)[lb]{\smash{\SetFigFont{25}{30.0}{\rmdefault}{\mddefault}{\updefault}\special{ps: gsave 0 0 0 setrgbcolor}b\special{ps: grestore}}}}
\put(2026,-2761){\makebox(0,0)[lb]{\smash{\SetFigFont{25}{30.0}{\rmdefault}{\mddefault}{\updefault}\special{ps: gsave 0 0 0 setrgbcolor}s\special{ps: grestore}}}}
\put(2026,-2086){\makebox(0,0)[lb]{\smash{\SetFigFont{25}{30.0}{\rmdefault}{\mddefault}{\updefault}\special{ps: gsave 0 0 0 setrgbcolor}i\special{ps: grestore}}}}
\put(2026,-1411){\makebox(0,0)[lb]{\smash{\SetFigFont{25}{30.0}{\rmdefault}{\mddefault}{\updefault}\special{ps: gsave 0 0 0 setrgbcolor}e\special{ps: grestore}}}}
\put(2026,-961){\makebox(0,0)[lb]{\smash{\SetFigFont{25}{30.0}{\rmdefault}{\mddefault}{\updefault}\special{ps: gsave 0 0 0 setrgbcolor}f\special{ps: grestore}}}}
\put(2901,-2086){\makebox(0,0)[lb]{\smash{\SetFigFont{25}{30.0}{\rmdefault}{\mddefault}{\updefault}\special{ps: gsave 0 0 0 setrgbcolor}r\special{ps: grestore}}}}
\put(1776,-2761){\makebox(0,0)[lb]{\smash{\SetFigFont{12}{14.4}{\rmdefault}{\mddefault}{\updefault}\special{ps: gsave 0 0 0 setrgbcolor}$\dipolese _A$\special{ps: grestore}}}}
\put(1776,-1411){\makebox(0,0)[lb]{\smash{\SetFigFont{12}{14.4}{\rmdefault}{\mddefault}{\updefault}\special{ps: gsave 0 0 0 setrgbcolor}$\dipoleee _A$\special{ps: grestore}}}}
\end{picture}
  \caption{Partition of plane determined by a dipole $A=(\dipolese _A,\dipoleee _A)$ \cite{Moratz00a}.}\label{dipolepartition}
\end{center}
\end{figure}

A dipole-dipole relation is denoted in \cite{Moratz00a} by a word of length $4$ over the alphabet
$\{l,b,s,i,e,f,r\}$, of the form $R^1R^2R^3R^4$. Such a relation on two dipoles $A$ and $B$,
denoted by $A\mbox{ }R^1R^2R^3R^4\mbox{ }B$, is interpreted as follows:
\begin{equation}
A\mbox{ }R^1R^2R^3R^4\mbox{ }B\Leftrightarrow (A\mbox{ }R^1\mbox{ }\dipolese _B)\wedge
                                              (A\mbox{ }R^2\mbox{ }\dipoleee _B)\wedge
                                              (B\mbox{ }R^3\mbox{ }\dipolese _A)\wedge
                                              (B\mbox{ }R^4\mbox{ }\dipoleee _A)
\end{equation}
In order to get a $\dlalg$ representation for each of the 69 dipole relations in \cite{Moratz00a},
we need to provide a $\dlalg$ representation for each of the seven dipole-point relations
$l,b,s,i,e,f,r$. For this purpose, we consider a dipole $A=(\dipolese _A,\dipoleee _A)$ and a
point $P$:
\begin{enumerate}
  \item Each of the three points $P$, $\dipolese _A$ and $\dipoleee _A$ is associated with a pair
    of $\dlignes$ consisting of its $\dlalg$ representations (see Subsection \ref{IncGeometry}
    Item \ref{IncGeometryItem1}):
    $\psi (P)=(\ell _P^1,\ell _P^2)$,
    $\psi (\dipolese _A)=(\ell _{\dipolese _A}^1,\ell _{\dipolese _A}^2)$ and
    $\psi (\dipoleee _A)=(\ell _{\dipoleee _A}^1,\ell _{\dipoleee _A}^2)$.
    The corresponding set of $\dlalg$ constraints is $\{\{\langle\cp _c,\lre\rangle\}(\ell _P^1,\ell _P^2,\ell _P^1),
                                                        \{\langle\cp _c,\lre\rangle\}(\ell _{\dipolese _A}^1,
                                                                   \ell _{\dipolese _A}^2,\ell _{\dipolese _A}^1),$
                                                        $\{\langle\cp _c,\lre\rangle\}(\ell _{\dipoleee _A}^1,
                                                                   \ell _{\dipoleee _A}^2,\ell _{\dipoleee _A}^1)\}$
  \item We associate with $A$ a $\dligne$ $\ell _A$ oriented from $\dipolese _A$ to $\dipoleee _A$:
    \begin{enumerate}
      \item From $\dipolese _A\in\ell _A$ and $\dipoleee _A\in\ell _A$, we get the following set of constraints
        (see Subsection \ref{SubsectionScivos}, third enumeration, Item \ref{etape4}):\\
        $\{\{\cc _=,\cp _c,\pc _c\}(\ell _A,\ell _{\dipolese _A}^1,\ell _{\dipolese _A}^2),
           \{\cc _=,\cp _c,\pc _c\}(\ell _A,\ell _{\dipoleee _A}^1,\ell _{\dipoleee _A}^2)\}$
      \item From $\ell _A$ oriented from $\dipolese _A$ to $\dipoleee _A$, we get the following set of constraints
        (see Subsection \ref{SubsectionScivos}, third enumeration, Item \ref{etape5}):\\
        $\{\{\cc _<,\cp _c,\pc _c,\pp _{c1}\}(\ell _A,\ell _{\dipolese _A}^1,\ell _{\dipoleee _A}^1),
           \{\cc _<,\cp _c,\pc _c,\pp _{c1}\}(\ell _A,\ell _{\dipolese _A}^2,\ell _{\dipoleee _A}^2)\}$
    \end{enumerate}
  \item The relations $l$, $b$, $i$, $f$ and $r$ are translated into $\dlalg$ in a similar way as
    the relations $f_l$, $f_6$, $f_8$, $f_{10}$ and $f_r$ of Freksa's calculus (see Subsection
    \ref{SubsectionScivos}, third enumeration, Items \ref{etape6}-\ref{etape10}), thanks to the
    following equivalences:
    $(A\mbox{ }l\mbox{ }P)$ iff $f_l(\dipolese _A,\dipoleee _A,P)$;
    $(A\mbox{ }b\mbox{ }P)$ iff $f_6(\dipolese _A,\dipoleee _A,P)$;
    $(A\mbox{ }i\mbox{ }P)$ iff $f_8(\dipolese _A,\dipoleee _A,P)$;
    $(A\mbox{ }f\mbox{ }P)$ iff $f_{10}(\dipolese _A,\dipoleee _A,P)$; and
    $(A\mbox{ }r\mbox{ }P)$ iff $f_r(\dipolese _A,\dipoleee _A,P)$.
  \item $A\mbox{ }s\mbox{ }P$ ($P$ coincides with $\dipolese _A$) is expressed by concurrency
    of $\ell _P^1$, $\ell _P^2$ and $\ell _{\dipolese _A}^1$, on the
    one hand, and concurrency of $\ell _P^1$, $\ell _P^2$ and
    $\ell _{\dipolese _A}^2$, on the other hand:\\
    $\{\{\cc _=,\cp _c,\pc _c\}(\ell _P^1,\ell _P^2,\ell _{\dipolese _A}^1),
       \{\cc _=,\cp _c,\pc _c\}(\ell _P^1,\ell _P^2,\ell _{\dipolese _A}^2)\}$
  \item In a similar way, we translate $A\mbox{ }e\mbox{ }P$ ($P$ coincides with
    $\dipoleee _A$) using a double concurrency:\\
    $\{\{\cc _=,\cp _c,\pc _c\}(\ell _P^1,\ell _P^2,\ell _{\dipoleee _A}^1),
       \{\cc _=,\cp _c,\pc _c\}(\ell _P^1,\ell _P^2,\ell _{\dipoleee _A}^2)\}$
\end{enumerate}
Renz's work \cite{Renz01a} was motivated by applications such as traffic
scenarios, where cars and their regions of influence can be represented as
directed intervals of an underlying line representing the road. We refer to the
underlying line as $\renzrl$. The 26 atomic relations of Renz's algebra of
directed intervals \cite{Renz01a} can be seen as particular relations of Moratz
et al.'s $\dipoleca$ algebra \cite{Moratz00a}, as long as we have a mean of
constraining all involved directed intervals to belong to the underlying line
$\renzrl$.

Consider two directed intervals $x$ and $y$. $x$ and $y$ are particular oriented
segments $x=(\dipolese _x,\dipoleee _x)$ and $y=(\dipolese _y,\dipoleee _y)$. As
in the above discussion of Moratz et al.'s $\dipoleca$ algebra, we associate
with each point $P$ in $\{\dipolese _x,\dipoleee _x,\dipolese _y,\dipoleee _y\}$
its $\dlalg$ representation $\psi (P)=(\ell _P^1,\ell _P^2)$, and with each
oriented segments $S$ in $\{x,y\}$ a $\dligne$ $\ell _S$ oriented from the left
endpoint to the right endpoint of the segment.
\begin{enumerate}
  \item Constraining a directed interval, for instance $x$, to be part of the
    underlying line $\renzrl$ can be easily done with the RA $\dlalg$, by saying
    that the line $\ell _x$ coincides with $\renzrl$:\\
    $\{\langle\pp _{c1},\eee\rangle ,\langle\pp _{c1},\oeo\rangle\}(\renzrl ,\ell _x,\ell _x)$
  \item Once each of the directed intervals is constrained to be part of $\renzrl$,
    the atomic relations can be translated into $\dlalg$ using what has been done
    above for Moratz et al.'s $\dipoleca$ dipole algebra, thanks to the
    equivalences of Figure \ref{from-dia-to-pat} between directed intervals base
    relations and $\dipoleca$ base relations.
\end{enumerate}
    \begin{figure}
    \begin{scriptsize}
    \begin{center}
    $
    \begin{array}{|l|l|l|l|}  \hline
    \mbox{Directed Intervals}&\mbox{Sym-}&\mbox{Pictorial}&\dipoleca\mbox{ Base}\\
    \mbox{Base Relation}     &\mbox{bol} &\mbox{Example}  &\mbox{Relation}\\  \hline\hline
    x\mbox{ }\renzbe\mbox{ }y&\mbox{b}_=&-x->                                &x\mbox{ }ffbb\mbox{ }y\\
    ................................................&..........&&\\
    y\mbox{ }\renzfe\mbox{ }x&\mbox{f}_=&\;\;\;\;\;\;\;\;\;\;\;\;\;\;\;\;-y->&y\mbox{ }bbff\mbox{ }x\\  \hline
    x\mbox{ }\renzbd\mbox{ }y&\mbox{b}_{\not =}&<-x-                         &x\mbox{ }bbbb\mbox{ }y\\
                             &          &\;\;\;\;\;\;\;\;\;\;\;\;\;\;\;\;-y->&\\  \hline
    x\mbox{ }\renzfd\mbox{ }y&\mbox{f}_{\not =}&-x->                         &x\mbox{ }ffff\mbox{ }y\\
                             &          &\;\;\;\;\;\;\;\;\;\;\;\;\;\;\;\;<-y-&\\  \hline
    x\mbox{ }\renzmbe\mbox{ }y&\mbox{mb}_=&-x->                      &x\mbox{ }efbs\mbox{ }y\\
    ................................................&..........&&\\
    y\mbox{ }\renzmfe\mbox{ }x&\mbox{mf}_=&\;\;\;\;\;\;\;\;\;\;\;-y->&y\mbox{ }bsef\mbox{ }x\\  \hline
    x\mbox{ }\renzmbd\mbox{ }y&\mbox{mb}_{\not =}&<-x-                      &x\mbox{ }sbsb\mbox{ }y\\
                              &                  &\;\;\;\;\;\;\;\;\;\;\;-y->&\\  \hline
    x\mbox{ }\renzmfd\mbox{ }y&\mbox{mf}_{\not =}&-x->        &x\mbox{ }fefe\mbox{ }y\\
                              &    &\;\;\;\;\;\;\;\;\;\;\;<-y-&\\  \hline
    x\mbox{ }\renzobe\mbox{ }y&\mbox{ob}_=&-x->          &x\mbox{ }ifbi\mbox{ }y\\
    ................................................&..........&&\\
    y\mbox{ }\renzofe\mbox{ }x&\mbox{of}_=&\;\;\;\;\;-y->&y\mbox{ }biif\mbox{ }x\\  \hline
    x\mbox{ }\renzobd\mbox{ }y&\mbox{ob}_{\not =}&<-x-   &x\mbox{ }ibib\mbox{ }y\\
                              &           &\;\;\;\;\;-y->&\\  \hline
    x\mbox{ }\renzofd\mbox{ }y&\mbox{of}_{\not =}&-x->   &x\mbox{ }fifi\mbox{ }y\\
                              &           &\;\;\;\;\;<-y-&\\  \hline
    x\mbox{ }\renzce\mbox{ }y&\mbox{c}_=&\;\;\;-x->                    &x\mbox{ }bfii\mbox{ }y\\
    ................................................&..........&&\\
    y\mbox{ }\renzee\mbox{ }x&\mbox{e}_=&\mbox{---}\;\;y\;\;\mbox{---}>&y\mbox{ }iibf\mbox{ }x\\  \hline
    x\mbox{ }\renzcd\mbox{ }y&\mbox{c}_{\not =}&\;\;\;<-x-                    &x\mbox{ }fbii\mbox{ }y\\
    ................................................&..........&&\\
    y\mbox{ }\renzed\mbox{ }x&\mbox{e}_{\not =}&\mbox{---}\;\;y\;\;\mbox{---}>&y\mbox{ }iifb\mbox{ }x\\  \hline
    x\mbox{ }\renzcbe\mbox{ }y&\mbox{cb}_=&-x->                          &x\mbox{ }sfsi\mbox{ }y\\
    ................................................&..........&&\\
    y\mbox{ }\renzefe\mbox{ }x&\mbox{ef}_=&\mbox{---}\;\;y\;\;\mbox{---}>&y\mbox{ }sisf\mbox{ }x\\  \hline
    x\mbox{ }\renzcbd\mbox{ }y&\mbox{cb}_{\not =}&<-x-                          &x\mbox{ }ebis\mbox{ }y\\
    ................................................&..........&&\\
    y\mbox{ }\renzebd\mbox{ }x&\mbox{eb}_{\not =}&\mbox{---}\;\;y\;\;\mbox{---}>&y\mbox{ }iseb\mbox{ }x\\  \hline
    x\mbox{ }\renzcfe\mbox{ }y&\mbox{cf}_=&\;\;\;\;\;-x->                &x\mbox{ }beie\mbox{ }y\\
    ................................................&..........&&\\
    y\mbox{ }\renzebe\mbox{ }x&\mbox{eb}_=&\mbox{---}\;\;y\;\;\mbox{---}>&y\mbox{ }iebe\mbox{ }x\\  \hline
    x\mbox{ }\renzcfd\mbox{ }y&\mbox{cf}_{\not =}&\;\;\;\;\;<-x-                &x\mbox{ }fsei\mbox{ }y\\
    ................................................&..........&&\\
    y\mbox{ }\renzefd\mbox{ }x&\mbox{ef}_{\not =}&\mbox{---}\;\;y\;\;\mbox{---}>&y\mbox{ }eifs\mbox{ }x\\  \hline
    x\mbox{ }\renzeqe\mbox{ }y&\mbox{eq}_=&-x->&x\mbox{ }sese\mbox{ }y\\
                              &           &-y->&\\  \hline
    x\mbox{ }\renzeqd\mbox{ }y&\mbox{eq}_{\not =}&-x->&x\mbox{ }eses\mbox{ }y\\
                              &                  &<-y-&\\  \hline
    \end{array}
    $
    \end{center}
    \end{scriptsize}
\caption{The 26 base relations of the directed intervals algebra, and their translation
into the RA $\dlalg$.}\label{from-dia-to-pat}
\end{figure}
\subsection{Reasoning about parallel-to-the-axes parallelograms}
Approaches to reasoning about parallelograms of the 2D space, whose sides are
parallel to the axes of some system $(x,O,y)$ of coordinates, can be found in the literature
\cite{Balbiani98a,Guesgen89a,Mukerjee90a}. These approaches
are straighforward extensions of Allen's algebra of temporal intervals
\cite{Allen83b}. Given such a parallelogram, say $P$, we denote by $P^x$ and $P^y$ the
intervals consisting of the projections of $P$ on the $x$- and $y$-axes, respectively. The
atoms of the corresponding rectangle algebra, $\rectalg$, are of the form $(r_1,r_2)$, where
$r_1$ and $r_2$ are Allen's atoms \cite{Allen83b} (the Allen's atoms are $\allenb$ (before),
$\allenm$ (meets), $\alleno$ (overlaps), $\allens$ (starts), $\allend$ (during), $\allenf$
(finishes); their respective converses $\allena$ (after), $\allenmi$ (met-by), $\allenoi$
(overlapped-by, $\allensi$ (started-by), $\allendi$ (contains), $\allenfi$ (finished-by); and
$\alleneq$ (equals), which is its proper converse). If $P_1$ and $P_2$ are two parallelograms
as described, then:
\begin{equation}
(r_1,r_2)(P_1,P_2)\Leftrightarrow r_1(P_1^x,P_2^x)\wedge r_2(P_1^y,P_2^y)
\end{equation}
The translation of the atoms of the rectangle algebra into the RA $\dlalg$ can thus be obtained
from the procedure already described on translating into $\dlalg$ Moratz et al.'s relations
\cite{Moratz00a}, on the one hand, and Renz's relations \cite{Renz01a}, on the other hand (see
Figure \ref{from-dia-to-pat}).
\section{Conclusion and further work}\label{conclusion}
We have presented a Relation Algebra (RA) \cite{Tarski41b,Ladkin94a,Isli00b} of relative
position relations on 2-dimensional directed lines ($\dlignes$). The converse table, the
rotation table and the composition tables of the RA have been provided. Furthermore,
thanks to its inspiration from the theory of degrees of freedom analysis
\cite{Hartenberg64a,Kramer92a}, the work can be seen as answering, at least partly, the
challenges in \cite{Randell92c} for the particular case of qualitative spatial
reasoning: computing, for instance, the composition of two relations is derived from:
\begin{enumerate}
  \item the composition of the rotational projections of the two relations, on the one
    hand; and
  \item the composition of the translational projections, on the other hand.
\end{enumerate}
More importantly, current research shows clearly the importance of developing
spatial RAs: specialising an $\alcd$-like Description Logic (DL) \cite{Baader91a}, so that the
roles are temporal immediate-successor (accessibility) relations, and the concrete domain is generated by a
decidable spatial RA in the style of the well-known Region-Connection Calculus RCC-8 \cite{Randell92a},
such as the RA $\cdlalg$ defined in this paper, leads to a computationally well-behaving family of
languages for spatial change in general, and for motion of spatial scenes in particular:
\begin{enumerate}
  \item Deciding satisfiability of an $\alcd$ concept $\wrt$ to a cyclic TBox is, in
    general, undecidable (see, for instance, \cite{Lutz01b}).
  \item In the case of the spatio-temporalisation, however, if we use what is called
    weakly cyclic TBoxes in \cite{Isli03a} (see also the related work \cite{Isli02b}), then
    satisfiability of a concept $\wrt$ such a TBox is decidable. The axioms of a
    weakly cyclic TBox capture the properties of modal temporal operators. The
    reader is referred to \cite{Isli03a} for details.\footnote{A full version of
    \cite{Isli03a}, including the decidability proof, will be made downloadable soon
    \cite{Isli02d}.}
\end{enumerate}

Extending the presented RA to 3D would allow, for instance, for the representation of 3D
shapes, such as polyhedra, which are the 3D counterpart of polygons in the 2D space, which
are themselves the 2D counterpart of (convex) intervals in the 1D space. This could be
achieved thus:
\begin{enumerate}
  \item extend Isli and Cohn's RA $\atra$ \cite{Isli98a,Isli00b} to orientations of the 3D space;
  \item extend the $\dltalg$ algebra to directed lines of the 3D space; and
  \item combine the two calculi, as done in this work for the 2D counterparts, to get a
    calculus for reasoning about relative position of directed lines of the 3D space.
\end{enumerate}
\begin{tiny}
\bibliographystyle{plain}
\bibliography{biblio}
\end{tiny}
\newpage\noindent
\appendix
\section{Checking that $\dlalg$ satisfies the ternary RA properties}\label{appendixb}
$\dlalg$ is the structure
$\dlalg =\langle 2^{\dlats},\cup ,\cap ,^-,\emptyset ,\dlats ,\circ ,^\smile ,^\frown ,\traidelt\rangle$, where:
\begin{enumerate}
  \item the set $2^{\dlats}$ of subsets of $\dlats$ is the set of elements, or relations, of $\dlalg$;
  \item the Boolean operations of addition, product and complement are given by the set-theoretic operations of
    union ($\cup$), intersection ($\cap$) and complement ($^-$);
  \item the empty set provides the empty, or bottom, element of $\dlalg$;
  \item the set $\dlats$ of atoms provides the universal, or top, element of $\dlalg$;
  \item the composition, converse and rotation of elements of $\dlalg$ are, respectively, the operations
    $\circ$, $^{\smile}$ and $^{\frown}$, as defined in \cite{Isli00b} ---see also Section \ref{csps},
    Equations (\ref{tothree})-(\ref{tofour})-(\ref{tofive}); and
  \item the identity element is given by $\traidelt =\{(a,b,b):a,b\in\dlines\}$.
\end{enumerate}
The verification that $\dlalg$ satisfies the nine ternary RA properties
(\ref{tra-pone})$-\cdots -$(\ref{tra-pnine}) is done in a similar way as for $\atra$ -see \cite{Isli00b},
Appendix B. The only thing that remains to be checked is that the converse table, the rotation table and the
composition tables record, respectively, the exact converses, the exact rotations and the exact compositions.
But this follows straighforwardly from the facts:
\begin{enumerate}
  \item that  $\atra$ is an RA \cite{Isli00b}, and
  \item that Vilain and Kautz's calculus of time points \cite{Vilain86a} is an RA \cite{Ladkin94a}. \cqfd
\end{enumerate}
\end{document}